\documentclass{llncs}
\usepackage[a4paper, margin=1in]{geometry}
\usepackage{graphicx}
\usepackage{bussproofs}
\usepackage{amsmath}
\usepackage{amssymb}
\usepackage{multicol}
\usepackage{fancyvrb}
\usepackage{qtree}
\usepackage{enumerate}
\usepackage{url}
\usepackage{listings}
\usepackage{holindex}
\usepackage{alltt}

\renewcommand{\HOLTokenProd}{$\cdot$}

\begin{document}

\title{Formalized Lambek Calculus in Higher Order Logic (HOL4)}
\author{Chun Tian}
\institute{Scuola di Scienze, Universit\`{a} di Bologna\\
\email{chun.tian@studio.unibo.it}\\
Numero di matricola: 0000735539}
\maketitle

\begin{abstract}
In this project, a rather complete proof-theoretical formalization of Lambek
Calculus (non-associative with arbitrary extensions) has been ported
from Coq proof assistent to HOL4 theorem prover, with some
improvements and new theorems.

Three deduction systems (Syntactic Calculus, Natural Deduction and Sequent Calculus)
of Lambek Calculus are defined with many related theorems proved.
The equivalance between these systems are formally proved. Finally, a
formalization of Sequent Calculus proofs (where Coq has built-in
supports) has been designed and implemented in HOL4. Some basic
results including the sub-formula properties of the
so-called ``cut-free'' proofs are formally proved.

This work can be considered as the preliminary work towards a language
parser based on category grammars which is not multimodal but still has
ability to support context-sensitive languages through customized
extensions.
\end{abstract}

\section{Introduction}

\subsection{Phrase structure grammars and Chomsky Hierarchy}

Roughly speaking, sentences in formal and natural languages are
constructed by sequences of smaller units, i.e. phrases and words. And determining
if any given sequence of phrases or words (coming from a finite lexicon of
certain language) forms a grammatically
correct sentence in that language, and when it's correct, finding out the
``structure'' (and even the ``meaning'') of that sentence, are the
central goals in both linguistics and Natural Language Processing in
computer science.

Since the year when Noam Chomsky published his famous ``Hierarchy''
\cite{CHomsky:1959ty} and his work on phrase structures of English
and Spanish \cite{Chomsky:2009vr} in 1950s, the concepts of \emph{context-free} and
\emph{context-sensitive} languages (and the intemediate areas between
them) with the uses of \emph{rewriting rules} to
represent the phrase structure grammar of any given language, has
dominated the parsing theory until today.

The rule-based grammar for context-free languages and certain
more restricted languages, together with the related parsing theory
and algorithms, worked so well in handing artificial languages
(formal languages and programming languages). But parsing natural
languages has always been a complex and difficult area.

The first clue is that, natural languages are NOT context-free,
although the precise
  statement only says: \emph{there exist some natural languages which are not
  context-free}, and the proof is totally based on some rare
  phenomenons in certain languages, e.g. the so-called
  \emph{cross-serial dependencies} in Swiss German and Dutch. For most
  of other natural languages, e.g.  English and Italian, a proof that
  these languages are not context-free, doesn't exist yet. Therefore,
  it's quite fair to say, after ignoring some rarely used part, all
  natural languages can indeed be defined by context-free grammars.

While it's quite common for any programming langauge compiler to use a
complete, hand-written or program-generated grammar to describe their targeting language,
the task of writing down all the grammar rules for a natural language,
e.g. English and Italian, is incredibly hard. (Nevertherless they do exist for
English). Efficiency is not a big problem (given the fact that each
sentence as parsing unit is relative small, e. g.. almost always less
than 100 words), ambiguities are indeed problems, but it's now
generally accepted for any grammar parsing tool to output more than
one possible interpretations for any given sentences (then there're
higher level linguistic theories to help disambiguiting them).

The real problem is, the grammar rules for each languages are so
different, none of rewriting rules can be considered as
\emph{universal} to all human languages. For example, one may think
that all sentences have a Noun Phrase (NP) followed by a Verb Phrase
(VP) at top level, e.g. in the Italian sentence ``molti esperti
arriveranno'', NP is ``molti esperti'' and VP is ``arriverranno''.
But in Italian one can also put the subject after the verb and say
``arriveranno esperti'' instead. If we have to maintain the order
between NP and VP, then this new sentence must be interpreted in a
different way (see p. 33 of \cite{Napoli:1988en}) with an empty NP at
the beginning (see Fig \ref{fig:1}). \footnote{Relaxing the order of NP and VP
in the grammar is not an option, because it will be illegal in English}
\begin{figure}
\label{fig:1}
\begin{multicols}{2}
\Tree [.S [.NP \textit{(Molti esperti)} ] [.VP \textit{arriveranno} ] ]

\Tree [.S [.NP $\epsilon$ ] [.VP [.V \textit{Arriveranno} ] [.NP
\textit{(molti esperti)} ] ] ]
\end{multicols}
\end{figure}.
Here the main argument is, phrase structures of the two sentences
shouldn't be such different when Italian speakers simply wanted to
emphasize the action (Verb) by speaking that word first. But within the
framework of Chomksy, such complexity is inevitable.

Chomsky further developped his phrase structure theory into a
more complex theory called ``Government and Binding'' (GB) in 1990s, and
by distinguishing the so-called \emph{surface structure} and
\emph{deep structure}, different speaking ways of the same sentence
now can be finally turned into the same structure. This new GB theory is
dedicated for parsing only natural languages, but the attepmts to find
evidences for the existence of universal grammars failed again.

Several years later, Chomsky almost completely abandoned GB theory and
turned into the ``Minimalist Program'', which this new
grammar theory is much simplier than GB theory while is still capable to analyze the
phrase structure of context-sensitive languages.

\subsection{Categorial Grammars and Lambek Calculus}

On the other side, different grammar theories have been introduced in
parallel. One of them is the so-called \emph{Categorial Grammar}.

Categorial Grammar was firstly introduced by Polish philosopher and logician
Kazimierz Ajdukiewicz in 1935 \cite{Ajdukiewics:2011wa}, which is
based on ideas from precedent Polish logicians. In this whole new
grammar system, Ajdukiewicz
assigned each word (or lexical entry) a ``category'' which is defined
inductively:
\begin{enumerate}
\item \emph{Basic categories}. The two primitive types $n$ (for entities or individuals or first order
terms) and $s$ (for propositions or truth values) are categories.
\item \emph{Functor categorites}. Whenever $N$ is a category and $D_1,\ldots,D_n$ is a sequence or multiset of
categories, then $\displaystyle\frac{N}{D_1 \cdots D_n}$ is itself a category.
\end{enumerate}

To decide the category of syntactically connected expressions, the following rules are used:
\begin{enumerate}
\item a word or lexical entry is syntactically connected, and its
  category is the basic their assigned category,
\item given $n$ syntactically connected expressions $d1,\ldots,d_n$ of respective categories
$D_1,\ldots,D_n$, and an expression $f$ of category $\displaystyle\frac{N}{D_1
  \cdots D_n}$, the expression $f d_1 \cdots d_n$ (or \emph{any permutation of it!}) is syntactically connected and
has category $N$.
\end{enumerate}

There're two notable observations about Ajdukiewicz's categorial
grammar:
\begin{enumerate}
\item The calculus of categorites works just like elementary
  arithmetics: functor categorites are divisions of categories, and
  syntactic connections are multiplications of categories. Thus if one
  word has category $\displaystyle\frac{A}{B}$, the other has category $B$, their
  connecion has category $\displaystyle\frac{A}{B} \cdot B = A$.
\item Although the original purpose of the grammar is to check logical
  formulae in Polish notation, in which the word order is not really a
  problem, but we can imagine that, for highly inflected languages
  like Latin and Sanskrit, which has rather flexible word orders, this
  category grammar may result into quite compat while still correct
  grammars (for at least a small portion). Nevertheless it's impossible to handle English.
\end{enumerate}

In 1953, just two years before Chomsky published his phrase structure grammar
theory and the famous hierarchy, Israeli philosopher, mathematician,
and linguist Yehoshua Bar-Hillel made an important
enhancement \cite{BarHillel:1953dd} to Ajdukiewicz's categorial
grammar. The resulting new categorial grammar is now called
\emph{Ajdukiewicz-Bar grammar} or \emph{AB grammar}.

In AB grammar, word orders are now respected, and the single
``division'' operator in categories are not divided into two different
directions, i.e. $\displaystyle\frac{A}{B}$ now becomes $A/B$ (read as ``$A$ over
$B$'') and $A\setminus B$ (read as ``$B$ under $A$'', and the
syntactic connections cannot be permutated with changing the overall
category, i.e. in general, $A B \neq B A$. In formal languages, the
so-called \emph{category} is defined as follows:
\begin{equation*}
C ::= P \quad | \quad (C / C) \quad | \quad (C \setminus C)
\end{equation*}
where $P$ is a basic category usually containing $S$ (for sentences)
and $np$ (for noun phrases) and $n$ (for nouns). And the only rules in
this grammar systems are, $\forall u,v \in C$,
\begin{align*}
u (u \setminus v) &\longrightarrow v\qquad (\setminus_e) \\
(v / u) u &\longrightarrow v\qquad (/_e)
\end{align*}

Although AB grammar system looks quite simple, it's proved that AB
grammar is weakly equivalent to a corresponding context-free
grammar, and there're algorithms to learn AB grammars from parsing
results. \cite{Moot:2012br}.
However, having only elimination rules, it's impossible to derive fomulae like
$(x/y) (y/z) \longrightarrow (x/z)$, because the only purposes of the
rules is to reduce two categories into a small one, which is part of
the two categories. AB grammar is just a rule-based system but formal system.

Finally, in 1958, Joachim Lambek \cite{Lambek:1958fh} succesfully
defined a formal system for syntactic calculus in which all category formulae can
be derived from a basic set of rules (as axioms) and basic logic formulae. It's later called
\emph{(Associative) Lambek Calculus}, or \textbf{L}. In Lambek
Calculus, now the syntactic connections are representated explicitly
by a new connective ``dot'' ($\cdot$) and the associativity $(A \cdot B)
\cdot C = A \cdot (B \cdot C)$ is assumed as axiom (or rule). Thus,
the category in Lambek calculus is defined as follows:
\begin{equation*}
L ::= P \quad | \quad (L / L) \quad | \quad (L \setminus L) \quad | \quad L \cdot L
\end{equation*}
where $P$ is a basic category. Lambek introduced a relation
$\rightarrow$ and use $x\rightarrow y$ to  indicate that any
expression (string of words) of type (or category) $x$ also has type
$y$, and the following rules were used as axioms of the formal system:
\begin{enumerate}
\item[(a)] $x\rightarrow x$;
\item[(b)] $(x \cdot y) \cdot z \rightarrow x \cdot (y \cdot z)$;
\item[(b')] $x \cdot (y \cdot z) \rightarrow (x \cdot y) \cdot z$;
\item[(c)] if $x \cdot y \rightarrow z$, then $x \rightarrow z/y$;
\item[(c')] if $x \cdot y \rightarrow z$, then $y \rightarrow
  x\setminus z$;
\item[(d)] if $x\rightarrow z/y$, then $x \cdot y \rightarrow z$;
\item[(d')] if $y \rightarrow x\setminus z$, then $x \cdot y
  \rightarrow z$;
\item[(e)] if $x\rightarrow y$ and $y\rightarrow z$, then
  $x\rightarrow z$.
\end{enumerate}
Here are a few observations about Lambek Calculus:
\begin{enumerate}
\item Although Lambek calculues has more rules then AB grammar, but
 they're actually equivalent: ``a set of strings of words forms a categorial language
of one type if and only if it forms a categorial language of the other
type'' \cite{Cohen:1967iy};
\item The $\rightarrow$ relation is reflexisive (by (a)) and
  transitive (by (e)), but it's not symmetric, i.e. in general
  $x\rightarrow y \nRightarrow y \rightarrow x$, thus it's not an
  equivalence relation at all.
\item The dot ($\cdot$) connective usually doesn't appears in the lexion, in
  which each word is associated with one or more categories which is
  made by only basic categories and two connectives slash ($/$) and
  backslash ($\setminus$).
\item Lambek calculus can only be used to decide if a given string of
  words has a specific category (e. g. the category of a legal
  sentence, $S$), the phrase structure, however, is not immediately
  known.
\item Lambek calculus has a decision procedure through Gentzen's
  Sequent Calculus and the corresponding Cut-elimination theorem. This was proved by Lambek (1958)
  \cite{Lambek:1958fh}.
\item Lambek calculus is NP-complete \cite{Pentus:2006ct}, L-complete
  \cite{Pentus:1993wd}, and Lambek grammars are context-free \cite{Pentus:1993cy}.
\end{enumerate}

Among other issues, to resolve the key limitation that Lambek calculus cannot be
used as a language parser, in 1961, Lambek introduced the so-called Non-associative Lambek Calculus
\cite{Lambek:1961bx}, or \textbf{NL}. There're two major modifications comparing to
previous associative Lambek Calculus:
\begin{enumerate}
\item The associative rule (b) and (b') have been removed from axiom
  rules, and all theorems derivated from them are not derivable any
  more. It's found that such changes also removed some strange
  non-language sentences which are acceptable before.
\item To decide the category of a sentence, now the first step is to
  put all the words into a binary tree with the original order
  respected, i. e. the \emph{bracketing} process, then applications of
  inference rules are limited at each tree node.
\end{enumerate}

Using as a language parser, non-associative Lambek Calculus works in this
way: for any given string of words, if there exists a bracketing such
that the resulting category is the target category (e. g. $S$), then
it's grammatically correct and the corresponding binary tree represent
the phrase structures, with the category of each node representing the
phrase structure at different levels. It's proven that non-associative
Lambek Calculus also has a decision procedure, which surprisingly is
polynomial. And for product-free \textbf{NL} (there's no dots in lexicon), the parsing
process is also polynomial. (see Chapter 4.6 of \cite{Moot:2012br} for
detailed discussions and links tooriginal papers)

Beside \textbf{L} and \textbf{NL}, there's also \textbf{NLP} in which
the commutativity of products is respect based on \textbf{NL}, i. e. $x \cdot y = y \cdot
x$. With such assumption, the resulting grammar is quite similar to the
original Ajdukiewicz grammar.  When both commutativity and
associativity are respected, the resulting Lambek calculus is called \textbf{LP}.

\subsection{Categorial grammars and universal grammar}

Chomsky believes that there's an innate \emph{universal grammar} in human mind, and
when children start learning languages from very limited vocabulary
and extremely incomplete corpus (where machine learning is impossible
to converge), what they actually do is to adapt this universal grammar to their
mother languages. However, from Chomsky's phrase structure theory and
the related grammars based on rewriting rules, terminals and
non-terminals, we can't see any evidence for the existence of
Universal gammars.

On the other side, in Categorial grammars we see two separated parts:
\begin{enumerate}
\item Language independent part: the inference rules for categories;
\item Language dependent part: the lexicon.
\end{enumerate}
Therefore, if we treat the first part at ``unversal grammar'', what
remains to be learn by children (and machine learning program), is just the lexion which associates
words to their syntactic categories (which then indicate how a word
gets used in this language) and their meanings.  This connection (to
universal grammars) has made
Category Grammars quite attractive. Maybe the language parsing process
based on categorial grammars are more natural and close to the natural
language processing in human mind.

\subsection{Categorial Type Logics}

Since natural language is not context-free, while either \textbf{L}
and \textbf{NL} can only handle context-free languages, people seek
ways to extend Lambek Calculus to support context-sensitive
languages. On the other side, none of \textbf{L}, \textbf{NL} and
\textbf{NLP} is perfect for all languages, thus people seek ways to
combine different Lambek calculus and use them for different portions
of the target language.

One such attempts is the so-called \emph{Multimodal Lambek
  Calculus}. In this calculus system, there're many copies of category
connectives, each associated with a mode index $i$:
\begin{equation*}
L ::= P \quad | \quad (L /_i L) \quad | \quad (L \setminus_i L) \quad | \quad L \cdot_i L
\end{equation*}
Different versions of Lambek Calculus were identified with different
mode identifiers, and properties like commutativity and associativity
are respected for only connectives of certain modes. Then there's
structure rules to bring different modes together. (see Chapter 5 of \cite{Moot:2012br}).

Even further extension includes the unary connectives from linear
logics: $\Diamond$ and $\Box$.  With all such things, the resulting
calculus system has so many rules and operators, and goes far beyond
Lambek Calculus, has a new name called \emph{Categorial Type Logics (CTL)}.

Categorial Type Logics  has the folloing characteristics:
\begin{enumerate}
\item From the view of proof theory, there're still decision procedures, and the cut-elimination
  theorem can still be proved. However, to find the proofs for
  category derivations, more complex algorithms must be used
  (e.g. proof nets) and the time complexity is at least NP-complete.
\item From the view of model theory, CTL were proved to be sound and
  complete.
\item Learning CTL grammars from corpus has no known results. Given
the fact that the structure of categories now contains more
connectives in different modes, the learning process is much harder
than the normal uni-modal Lambek calculus.
\end{enumerate}
Thus CTL is not quite practical so far, as one of the main authors
said in his book: \emph{``So the big open question for multimodal categorial grammars is to find the `right'
combination of structural rules both from the point of view of being able to make
the relevant linguistic generalizations and from the point of view of computational
complexity.''.} (page. 184 of \cite{Moot:2012br})

In this paper, we have tried another possibility to extend Lambek
Calculus. We think \textbf{NL} is good enough as a basis, because it's
as simple as rewriting rules in context-free grammars, and it
has known polinomial parsing alghrithm and reasonable learning
algorithms. To handle certain context-sentive portions of the target language, we
use restrictive \emph{extensions} to extend the \textbf{NL} inference
rules. These extensions, when satisfying certain restrictions,
won't break the validity of existing parsing and learning algorithms
while it's possible to handle context-sensitive languages.

\subsection{Trusted software and theorem provers}

Software cannot be trusted in general, their outputs may be wrong due
to either issues in their program code, or issues in the development
platform in which they were written. On the other side, either
operating systems or development platform (including programming
languages) evolve quickly. Any program, once being
written, must be maintained continuously, otherwise it soon becomes unusable.

Therefore, we try to convince the audiences the following principles:
\begin{enumerate}
\item For any research task in computer science, the last solution is to write
  a whole new software. Because once the research is done (or paper is
  published), the software may not be well maintained any more, and in
  a few years it's dead.
\item Choose programming languages which has long history (at least 10
  years), with more than one stable implemenations.
\item Whenever possible, generate program code from higher level tools
  instead of writing them directly.
\end{enumerate}

On the other side, to gurantee the correctness of algorithms or their
implementations, it's preferred to use theorem provers. Many theorem
proving software have the ability to generte program code from
proved theorems. Coq \footnote{\url{https://coq.inria.fr}} and
Isabelle \footnote{\url{http://isabelle.in.tum.de}} are such examples.

Sometimes, the theorem prover itself is written as an
extension to the development platform, and in this case user
can continue extending the theorem prover to make the whole platform
into any application software. ACL2 \footnote{\url{http://www.cs.utexas.edu/users/moore/acl2/}} and HOL4
\footnote{\url{https://hol-theorem-prover.org}} are two notable
examples written in this way.

\subsection{Why Higher Order Logic?}

Comparing to other theorem provers, the Higher Order Logic (HOL) family of
theorem provers (HOL4, HOL light, Isabelle) has relatively simple
logic foundation: just simple typed lambda calculus with type
variables. Comparing to the logic foundation of Coq (Calculus of
Inductive Construction, CIC), higier order logic is easier to
understand and use, because it has few primitive derivation rules, and
students who has just finished studying $\lambda$-calculus (typed and untyped) could
easily get started with HOL.

HOL has also a small and verified kernel. And when HOL is implemented in Standard ML language, which is
the case of HOL4 and Isabelle, the
programming platform itself could be also formally verified \footnote{The
  CakeML project (\url{https://cakeml.org}) represents one such
  efforts.}

Therefore, we choose HOL4 for several reasons:
\begin{enumerate}
\item It's a theorem prover well maintained and with long history (30
  years). Wigh high chances it will continue living for next 30 years, so is the proof scripts
  we wrote in it today.
\item The logical foundation of HOL4 is easy to understand, and the
  primitive inference rule set is small and clear.
\item HOL4 has a rich builtin theory library and a very large set of
  example code library for reference purposes.
\item The software is written as a natural extension to Standard ML
  programming language. The theorem prover, the proof scripts, and the
  tacticals, they're all written in Standard ML. In theory, user can
  build any customized
  software upon the theorem prover. Besides, the CakeML project
  has demonstrated how to write a formally verified programming
  language compiler and interpreter in HOL4.
\item The type variable in HOL4 types are especially suited for
expresing syntactic categories in Categorial Grammars. (this will be explained in next section).
\end{enumerate}

Some more differences between HOL and Coq will be discussed later, in
section \ref{sec:differences-hol-coq}.

\section{Lambek Calculus in HOL4}

In this project, we have implemented Lambek Calculus in HOL4. The
exact variant of Lambek Calculus is \textbf{NL} (non-associative) with arbitrary
extensions. Three deduction systems are implemented: (Axiomatic)
Syntactic Calculus, Natural Deduction, and Gentzen's Sequent
Calculus. \footnote{Project code can be found at \url{https://github.com/binghe/informatica-public/tree/master/lambek}}

This work is based on an implementation of Lambek Calculus
\footnote{\url{https://github.com/coq-contribs/lambek}} in Coq
proof assistant. In fact, our work so far can be considered as a
partial porting of the Coq-based proof scripts: most definitions and theorems are
from previous work, with necessary modifications. Here are some major
differences:
\begin{enumerate}
\item Whenever a relation can be defined as a reflexitive transivive
  closure (RTC) of another relation, instead of define it manuall as
  in Coq, now we use HOL's builtin relationTheory to define it and get the related
  theorems.
\item Coq's built-in support on proofs are not directly portable to
  HOL, therefore we defined whole new data structures and fill the
  gaps.
\end{enumerate}

\subsection{Syntactic Categories}

In HOL, syntactic categories are defined by an algebraic datatype
\HOLinline{\HOLFreeVar{Form}} with a type variable $\alpha$:
\begin{lstlisting}
Datatype `Form = At 'a | Slash Form Form | Backslash Form Form | Dot Form Form`
\end{lstlisting}
In this way, all theorems we proved are actually theorem schemas in
which there's always a type variable. In practice, user can either
define another data structure with enumerated categories ($S, n, np,
\ldots$), or directly use strings. For example, in later mode, the basic category
``$S$'' can be simply represented as \HOLinline{\HOLConst{At} \HOLStringLit{S}}, and category
``$S/np$'' will be \texttt{Slash (At "S") (At "np")} or \HOLinline{\HOLConst{At} \HOLStringLit{S} \HOLSymConst{/} \HOLConst{At} \HOLStringLit{np}} when grammar supports are enabled.

\subsection{Syntactic Calculus}

(Axiomatic) Syntactic Calculus is a theory about the
\HOLinline{\HOLConst{arrow}} relation which indicates that a string of words having
the first category will also have the seoncd category, it's
reflexitive and transitive. In HOL4, such a inductive relation
can be easily defined by \texttt{Hol_reln} \cite{Anonymous:Bxz1gZYL}
command:
\begin{lstlisting}
val (arrow_rules, arrow_ind, arrow_cases) = Hol_reln `
    (!X A. arrow X A A) /\
    (!X A B C. arrow X (Dot A B) C ==> arrow X A (Slash C B)) /\
    (!X A B C. arrow X A (Slash C B) ==> arrow X (Dot A B) C) /\
    (!X A B C. arrow X (Dot A B) C ==> arrow X B (Backslash A C)) /\
    (!X A B C. arrow X B (Backslash A C) ==> arrow X (Dot A B) C) /\
    (!X A B C. arrow X A B /\ arrow X B C ==> arrow X A C) /\
    (!(X :'a arrow_extension) A B. X A B ==> arrow X A B) `;
\end{lstlisting}
which is then broken into the following separated rules:
\begin{alltt}
one:
\HOLTokenTurnstile{} \HOLConst{arrow} \HOLFreeVar{X} \HOLFreeVar{A} \HOLFreeVar{A}
beta:
\HOLTokenTurnstile{} \HOLConst{arrow} \HOLFreeVar{X} (\HOLFreeVar{A} \HOLSymConst{\HOLTokenProd{}} \HOLFreeVar{B}) \HOLFreeVar{C} \HOLSymConst{\HOLTokenImp{}} \HOLConst{arrow} \HOLFreeVar{X} \HOLFreeVar{A} (\HOLFreeVar{C} \HOLSymConst{/} \HOLFreeVar{B})
beta':
\HOLTokenTurnstile{} \HOLConst{arrow} \HOLFreeVar{X} \HOLFreeVar{A} (\HOLFreeVar{C} \HOLSymConst{/} \HOLFreeVar{B}) \HOLSymConst{\HOLTokenImp{}} \HOLConst{arrow} \HOLFreeVar{X} (\HOLFreeVar{A} \HOLSymConst{\HOLTokenProd{}} \HOLFreeVar{B}) \HOLFreeVar{C}
gamma:
\HOLTokenTurnstile{} \HOLConst{arrow} \HOLFreeVar{X} (\HOLFreeVar{A} \HOLSymConst{\HOLTokenProd{}} \HOLFreeVar{B}) \HOLFreeVar{C} \HOLSymConst{\HOLTokenImp{}} \HOLConst{arrow} \HOLFreeVar{X} \HOLFreeVar{B} (\HOLFreeVar{A} \HOLSymConst{\HOLTokenBackslash} \HOLFreeVar{C})
gamma':
\HOLTokenTurnstile{} \HOLConst{arrow} \HOLFreeVar{X} \HOLFreeVar{B} (\HOLFreeVar{A} \HOLSymConst{\HOLTokenBackslash} \HOLFreeVar{C}) \HOLSymConst{\HOLTokenImp{}} \HOLConst{arrow} \HOLFreeVar{X} (\HOLFreeVar{A} \HOLSymConst{\HOLTokenProd{}} \HOLFreeVar{B}) \HOLFreeVar{C}
comp:
\HOLTokenTurnstile{} \HOLConst{arrow} \HOLFreeVar{X} \HOLFreeVar{A} \HOLFreeVar{B} \HOLSymConst{\HOLTokenConj{}} \HOLConst{arrow} \HOLFreeVar{X} \HOLFreeVar{B} \HOLFreeVar{C} \HOLSymConst{\HOLTokenImp{}} \HOLConst{arrow} \HOLFreeVar{X} \HOLFreeVar{A} \HOLFreeVar{C}
arrow_plus:
\HOLTokenTurnstile{} \HOLFreeVar{X} \HOLFreeVar{A} \HOLFreeVar{B} \HOLSymConst{\HOLTokenImp{}} \HOLConst{arrow} \HOLFreeVar{X} \HOLFreeVar{A} \HOLFreeVar{B}
\end{alltt}

Noticed that, the relation \HOLinline{\HOLConst{arrow}} is a 3-ary relation with the
type ``\HOLinline{\ensuremath{\alpha} \HOLTyOp{arrow_extension} -> \ensuremath{\alpha} \HOLTyOp{arrow_extension}}'', in
which the type abbreviation ``arrow_extension'' is defined as 
\begin{lstlisting}
type_abbrev ("arrow_extension", ``:'a Form -> 'a Form -> bool``);
\end{lstlisting}
An arrow extension defined some extra properties beyond the basic
rules. This connection was made by the \texttt{arrow_plus} theorem.
Thus for any arrow extension \HOLinline{\HOLFreeVar{X}}, \HOLinline{\HOLConst{arrow} \HOLFreeVar{X}} can be considered as a normal
2-ary relation between two categories with the type ``\HOLinline{\ensuremath{\alpha} \HOLTyOp{Form}}''.

Since arrow extensions are nothing but normal relations, the union of
two such relations can be considered as the ``sum'' of two arrow
extensions, and the subset/superset of relations can be considered as
restrictions (or further extensions) to an arrow extension.
Then, the arrow extensions of the four different Lambek Calculus, \textbf{NL}, \textbf{L}, \textbf{NLP} and
\textbf{LP} are defined respectively:
\begin{alltt}
\HOLTokenTurnstile{} \HOLConst{NL} \HOLSymConst{=} \HOLSymConst{∅ᵣ}
\HOLTokenTurnstile{} (\HOLSymConst{\HOLTokenForall{}}\HOLBoundVar{A} \HOLBoundVar{B} \HOLBoundVar{C}. \HOLConst{L} (\HOLBoundVar{A} \HOLSymConst{\HOLTokenProd{}} (\HOLBoundVar{B} \HOLSymConst{\HOLTokenProd{}} \HOLBoundVar{C})) (\HOLBoundVar{A} \HOLSymConst{\HOLTokenProd{}} \HOLBoundVar{B} \HOLSymConst{\HOLTokenProd{}} \HOLBoundVar{C})) \HOLSymConst{\HOLTokenConj{}}
   \HOLSymConst{\HOLTokenForall{}}\HOLBoundVar{A} \HOLBoundVar{B} \HOLBoundVar{C}. \HOLConst{L} (\HOLBoundVar{A} \HOLSymConst{\HOLTokenProd{}} \HOLBoundVar{B} \HOLSymConst{\HOLTokenProd{}} \HOLBoundVar{C}) (\HOLBoundVar{A} \HOLSymConst{\HOLTokenProd{}} (\HOLBoundVar{B} \HOLSymConst{\HOLTokenProd{}} \HOLBoundVar{C}))
\HOLTokenTurnstile{} \HOLConst{NLP} (\HOLFreeVar{A} \HOLSymConst{\HOLTokenProd{}} \HOLFreeVar{B}) (\HOLFreeVar{B} \HOLSymConst{\HOLTokenProd{}} \HOLFreeVar{A})
\HOLTokenTurnstile{} \HOLConst{LP} \HOLSymConst{=} \HOLConst{add_extension} \HOLConst{NLP} \HOLConst{L}
\end{alltt}
in which \texttt{EMPTY_REL} represents an empty relation, and
\HOLinline{\HOLConst{add_extension}} equals to the union of relations. Thus
\textbf{LP} is nothing but a sum of \textbf{NLP} and \textbf{L}.

Then we have proved some common theorems for all Lambek Calculus
(i. e. for whatever arrow extensions):
\begin{alltt}
Dot_mono_right:
\HOLTokenTurnstile{} \HOLConst{arrow} \HOLFreeVar{X} \HOLFreeVar{B\sp{\prime}} \HOLFreeVar{B} \HOLSymConst{\HOLTokenImp{}} \HOLConst{arrow} \HOLFreeVar{X} (\HOLFreeVar{A} \HOLSymConst{\HOLTokenProd{}} \HOLFreeVar{B\sp{\prime}}) (\HOLFreeVar{A} \HOLSymConst{\HOLTokenProd{}} \HOLFreeVar{B})
Dot_mono_left:
\HOLTokenTurnstile{} \HOLConst{arrow} \HOLFreeVar{X} \HOLFreeVar{A\sp{\prime}} \HOLFreeVar{A} \HOLSymConst{\HOLTokenImp{}} \HOLConst{arrow} \HOLFreeVar{X} (\HOLFreeVar{A\sp{\prime}} \HOLSymConst{\HOLTokenProd{}} \HOLFreeVar{B}) (\HOLFreeVar{A} \HOLSymConst{\HOLTokenProd{}} \HOLFreeVar{B})
Dot_mono:
\HOLTokenTurnstile{} \HOLConst{arrow} \HOLFreeVar{X} \HOLFreeVar{A} \HOLFreeVar{C} \HOLSymConst{\HOLTokenConj{}} \HOLConst{arrow} \HOLFreeVar{X} \HOLFreeVar{B} \HOLFreeVar{D} \HOLSymConst{\HOLTokenImp{}} \HOLConst{arrow} \HOLFreeVar{X} (\HOLFreeVar{A} \HOLSymConst{\HOLTokenProd{}} \HOLFreeVar{B}) (\HOLFreeVar{C} \HOLSymConst{\HOLTokenProd{}} \HOLFreeVar{D})
Slash_mono_left:
\HOLTokenTurnstile{} \HOLConst{arrow} \HOLFreeVar{X} \HOLFreeVar{C\sp{\prime}} \HOLFreeVar{C} \HOLSymConst{\HOLTokenImp{}} \HOLConst{arrow} \HOLFreeVar{X} (\HOLFreeVar{C\sp{\prime}} \HOLSymConst{/} \HOLFreeVar{B}) (\HOLFreeVar{C} \HOLSymConst{/} \HOLFreeVar{B})
Slash_antimono_right:
\HOLTokenTurnstile{} \HOLConst{arrow} \HOLFreeVar{X} \HOLFreeVar{B\sp{\prime}} \HOLFreeVar{B} \HOLSymConst{\HOLTokenImp{}} \HOLConst{arrow} \HOLFreeVar{X} (\HOLFreeVar{C} \HOLSymConst{/} \HOLFreeVar{B}) (\HOLFreeVar{C} \HOLSymConst{/} \HOLFreeVar{B\sp{\prime}})
Backslash_antimono_left:
\HOLTokenTurnstile{} \HOLConst{arrow} \HOLFreeVar{X} \HOLFreeVar{A} \HOLFreeVar{A\sp{\prime}} \HOLSymConst{\HOLTokenImp{}} \HOLConst{arrow} \HOLFreeVar{X} (\HOLFreeVar{A\sp{\prime}} \HOLSymConst{\HOLTokenBackslash} \HOLFreeVar{C}) (\HOLFreeVar{A} \HOLSymConst{\HOLTokenBackslash} \HOLFreeVar{C})
Backslash_mono_right:
\HOLTokenTurnstile{} \HOLConst{arrow} \HOLFreeVar{X} \HOLFreeVar{C\sp{\prime}} \HOLFreeVar{C} \HOLSymConst{\HOLTokenImp{}} \HOLConst{arrow} \HOLFreeVar{X} (\HOLFreeVar{A} \HOLSymConst{\HOLTokenBackslash} \HOLFreeVar{C\sp{\prime}}) (\HOLFreeVar{A} \HOLSymConst{\HOLTokenBackslash} \HOLFreeVar{C})
\end{alltt}

The monotonity of \HOLinline{\HOLConst{arrow}} relation with respect to arrow
extensions is proved too (by induction on all basic arrow rules):
\begin{alltt}
mono_X:
\HOLTokenTurnstile{} \HOLConst{arrow} \HOLFreeVar{X} \HOLFreeVar{A} \HOLFreeVar{B} \HOLSymConst{\HOLTokenImp{}} \HOLConst{extends} \HOLFreeVar{X} \HOLFreeVar{X\sp{\prime}} \HOLSymConst{\HOLTokenImp{}} \HOLConst{arrow} \HOLFreeVar{X\sp{\prime}} \HOLFreeVar{A} \HOLFreeVar{B}
\end{alltt}

For Lambek extensions supporting permutations, the following theorems
are proved:
\begin{alltt}
pi:
\HOLTokenTurnstile{} \HOLConst{extends} \HOLConst{NLP} \HOLFreeVar{X} \HOLSymConst{\HOLTokenImp{}} \HOLSymConst{\HOLTokenForall{}}\HOLBoundVar{A} \HOLBoundVar{B}. \HOLConst{arrow} \HOLFreeVar{X} (\HOLBoundVar{A} \HOLSymConst{\HOLTokenProd{}} \HOLBoundVar{B}) (\HOLBoundVar{B} \HOLSymConst{\HOLTokenProd{}} \HOLBoundVar{A})
pi_NLP:
\HOLTokenTurnstile{} \HOLConst{arrow} \HOLConst{NLP} (\HOLFreeVar{A} \HOLSymConst{\HOLTokenProd{}} \HOLFreeVar{B}) (\HOLFreeVar{B} \HOLSymConst{\HOLTokenProd{}} \HOLFreeVar{A})
pi_LP:
\HOLTokenTurnstile{} \HOLConst{arrow} \HOLConst{LP} (\HOLFreeVar{A} \HOLSymConst{\HOLTokenProd{}} \HOLFreeVar{B}) (\HOLFreeVar{B} \HOLSymConst{\HOLTokenProd{}} \HOLFreeVar{A})
\end{alltt}

For Lambek extensions supporting associativity, the following theorems
are proved:
\begin{alltt}
alfa:
\HOLTokenTurnstile{} \HOLConst{extends} \HOLConst{L} \HOLFreeVar{X} \HOLSymConst{\HOLTokenImp{}} \HOLSymConst{\HOLTokenForall{}}\HOLBoundVar{A} \HOLBoundVar{B} \HOLBoundVar{C}. \HOLConst{arrow} \HOLFreeVar{X} (\HOLBoundVar{A} \HOLSymConst{\HOLTokenProd{}} (\HOLBoundVar{B} \HOLSymConst{\HOLTokenProd{}} \HOLBoundVar{C})) (\HOLBoundVar{A} \HOLSymConst{\HOLTokenProd{}} \HOLBoundVar{B} \HOLSymConst{\HOLTokenProd{}} \HOLBoundVar{C})
alfa':
\HOLTokenTurnstile{} \HOLConst{extends} \HOLConst{L} \HOLFreeVar{X} \HOLSymConst{\HOLTokenImp{}} \HOLSymConst{\HOLTokenForall{}}\HOLBoundVar{A} \HOLBoundVar{B} \HOLBoundVar{C}. \HOLConst{arrow} \HOLFreeVar{X} (\HOLBoundVar{A} \HOLSymConst{\HOLTokenProd{}} \HOLBoundVar{B} \HOLSymConst{\HOLTokenProd{}} \HOLBoundVar{C}) (\HOLBoundVar{A} \HOLSymConst{\HOLTokenProd{}} (\HOLBoundVar{B} \HOLSymConst{\HOLTokenProd{}} \HOLBoundVar{C}))
alfa_L:
\HOLTokenTurnstile{} \HOLConst{arrow} \HOLConst{L} (\HOLFreeVar{A} \HOLSymConst{\HOLTokenProd{}} (\HOLFreeVar{B} \HOLSymConst{\HOLTokenProd{}} \HOLFreeVar{C})) (\HOLFreeVar{A} \HOLSymConst{\HOLTokenProd{}} \HOLFreeVar{B} \HOLSymConst{\HOLTokenProd{}} \HOLFreeVar{C})
alfa'_L:
\HOLTokenTurnstile{} \HOLConst{arrow} \HOLConst{L} (\HOLFreeVar{A} \HOLSymConst{\HOLTokenProd{}} \HOLFreeVar{B} \HOLSymConst{\HOLTokenProd{}} \HOLFreeVar{C}) (\HOLFreeVar{A} \HOLSymConst{\HOLTokenProd{}} (\HOLFreeVar{B} \HOLSymConst{\HOLTokenProd{}} \HOLFreeVar{C}))
alfa_LP:
\HOLTokenTurnstile{} \HOLConst{arrow} \HOLConst{LP} (\HOLFreeVar{A} \HOLSymConst{\HOLTokenProd{}} (\HOLFreeVar{B} \HOLSymConst{\HOLTokenProd{}} \HOLFreeVar{C})) (\HOLFreeVar{A} \HOLSymConst{\HOLTokenProd{}} \HOLFreeVar{B} \HOLSymConst{\HOLTokenProd{}} \HOLFreeVar{C})
alfa'_LP:
\HOLTokenTurnstile{} \HOLConst{arrow} \HOLConst{LP} (\HOLFreeVar{A} \HOLSymConst{\HOLTokenProd{}} \HOLFreeVar{B} \HOLSymConst{\HOLTokenProd{}} \HOLFreeVar{C}) (\HOLFreeVar{A} \HOLSymConst{\HOLTokenProd{}} (\HOLFreeVar{B} \HOLSymConst{\HOLTokenProd{}} \HOLFreeVar{C}))
\end{alltt}

\subsection{Associative Lambek Calculus}

For the original associative Lambek Calculus, we have proved all arrow
theorems mentioned in Lambek (1958) \cite{Lambek:1958fh}. All these
theorems are named as \texttt{L_} plus single letters (with optional
prim). The first five are \textbf{L} axioms (but actually they're
proved from previous definition on the arrow relation:
\begin{alltt}
L_a:  \HOLTokenTurnstile{} \HOLConst{arrow} \HOLConst{L} \HOLFreeVar{x} \HOLFreeVar{x}
L_b:  \HOLTokenTurnstile{} \HOLConst{arrow} \HOLConst{L} (\HOLFreeVar{x} \HOLSymConst{\HOLTokenProd{}} \HOLFreeVar{y} \HOLSymConst{\HOLTokenProd{}} \HOLFreeVar{z}) (\HOLFreeVar{x} \HOLSymConst{\HOLTokenProd{}} (\HOLFreeVar{y} \HOLSymConst{\HOLTokenProd{}} \HOLFreeVar{z}))
L_b': \HOLTokenTurnstile{} \HOLConst{arrow} \HOLConst{L} (\HOLFreeVar{x} \HOLSymConst{\HOLTokenProd{}} (\HOLFreeVar{y} \HOLSymConst{\HOLTokenProd{}} \HOLFreeVar{z})) (\HOLFreeVar{x} \HOLSymConst{\HOLTokenProd{}} \HOLFreeVar{y} \HOLSymConst{\HOLTokenProd{}} \HOLFreeVar{z})
L_c:  \HOLTokenTurnstile{} \HOLConst{arrow} \HOLConst{L} (\HOLFreeVar{x} \HOLSymConst{\HOLTokenProd{}} \HOLFreeVar{y}) \HOLFreeVar{z} \HOLSymConst{\HOLTokenImp{}} \HOLConst{arrow} \HOLConst{L} \HOLFreeVar{x} (\HOLFreeVar{z} \HOLSymConst{/} \HOLFreeVar{y})
L_c': \HOLTokenTurnstile{} \HOLConst{arrow} \HOLConst{L} (\HOLFreeVar{x} \HOLSymConst{\HOLTokenProd{}} \HOLFreeVar{y}) \HOLFreeVar{z} \HOLSymConst{\HOLTokenImp{}} \HOLConst{arrow} \HOLConst{L} \HOLFreeVar{y} (\HOLFreeVar{x} \HOLSymConst{\HOLTokenBackslash} \HOLFreeVar{z})
L_d:  \HOLTokenTurnstile{} \HOLConst{arrow} \HOLConst{L} \HOLFreeVar{x} (\HOLFreeVar{z} \HOLSymConst{/} \HOLFreeVar{y}) \HOLSymConst{\HOLTokenImp{}} \HOLConst{arrow} \HOLConst{L} (\HOLFreeVar{x} \HOLSymConst{\HOLTokenProd{}} \HOLFreeVar{y}) \HOLFreeVar{z}
L_d': \HOLTokenTurnstile{} \HOLConst{arrow} \HOLConst{L} \HOLFreeVar{y} (\HOLFreeVar{x} \HOLSymConst{\HOLTokenBackslash} \HOLFreeVar{z}) \HOLSymConst{\HOLTokenImp{}} \HOLConst{arrow} \HOLConst{L} (\HOLFreeVar{x} \HOLSymConst{\HOLTokenProd{}} \HOLFreeVar{y}) \HOLFreeVar{z}
L_e:  \HOLTokenTurnstile{} \HOLConst{arrow} \HOLConst{L} \HOLFreeVar{x} \HOLFreeVar{y} \HOLSymConst{\HOLTokenConj{}} \HOLConst{arrow} \HOLConst{L} \HOLFreeVar{y} \HOLFreeVar{z} \HOLSymConst{\HOLTokenImp{}} \HOLConst{arrow} \HOLConst{L} \HOLFreeVar{x} \HOLFreeVar{z}
\end{alltt}

Based on these axioms, the rest \textbf{L} theorems about arrow
relations can easily be proved by automatic first-order proof
searching:
\begin{alltt}
L_f:  \HOLTokenTurnstile{} \HOLConst{arrow} \HOLConst{L} \HOLFreeVar{x} (\HOLFreeVar{x} \HOLSymConst{\HOLTokenProd{}} \HOLFreeVar{y} \HOLSymConst{/} \HOLFreeVar{y})
L_g:  \HOLTokenTurnstile{} \HOLConst{arrow} \HOLConst{L} (\HOLFreeVar{z} \HOLSymConst{/} \HOLFreeVar{y} \HOLSymConst{\HOLTokenProd{}} \HOLFreeVar{y}) \HOLFreeVar{z}
L_h:  \HOLTokenTurnstile{} \HOLConst{arrow} \HOLConst{L} \HOLFreeVar{y} ((\HOLFreeVar{z} \HOLSymConst{/} \HOLFreeVar{y}) \HOLSymConst{\HOLTokenBackslash} \HOLFreeVar{z})
L_i:  \HOLTokenTurnstile{} \HOLConst{arrow} \HOLConst{L} (\HOLFreeVar{z} \HOLSymConst{/} \HOLFreeVar{y} \HOLSymConst{\HOLTokenProd{}} (\HOLFreeVar{y} \HOLSymConst{/} \HOLFreeVar{x})) (\HOLFreeVar{z} \HOLSymConst{/} \HOLFreeVar{x})
L_j:  \HOLTokenTurnstile{} \HOLConst{arrow} \HOLConst{L} (\HOLFreeVar{z} \HOLSymConst{/} \HOLFreeVar{y}) (\HOLFreeVar{z} \HOLSymConst{/} \HOLFreeVar{x} \HOLSymConst{/} (\HOLFreeVar{y} \HOLSymConst{/} \HOLFreeVar{x}))
L_k:  \HOLTokenTurnstile{} \HOLConst{arrow} \HOLConst{L} (\HOLFreeVar{x} \HOLSymConst{\HOLTokenBackslash} \HOLFreeVar{y} \HOLSymConst{/} \HOLFreeVar{z}) (\HOLFreeVar{x} \HOLSymConst{\HOLTokenBackslash} (\HOLFreeVar{y} \HOLSymConst{/} \HOLFreeVar{z}))
L_k': \HOLTokenTurnstile{} \HOLConst{arrow} \HOLConst{L} (\HOLFreeVar{x} \HOLSymConst{\HOLTokenBackslash} (\HOLFreeVar{y} \HOLSymConst{/} \HOLFreeVar{z})) (\HOLFreeVar{x} \HOLSymConst{\HOLTokenBackslash} \HOLFreeVar{y} \HOLSymConst{/} \HOLFreeVar{z})
L_l:  \HOLTokenTurnstile{} \HOLConst{arrow} \HOLConst{L} (\HOLFreeVar{x} \HOLSymConst{/} \HOLFreeVar{y} \HOLSymConst{/} \HOLFreeVar{z}) (\HOLFreeVar{x} \HOLSymConst{/} (\HOLFreeVar{z} \HOLSymConst{\HOLTokenProd{}} \HOLFreeVar{y}))
L_l': \HOLTokenTurnstile{} \HOLConst{arrow} \HOLConst{L} (\HOLFreeVar{x} \HOLSymConst{/} (\HOLFreeVar{z} \HOLSymConst{\HOLTokenProd{}} \HOLFreeVar{y})) (\HOLFreeVar{x} \HOLSymConst{/} \HOLFreeVar{y} \HOLSymConst{/} \HOLFreeVar{z})
L_m:  \HOLTokenTurnstile{} \HOLConst{arrow} \HOLConst{L} \HOLFreeVar{x} \HOLFreeVar{x\sp{\prime}} \HOLSymConst{\HOLTokenConj{}} \HOLConst{arrow} \HOLConst{L} \HOLFreeVar{y} \HOLFreeVar{y\sp{\prime}} \HOLSymConst{\HOLTokenImp{}} \HOLConst{arrow} \HOLConst{L} (\HOLFreeVar{x} \HOLSymConst{\HOLTokenProd{}} \HOLFreeVar{y}) (\HOLFreeVar{x\sp{\prime}} \HOLSymConst{\HOLTokenProd{}} \HOLFreeVar{y\sp{\prime}})
L_n:  \HOLTokenTurnstile{} \HOLConst{arrow} \HOLConst{L} \HOLFreeVar{x} \HOLFreeVar{x\sp{\prime}} \HOLSymConst{\HOLTokenConj{}} \HOLConst{arrow} \HOLConst{L} \HOLFreeVar{y} \HOLFreeVar{y\sp{\prime}} \HOLSymConst{\HOLTokenImp{}} \HOLConst{arrow} \HOLConst{L} (\HOLFreeVar{x} \HOLSymConst{/} \HOLFreeVar{y\sp{\prime}}) (\HOLFreeVar{x\sp{\prime}} \HOLSymConst{/} \HOLFreeVar{y})
\end{alltt}

Finally, the monotone theorems specific to \textbf{L} are proved:
\begin{alltt}
L_dot_mono_r:
\HOLTokenTurnstile{} \HOLConst{arrow} \HOLConst{L} \HOLFreeVar{B} \HOLFreeVar{B\sp{\prime}} \HOLSymConst{\HOLTokenImp{}} \HOLConst{arrow} \HOLConst{L} (\HOLFreeVar{A} \HOLSymConst{\HOLTokenProd{}} \HOLFreeVar{B}) (\HOLFreeVar{A} \HOLSymConst{\HOLTokenProd{}} \HOLFreeVar{B\sp{\prime}})
L_dot_mono_l:
\HOLTokenTurnstile{} \HOLConst{arrow} \HOLConst{L} \HOLFreeVar{A} \HOLFreeVar{A\sp{\prime}} \HOLSymConst{\HOLTokenImp{}} \HOLConst{arrow} \HOLConst{L} (\HOLFreeVar{A} \HOLSymConst{\HOLTokenProd{}} \HOLFreeVar{B}) (\HOLFreeVar{A\sp{\prime}} \HOLSymConst{\HOLTokenProd{}} \HOLFreeVar{B})
L_slash_mono_l:
\HOLTokenTurnstile{} \HOLConst{arrow} \HOLConst{L} \HOLFreeVar{C} \HOLFreeVar{C\sp{\prime}} \HOLSymConst{\HOLTokenImp{}} \HOLConst{arrow} \HOLConst{L} (\HOLFreeVar{C} \HOLSymConst{/} \HOLFreeVar{B}) (\HOLFreeVar{C\sp{\prime}} \HOLSymConst{/} \HOLFreeVar{B})
L_slash_antimono_r:
\HOLTokenTurnstile{} \HOLConst{arrow} \HOLConst{L} \HOLFreeVar{B} \HOLFreeVar{B\sp{\prime}} \HOLSymConst{\HOLTokenImp{}} \HOLConst{arrow} \HOLConst{L} (\HOLFreeVar{C} \HOLSymConst{/} \HOLFreeVar{B\sp{\prime}}) (\HOLFreeVar{C} \HOLSymConst{/} \HOLFreeVar{B})
L_backslash_antimono_l:
\HOLTokenTurnstile{} \HOLConst{arrow} \HOLConst{L} \HOLFreeVar{A} \HOLFreeVar{A\sp{\prime}} \HOLSymConst{\HOLTokenImp{}} \HOLConst{arrow} \HOLConst{L} (\HOLFreeVar{A\sp{\prime}} \HOLSymConst{\HOLTokenBackslash} \HOLFreeVar{C}) (\HOLFreeVar{A} \HOLSymConst{\HOLTokenBackslash} \HOLFreeVar{C})
L_backslash_mono_r:
\HOLTokenTurnstile{} \HOLConst{arrow} \HOLConst{L} \HOLFreeVar{C} \HOLFreeVar{C\sp{\prime}} \HOLSymConst{\HOLTokenImp{}} \HOLConst{arrow} \HOLConst{L} (\HOLFreeVar{A} \HOLSymConst{\HOLTokenBackslash} \HOLFreeVar{C}) (\HOLFreeVar{A} \HOLSymConst{\HOLTokenBackslash} \HOLFreeVar{C\sp{\prime}})
\end{alltt}

The main problem of axiomatic syntactic calculus is that, for
complicated categories, it's not easy to ``find'' the
proofs. Automatic proof searching also doesn't work here, because the
searching space is infinite.  Further more, the arrow relation is not
a reduction, because there's no congruence here: if a sub-formula is
known to have another category, we can't just replace that sub-formula
and expect the whole formula still has the same category. Thus in
practice, we have to use other deduction systems which is more
convinence for manual proofs or automatic proofs. But for what ever
other deduction systems, to be useful they must be proved to be equivalent with
Syntactic calculus.

\subsection{Natural Deduction for Lambek Calculus}

Natural deduction was first invented by Dag Prawitz
\cite{Prawits:2001tc} as a non-semantic approach to derive
propositional logic formulae.\footnote{the semantic approach is to
build a truth table to explicitly check if the given formula is true
for all possible combinations of boolean values for each variables}
From a structure view, Natural deduction is nothing but a group of
inference rules, each concerns about the \emph{introduction} or
\emph{eliminatiob} of a logic connective.  Natural deduction is indeed
natural: it's successfully adopted in hand-proofs and also the primitive deduction systems
of many theorem provers.

Natural deduction has two styles, the original Prawitz style and the
Gentzen style (which is borrowed from Genzen's Sequent Calculus, we'll
talk about this in next section). The two styles are not totally
equivalent in expressive power, sometimes the Prawitz style is
unnatural to express certain logic extensions. Thus, in Lambek
Calculus, most of time, only the Gentzen
style of Natural Deduction is used, and when we're talking ``Natural
deduction'' in this paper, we always refer to its Gentzen style.

The basic unit in Natural Deduction (in Gentzen
style) is based on the concept of \emph{Sequent} which represents a
statement of the calculus in the following form:
\begin{equation*}
A_1, A_2, \ldots, A_n \vdash C
\end{equation*}
If we think the whole statement as a theorem, then $C$ is the
conclusion, and $A_1, A_2, \ldots, A_n$ is a list of hypotheses or assumptions in
which their orders are irrelevant. In terminalogy of sequents, such a
list is called \emph{the antecedent, or the hypotheses of the
  statement}. The purpose of inference rules is to derive new
statements from existing statements.

To adopt Natural Deduction for Lambek Calculus, especially the
non-associative version, two steps are done here:
\begin{enumerate}
\item The arrow relation between two categories,
  i.e. $A\longrightarrow B$, is now represented by logic theorem $A
  \Rightarrow B$, and then a sequent $A \vdash B$,
 which means by assuming $A$ we can prove $B$.
\item A sentence as a string of categories of each words, is now
  represented as the antecedents $A_1, A_2, \ldots, A_n$ of a
  sequent. However, now the order is essential (for Lambek calculus
  without \textbf{P}). And for Lambek calculus based on \textbf{NL},
  the binary bracketing is also necessary.
\end{enumerate}

To represent sentence structures as binary trees, a new data structure
called \emph{Term} is now defined in HOL4:
\begin{lstlisting}
Datatype `Term = OneForm ('a Form) | Comma Term Term`;
\end{lstlisting}
Notice the similarity between the \texttt{Comma} connective between
two Terms and the \texttt{Dot} connective between two Forms. In fact,
most of times they're convertable from each others, but it's
\texttt{Comma} representing the boundary of categories of words. To
translate a Term into Form, a recursive function called
\HOLinline{\HOLConst{deltaTranslation}} is defined as follows\footnote{The reverse
translation is not unique, and it's not needed to have such translations.}:
\begin{alltt}
\HOLTokenTurnstile{} (\HOLSymConst{\HOLTokenForall{}}\HOLBoundVar{f}. \HOLConst{deltaTranslation} (\HOLConst{OneForm} \HOLBoundVar{f}) \HOLSymConst{=} \HOLBoundVar{f}) \HOLSymConst{\HOLTokenConj{}}
   \HOLSymConst{\HOLTokenForall{}}\HOLBoundVar{t\sb{\mathrm{1}}} \HOLBoundVar{t\sb{\mathrm{2}}}.
     \HOLConst{deltaTranslation} (\HOLConst{Comma} \HOLBoundVar{t\sb{\mathrm{1}}} \HOLBoundVar{t\sb{\mathrm{2}}}) \HOLSymConst{=}
     \HOLConst{deltaTranslation} \HOLBoundVar{t\sb{\mathrm{1}}} \HOLSymConst{\HOLTokenProd{}} \HOLConst{deltaTranslation} \HOLBoundVar{t\sb{\mathrm{2}}}
\end{alltt}

Simiar to arrow extensions, which is a 2-ary relation between Terms,
now we have another kind of extensions, called Sequent extensions or
\HOLinline{\HOLFreeVar{gentzen\HOLTokenUnderscore{}extension}}, which is again abbreviated type as 2-ary
relation between Terms:
\begin{lstlisting}
type_abbrev ("gentzen_extension", ``:'a Term -> 'a Term -> bool``);
\end{lstlisting}

Sequent extensions for \textbf{L}, \textbf{NL}, \textbf{NLP} and
\textbf{LP} are defined as follows:
\begin{alltt}
\HOLTokenTurnstile{} \HOLConst{NL_Sequent} \HOLSymConst{=} \HOLSymConst{∅ᵣ}
\HOLTokenTurnstile{} \HOLConst{NLP_Sequent} (\HOLConst{Comma} \HOLFreeVar{A} \HOLFreeVar{B}) (\HOLConst{Comma} \HOLFreeVar{B} \HOLFreeVar{A})
\HOLTokenTurnstile{} (\HOLSymConst{\HOLTokenForall{}}\HOLBoundVar{A} \HOLBoundVar{B} \HOLBoundVar{C}.
      \HOLConst{L_Sequent} (\HOLConst{Comma} \HOLBoundVar{A} (\HOLConst{Comma} \HOLBoundVar{B} \HOLBoundVar{C})) (\HOLConst{Comma} (\HOLConst{Comma} \HOLBoundVar{A} \HOLBoundVar{B}) \HOLBoundVar{C})) \HOLSymConst{\HOLTokenConj{}}
   \HOLSymConst{\HOLTokenForall{}}\HOLBoundVar{A} \HOLBoundVar{B} \HOLBoundVar{C}.
     \HOLConst{L_Sequent} (\HOLConst{Comma} (\HOLConst{Comma} \HOLBoundVar{A} \HOLBoundVar{B}) \HOLBoundVar{C}) (\HOLConst{Comma} \HOLBoundVar{A} (\HOLConst{Comma} \HOLBoundVar{B} \HOLBoundVar{C}))
\HOLTokenTurnstile{} \HOLConst{LP_Sequent} \HOLSymConst{=} \HOLConst{add_extension} \HOLConst{NLP_Sequent} \HOLConst{L_Sequent}
\end{alltt}

To define the inference rules for Natural Deduction, we still need one
more device: the ability to \emph{replace} the sub-formula of a Term
into another Term:
\begin{alltt}
\HOLTokenTurnstile{} (\HOLSymConst{\HOLTokenForall{}}\HOLBoundVar{F\sb{\mathrm{1}}} \HOLBoundVar{F\sb{\mathrm{2}}}. \HOLConst{replace} \HOLBoundVar{F\sb{\mathrm{1}}} \HOLBoundVar{F\sb{\mathrm{2}}} \HOLBoundVar{F\sb{\mathrm{1}}} \HOLBoundVar{F\sb{\mathrm{2}}}) \HOLSymConst{\HOLTokenConj{}}
   (\HOLSymConst{\HOLTokenForall{}}\HOLBoundVar{Gamma\sb{\mathrm{1}}} \HOLBoundVar{Gamma\sb{\mathrm{2}}} \HOLBoundVar{Delta} \HOLBoundVar{F\sb{\mathrm{1}}} \HOLBoundVar{F\sb{\mathrm{2}}}.
      \HOLConst{replace} \HOLBoundVar{Gamma\sb{\mathrm{1}}} \HOLBoundVar{Gamma\sb{\mathrm{2}}} \HOLBoundVar{F\sb{\mathrm{1}}} \HOLBoundVar{F\sb{\mathrm{2}}} \HOLSymConst{\HOLTokenImp{}}
      \HOLConst{replace} (\HOLConst{Comma} \HOLBoundVar{Gamma\sb{\mathrm{1}}} \HOLBoundVar{Delta}) (\HOLConst{Comma} \HOLBoundVar{Gamma\sb{\mathrm{2}}} \HOLBoundVar{Delta}) \HOLBoundVar{F\sb{\mathrm{1}}}
        \HOLBoundVar{F\sb{\mathrm{2}}}) \HOLSymConst{\HOLTokenConj{}}
   \HOLSymConst{\HOLTokenForall{}}\HOLBoundVar{Gamma\sb{\mathrm{1}}} \HOLBoundVar{Gamma\sb{\mathrm{2}}} \HOLBoundVar{Delta} \HOLBoundVar{F\sb{\mathrm{1}}} \HOLBoundVar{F\sb{\mathrm{2}}}.
     \HOLConst{replace} \HOLBoundVar{Gamma\sb{\mathrm{1}}} \HOLBoundVar{Gamma\sb{\mathrm{2}}} \HOLBoundVar{F\sb{\mathrm{1}}} \HOLBoundVar{F\sb{\mathrm{2}}} \HOLSymConst{\HOLTokenImp{}}
     \HOLConst{replace} (\HOLConst{Comma} \HOLBoundVar{Delta} \HOLBoundVar{Gamma\sb{\mathrm{1}}}) (\HOLConst{Comma} \HOLBoundVar{Delta} \HOLBoundVar{Gamma\sb{\mathrm{2}}}) \HOLBoundVar{F\sb{\mathrm{1}}} \HOLBoundVar{F\sb{\mathrm{2}}}
\end{alltt}
Therefore when \HOLinline{\HOLConst{replace} \HOLFreeVar{A} \HOLFreeVar{B} \HOLFreeVar{C} \HOLFreeVar{D}} is true, it means in Term $A$, there's a
sub-formula $C$, and by replacing it with $D$, we obtain $B$. In many
papers, the Term $A$ is actually $A[C]$, and $B$ is written as
$A[D]$.

Related to \HOLinline{\HOLConst{replace}}, if the goal is to just replace some Commas
into Dots, there's a simplified relation called
\HOLinline{\HOLConst{replaceCommaDot}}. To define it, we first define its one-step version is called
\HOLinline{\HOLConst{replaceCommaDot1}}:
\begin{alltt}
\HOLTokenTurnstile{} \HOLConst{replace} \HOLFreeVar{T\sb{\mathrm{1}}} \HOLFreeVar{T\sb{\mathrm{2}}} (\HOLConst{Comma} (\HOLConst{OneForm} \HOLFreeVar{A}) (\HOLConst{OneForm} \HOLFreeVar{B}))
     (\HOLConst{OneForm} (\HOLFreeVar{A} \HOLSymConst{\HOLTokenProd{}} \HOLFreeVar{B})) \HOLSymConst{\HOLTokenImp{}}
   \HOLConst{replaceCommaDot1} \HOLFreeVar{T\sb{\mathrm{1}}} \HOLFreeVar{T\sb{\mathrm{2}}}
\end{alltt}
Then \HOLinline{\HOLConst{replaceCommaDot}} is nothing but the RTC of
\HOLinline{\HOLConst{replaceCommaDot1}}:
\begin{alltt}
\HOLTokenTurnstile{} \HOLConst{replaceCommaDot} \HOLSymConst{=} \HOLConst{replaceCommaDot1}\HOLSymConst{\HOLTokenSupStar{}}
\end{alltt}

To make these replace operators actually useful, we have proved many
theorems about them (some have complicated proofs):
\begin{alltt}
replaceTransitive':
\HOLTokenTurnstile{} \HOLConst{replaceCommaDot} \HOLFreeVar{T\sb{\mathrm{1}}} \HOLFreeVar{T\sb{\mathrm{2}}} \HOLSymConst{\HOLTokenConj{}} \HOLConst{replaceCommaDot} \HOLFreeVar{T\sb{\mathrm{2}}} \HOLFreeVar{T\sb{\mathrm{3}}} \HOLSymConst{\HOLTokenImp{}}
   \HOLConst{replaceCommaDot} \HOLFreeVar{T\sb{\mathrm{1}}} \HOLFreeVar{T\sb{\mathrm{3}}}
replaceCommaDot_rules:
\HOLTokenTurnstile{} (\HOLSymConst{\HOLTokenForall{}}\HOLBoundVar{T}. \HOLConst{replaceCommaDot} \HOLBoundVar{T} \HOLBoundVar{T}) \HOLSymConst{\HOLTokenConj{}}
   (\HOLSymConst{\HOLTokenForall{}}\HOLBoundVar{T\sb{\mathrm{1}}} \HOLBoundVar{T\sb{\mathrm{2}}} \HOLBoundVar{A} \HOLBoundVar{B}.
      \HOLConst{replace} \HOLBoundVar{T\sb{\mathrm{1}}} \HOLBoundVar{T\sb{\mathrm{2}}} (\HOLConst{Comma} (\HOLConst{OneForm} \HOLBoundVar{A}) (\HOLConst{OneForm} \HOLBoundVar{B}))
        (\HOLConst{OneForm} (\HOLBoundVar{A} \HOLSymConst{\HOLTokenProd{}} \HOLBoundVar{B})) \HOLSymConst{\HOLTokenImp{}}
      \HOLConst{replaceCommaDot} \HOLBoundVar{T\sb{\mathrm{1}}} \HOLBoundVar{T\sb{\mathrm{2}}}) \HOLSymConst{\HOLTokenConj{}}
   (\HOLSymConst{\HOLTokenForall{}}\HOLBoundVar{T\sb{\mathrm{1}}} \HOLBoundVar{T\sb{\mathrm{2}}} \HOLBoundVar{T\sb{\mathrm{3}}} \HOLBoundVar{A} \HOLBoundVar{B}.
      \HOLConst{replaceCommaDot} \HOLBoundVar{T\sb{\mathrm{1}}} \HOLBoundVar{T\sb{\mathrm{2}}} \HOLSymConst{\HOLTokenConj{}}
      \HOLConst{replace} \HOLBoundVar{T\sb{\mathrm{2}}} \HOLBoundVar{T\sb{\mathrm{3}}} (\HOLConst{Comma} (\HOLConst{OneForm} \HOLBoundVar{A}) (\HOLConst{OneForm} \HOLBoundVar{B}))
        (\HOLConst{OneForm} (\HOLBoundVar{A} \HOLSymConst{\HOLTokenProd{}} \HOLBoundVar{B})) \HOLSymConst{\HOLTokenImp{}}
      \HOLConst{replaceCommaDot} \HOLBoundVar{T\sb{\mathrm{1}}} \HOLBoundVar{T\sb{\mathrm{3}}}) \HOLSymConst{\HOLTokenConj{}}
   \HOLSymConst{\HOLTokenForall{}}\HOLBoundVar{T\sb{\mathrm{1}}} \HOLBoundVar{T\sb{\mathrm{2}}} \HOLBoundVar{T\sb{\mathrm{3}}} \HOLBoundVar{A} \HOLBoundVar{B}.
     \HOLConst{replace} \HOLBoundVar{T\sb{\mathrm{1}}} \HOLBoundVar{T\sb{\mathrm{2}}} (\HOLConst{Comma} (\HOLConst{OneForm} \HOLBoundVar{A}) (\HOLConst{OneForm} \HOLBoundVar{B}))
       (\HOLConst{OneForm} (\HOLBoundVar{A} \HOLSymConst{\HOLTokenProd{}} \HOLBoundVar{B})) \HOLSymConst{\HOLTokenConj{}} \HOLConst{replaceCommaDot} \HOLBoundVar{T\sb{\mathrm{2}}} \HOLBoundVar{T\sb{\mathrm{3}}} \HOLSymConst{\HOLTokenImp{}}
     \HOLConst{replaceCommaDot} \HOLBoundVar{T\sb{\mathrm{1}}} \HOLBoundVar{T\sb{\mathrm{3}}}
replaceMonoRight:
\HOLTokenTurnstile{} \HOLConst{replaceCommaDot} \HOLFreeVar{T\sb{\mathrm{1}}} \HOLFreeVar{T\sb{\mathrm{2}}} \HOLSymConst{\HOLTokenImp{}}
   \HOLConst{replaceCommaDot} (\HOLConst{Comma} \HOLFreeVar{T\sb{\mathrm{1}}} \HOLFreeVar{T\sb{\mathrm{3}}}) (\HOLConst{Comma} \HOLFreeVar{T\sb{\mathrm{2}}} \HOLFreeVar{T\sb{\mathrm{3}}})
replaceMonoLeft:
\HOLTokenTurnstile{} \HOLConst{replaceCommaDot} \HOLFreeVar{T\sb{\mathrm{1}}} \HOLFreeVar{T\sb{\mathrm{2}}} \HOLSymConst{\HOLTokenImp{}}
   \HOLConst{replaceCommaDot} (\HOLConst{Comma} \HOLFreeVar{T\sb{\mathrm{3}}} \HOLFreeVar{T\sb{\mathrm{1}}}) (\HOLConst{Comma} \HOLFreeVar{T\sb{\mathrm{3}}} \HOLFreeVar{T\sb{\mathrm{2}}})
replaceMono:
\HOLTokenTurnstile{} \HOLConst{replaceCommaDot} \HOLFreeVar{T\sb{\mathrm{1}}} \HOLFreeVar{T\sb{\mathrm{2}}} \HOLSymConst{\HOLTokenConj{}} \HOLConst{replaceCommaDot} \HOLFreeVar{T\sb{\mathrm{3}}} \HOLFreeVar{T\sb{\mathrm{4}}} \HOLSymConst{\HOLTokenImp{}}
   \HOLConst{replaceCommaDot} (\HOLConst{Comma} \HOLFreeVar{T\sb{\mathrm{1}}} \HOLFreeVar{T\sb{\mathrm{3}}}) (\HOLConst{Comma} \HOLFreeVar{T\sb{\mathrm{2}}} \HOLFreeVar{T\sb{\mathrm{4}}})
replaceTranslation:
\HOLTokenTurnstile{} \HOLConst{replaceCommaDot} \HOLFreeVar{T} (\HOLConst{OneForm} (\HOLConst{deltaTranslation} \HOLFreeVar{T}))
replace_inv1:
\HOLTokenTurnstile{} \HOLConst{replace} (\HOLConst{OneForm} \HOLFreeVar{C}) \HOLFreeVar{Gamma\sp{\prime}} (\HOLConst{OneForm} \HOLFreeVar{X}) \HOLFreeVar{Delta} \HOLSymConst{\HOLTokenImp{}}
   (\HOLFreeVar{Gamma\sp{\prime}} \HOLSymConst{=} \HOLFreeVar{Delta}) \HOLSymConst{\HOLTokenConj{}} (\HOLFreeVar{X} \HOLSymConst{=} \HOLFreeVar{C})
replace_inv2:
\HOLTokenTurnstile{} \HOLConst{replace} (\HOLConst{Comma} \HOLFreeVar{Gamma\sb{\mathrm{1}}} \HOLFreeVar{Gamma\sb{\mathrm{2}}}) \HOLFreeVar{Gamma\sp{\prime}} (\HOLConst{OneForm} \HOLFreeVar{X}) \HOLFreeVar{Delta} \HOLSymConst{\HOLTokenImp{}}
   (\HOLSymConst{\HOLTokenExists{}}\HOLBoundVar{G}.
      (\HOLFreeVar{Gamma\sp{\prime}} \HOLSymConst{=} \HOLConst{Comma} \HOLBoundVar{G} \HOLFreeVar{Gamma\sb{\mathrm{2}}}) \HOLSymConst{\HOLTokenConj{}}
      \HOLConst{replace} \HOLFreeVar{Gamma\sb{\mathrm{1}}} \HOLBoundVar{G} (\HOLConst{OneForm} \HOLFreeVar{X}) \HOLFreeVar{Delta}) \HOLSymConst{\HOLTokenDisj{}}
   \HOLSymConst{\HOLTokenExists{}}\HOLBoundVar{G}.
     (\HOLFreeVar{Gamma\sp{\prime}} \HOLSymConst{=} \HOLConst{Comma} \HOLFreeVar{Gamma\sb{\mathrm{1}}} \HOLBoundVar{G}) \HOLSymConst{\HOLTokenConj{}}
     \HOLConst{replace} \HOLFreeVar{Gamma\sb{\mathrm{2}}} \HOLBoundVar{G} (\HOLConst{OneForm} \HOLFreeVar{X}) \HOLFreeVar{Delta}
doubleReplace:
\HOLTokenTurnstile{} \HOLConst{replace} \HOLFreeVar{Gamma} \HOLFreeVar{Gamma\sp{\prime}} \HOLFreeVar{T\sb{\mathrm{1}}} \HOLFreeVar{T\sb{\mathrm{2}}} \HOLSymConst{\HOLTokenImp{}}
   \HOLSymConst{\HOLTokenForall{}}\HOLBoundVar{Gamma\sb{\mathrm{2}}} \HOLBoundVar{A} \HOLBoundVar{T\sb{\mathrm{3}}}.
     \HOLConst{replace} \HOLFreeVar{Gamma\sp{\prime}} \HOLBoundVar{Gamma\sb{\mathrm{2}}} (\HOLConst{OneForm} \HOLBoundVar{A}) \HOLBoundVar{T\sb{\mathrm{3}}} \HOLSymConst{\HOLTokenImp{}}
     (\HOLSymConst{\HOLTokenExists{}}\HOLBoundVar{G}.
        \HOLConst{replace} \HOLFreeVar{Gamma} \HOLBoundVar{G} (\HOLConst{OneForm} \HOLBoundVar{A}) \HOLBoundVar{T\sb{\mathrm{3}}} \HOLSymConst{\HOLTokenConj{}}
        \HOLConst{replace} \HOLBoundVar{G} \HOLBoundVar{Gamma\sb{\mathrm{2}}} \HOLFreeVar{T\sb{\mathrm{1}}} \HOLFreeVar{T\sb{\mathrm{2}}}) \HOLSymConst{\HOLTokenDisj{}}
     \HOLSymConst{\HOLTokenExists{}}\HOLBoundVar{G}.
       \HOLConst{replace} \HOLFreeVar{T\sb{\mathrm{2}}} \HOLBoundVar{G} (\HOLConst{OneForm} \HOLBoundVar{A}) \HOLBoundVar{T\sb{\mathrm{3}}} \HOLSymConst{\HOLTokenConj{}} \HOLConst{replace} \HOLFreeVar{Gamma} \HOLBoundVar{Gamma\sb{\mathrm{2}}} \HOLFreeVar{T\sb{\mathrm{1}}} \HOLBoundVar{G}
replaceSameP:
\HOLTokenTurnstile{} \HOLConst{replace} \HOLFreeVar{T\sb{\mathrm{1}}} \HOLFreeVar{T\sb{\mathrm{2}}} \HOLFreeVar{T\sb{\mathrm{3}}} \HOLFreeVar{T\sb{\mathrm{4}}} \HOLSymConst{\HOLTokenImp{}}
   \HOLSymConst{\HOLTokenForall{}}\HOLBoundVar{G}. \HOLSymConst{\HOLTokenExists{}}\HOLBoundVar{G\sp{\prime}}. \HOLConst{replace} \HOLFreeVar{T\sb{\mathrm{1}}} \HOLBoundVar{G\sp{\prime}} \HOLFreeVar{T\sb{\mathrm{3}}} \HOLBoundVar{G} \HOLSymConst{\HOLTokenConj{}} \HOLConst{replace} \HOLBoundVar{G\sp{\prime}} \HOLFreeVar{T\sb{\mathrm{2}}} \HOLBoundVar{G} \HOLFreeVar{T\sb{\mathrm{4}}}
replaceTrans:
\HOLTokenTurnstile{} \HOLConst{replace} \HOLFreeVar{T\sb{\mathrm{1}}} \HOLFreeVar{T\sb{\mathrm{2}}} \HOLFreeVar{T\sb{\mathrm{3}}} \HOLFreeVar{T\sb{\mathrm{4}}} \HOLSymConst{\HOLTokenImp{}}
   \HOLConst{replace} \HOLFreeVar{T\sb{\mathrm{3}}} \HOLFreeVar{T\sb{\mathrm{4}}} \HOLFreeVar{T\sb{\mathrm{5}}} \HOLFreeVar{T\sb{\mathrm{6}}} \HOLSymConst{\HOLTokenImp{}}
   \HOLConst{replace} \HOLFreeVar{T\sb{\mathrm{1}}} \HOLFreeVar{T\sb{\mathrm{2}}} \HOLFreeVar{T\sb{\mathrm{5}}} \HOLFreeVar{T\sb{\mathrm{6}}}
\end{alltt}

The replace theorems about three deduction systems are more difficult
to prove:
\begin{alltt}
replace_arrow:
\HOLTokenTurnstile{} \HOLConst{replace} \HOLFreeVar{Gamma} \HOLFreeVar{Gamma\sp{\prime}} \HOLFreeVar{T\sb{\mathrm{1}}} \HOLFreeVar{T\sb{\mathrm{2}}} \HOLSymConst{\HOLTokenImp{}}
   \HOLConst{arrow} \HOLFreeVar{X} (\HOLConst{deltaTranslation} \HOLFreeVar{T\sb{\mathrm{2}}}) (\HOLConst{deltaTranslation} \HOLFreeVar{T\sb{\mathrm{1}}}) \HOLSymConst{\HOLTokenImp{}}
   \HOLConst{arrow} \HOLFreeVar{X} (\HOLConst{deltaTranslation} \HOLFreeVar{Gamma\sp{\prime}}) (\HOLConst{deltaTranslation} \HOLFreeVar{Gamma})
replace_arrow':
\HOLTokenTurnstile{} \HOLConst{replace} \HOLFreeVar{Gamma} \HOLFreeVar{Gamma\sp{\prime}} \HOLFreeVar{T\sb{\mathrm{1}}} \HOLFreeVar{T\sb{\mathrm{2}}} \HOLSymConst{\HOLTokenImp{}}
   \HOLConst{arrow} \HOLFreeVar{X} (\HOLConst{deltaTranslation} \HOLFreeVar{T\sb{\mathrm{2}}}) (\HOLConst{deltaTranslation} \HOLFreeVar{T\sb{\mathrm{1}}}) \HOLSymConst{\HOLTokenImp{}}
   \HOLConst{arrow} \HOLFreeVar{X} (\HOLConst{deltaTranslation} \HOLFreeVar{Gamma}) \HOLFreeVar{C} \HOLSymConst{\HOLTokenImp{}}
   \HOLConst{arrow} \HOLFreeVar{X} (\HOLConst{deltaTranslation} \HOLFreeVar{Gamma\sp{\prime}}) \HOLFreeVar{C}
replaceGentzen:
\HOLTokenTurnstile{} \HOLConst{replace} \HOLFreeVar{Gamma} \HOLFreeVar{Gamma\sp{\prime}} \HOLFreeVar{Delta} \HOLFreeVar{Delta\sp{\prime}} \HOLSymConst{\HOLTokenImp{}}
   \HOLConst{gentzenSequent} \HOLFreeVar{E} \HOLFreeVar{Delta\sp{\prime}} (\HOLConst{deltaTranslation} \HOLFreeVar{Delta}) \HOLSymConst{\HOLTokenImp{}}
   \HOLConst{gentzenSequent} \HOLFreeVar{E} \HOLFreeVar{Gamma\sp{\prime}} (\HOLConst{deltaTranslation} \HOLFreeVar{Gamma})
replaceGentzen':
\HOLTokenTurnstile{} \HOLConst{replace} \HOLFreeVar{Gamma} \HOLFreeVar{Gamma\sp{\prime}} \HOLFreeVar{Delta} \HOLFreeVar{Delta\sp{\prime}} \HOLSymConst{\HOLTokenConj{}}
   \HOLConst{gentzenSequent} \HOLFreeVar{E} \HOLFreeVar{Delta\sp{\prime}} (\HOLConst{deltaTranslation} \HOLFreeVar{Delta}) \HOLSymConst{\HOLTokenConj{}}
   \HOLConst{gentzenSequent} \HOLFreeVar{E} \HOLFreeVar{Gamma} \HOLFreeVar{C} \HOLSymConst{\HOLTokenImp{}}
   \HOLConst{gentzenSequent} \HOLFreeVar{E} \HOLFreeVar{Gamma\sp{\prime}} \HOLFreeVar{C}
replaceNatDed:
\HOLTokenTurnstile{} \HOLConst{replace} \HOLFreeVar{Gamma} \HOLFreeVar{Gamma\sp{\prime}} \HOLFreeVar{Delta} \HOLFreeVar{Delta\sp{\prime}} \HOLSymConst{\HOLTokenImp{}}
   \HOLConst{natDed} \HOLFreeVar{E} \HOLFreeVar{Delta\sp{\prime}} (\HOLConst{deltaTranslation} \HOLFreeVar{Delta}) \HOLSymConst{\HOLTokenImp{}}
   \HOLConst{natDed} \HOLFreeVar{E} \HOLFreeVar{Gamma\sp{\prime}} (\HOLConst{deltaTranslation} \HOLFreeVar{Gamma})
\end{alltt}

Now we use \HOLinline{\HOLConst{natDed} \HOLFreeVar{E} \HOLFreeVar{Gamma} \HOLFreeVar{A}} to represent $Gamma \vdash A$
under extension $E$. The type of \HOLinline{\HOLConst{natDed}} is ``\HOLinline{\ensuremath{\alpha} \HOLTyOp{gentzen_extension} -> \ensuremath{\alpha} \HOLTyOp{Term} -> \ensuremath{\alpha} \HOLTyOp{Form} -> \HOLTyOp{bool}}''. And the
inference rules about Natural Deduction on Lambek Calculus are defined
by an inductive relation and then broken into the following separated rules:
\begin{alltt}
NatAxiom:
\HOLTokenTurnstile{} \HOLConst{natDed} \HOLFreeVar{E} (\HOLConst{OneForm} \HOLFreeVar{A}) \HOLFreeVar{A}
SlashIntro:
\HOLTokenTurnstile{} \HOLConst{natDed} \HOLFreeVar{E} (\HOLConst{Comma} \HOLFreeVar{Gamma} (\HOLConst{OneForm} \HOLFreeVar{B})) \HOLFreeVar{A} \HOLSymConst{\HOLTokenImp{}}
   \HOLConst{natDed} \HOLFreeVar{E} \HOLFreeVar{Gamma} (\HOLFreeVar{A} \HOLSymConst{/} \HOLFreeVar{B})
BackslashIntro:
\HOLTokenTurnstile{} \HOLConst{natDed} \HOLFreeVar{E} (\HOLConst{Comma} (\HOLConst{OneForm} \HOLFreeVar{B}) \HOLFreeVar{Gamma}) \HOLFreeVar{A} \HOLSymConst{\HOLTokenImp{}}
   \HOLConst{natDed} \HOLFreeVar{E} \HOLFreeVar{Gamma} (\HOLFreeVar{B} \HOLSymConst{\HOLTokenBackslash} \HOLFreeVar{A})
DotIntro:
\HOLTokenTurnstile{} \HOLConst{natDed} \HOLFreeVar{E} \HOLFreeVar{Gamma} \HOLFreeVar{A} \HOLSymConst{\HOLTokenConj{}} \HOLConst{natDed} \HOLFreeVar{E} \HOLFreeVar{Delta} \HOLFreeVar{B} \HOLSymConst{\HOLTokenImp{}}
   \HOLConst{natDed} \HOLFreeVar{E} (\HOLConst{Comma} \HOLFreeVar{Gamma} \HOLFreeVar{Delta}) (\HOLFreeVar{A} \HOLSymConst{\HOLTokenProd{}} \HOLFreeVar{B})
SlashElim:
\HOLTokenTurnstile{} \HOLConst{natDed} \HOLFreeVar{E} \HOLFreeVar{Gamma} (\HOLFreeVar{A} \HOLSymConst{/} \HOLFreeVar{B}) \HOLSymConst{\HOLTokenConj{}} \HOLConst{natDed} \HOLFreeVar{E} \HOLFreeVar{Delta} \HOLFreeVar{B} \HOLSymConst{\HOLTokenImp{}}
   \HOLConst{natDed} \HOLFreeVar{E} (\HOLConst{Comma} \HOLFreeVar{Gamma} \HOLFreeVar{Delta}) \HOLFreeVar{A}
BackslashElim:
\HOLTokenTurnstile{} \HOLConst{natDed} \HOLFreeVar{E} \HOLFreeVar{Gamma} \HOLFreeVar{B} \HOLSymConst{\HOLTokenConj{}} \HOLConst{natDed} \HOLFreeVar{E} \HOLFreeVar{Delta} (\HOLFreeVar{B} \HOLSymConst{\HOLTokenBackslash} \HOLFreeVar{A}) \HOLSymConst{\HOLTokenImp{}}
   \HOLConst{natDed} \HOLFreeVar{E} (\HOLConst{Comma} \HOLFreeVar{Gamma} \HOLFreeVar{Delta}) \HOLFreeVar{A}
DotElim:
\HOLTokenTurnstile{} \HOLConst{replace} \HOLFreeVar{Gamma} \HOLFreeVar{Gamma\sp{\prime}} (\HOLConst{Comma} (\HOLConst{OneForm} \HOLFreeVar{A}) (\HOLConst{OneForm} \HOLFreeVar{B})) \HOLFreeVar{Delta} \HOLSymConst{\HOLTokenConj{}}
   \HOLConst{natDed} \HOLFreeVar{E} \HOLFreeVar{Delta} (\HOLFreeVar{A} \HOLSymConst{\HOLTokenProd{}} \HOLFreeVar{B}) \HOLSymConst{\HOLTokenConj{}} \HOLConst{natDed} \HOLFreeVar{E} \HOLFreeVar{Gamma} \HOLFreeVar{C} \HOLSymConst{\HOLTokenImp{}}
   \HOLConst{natDed} \HOLFreeVar{E} \HOLFreeVar{Gamma\sp{\prime}} \HOLFreeVar{C}
NatExt:
\HOLTokenTurnstile{} \HOLConst{replace} \HOLFreeVar{Gamma} \HOLFreeVar{Gamma\sp{\prime}} \HOLFreeVar{Delta} \HOLFreeVar{Delta\sp{\prime}} \HOLSymConst{\HOLTokenConj{}} \HOLFreeVar{E} \HOLFreeVar{Delta} \HOLFreeVar{Delta\sp{\prime}} \HOLSymConst{\HOLTokenConj{}}
   \HOLConst{natDed} \HOLFreeVar{E} \HOLFreeVar{Gamma} \HOLFreeVar{C} \HOLSymConst{\HOLTokenImp{}}
   \HOLConst{natDed} \HOLFreeVar{E} \HOLFreeVar{Gamma\sp{\prime}} \HOLFreeVar{C}
\end{alltt}
These natural deduction rules are quite similar with those arrow rules
of Syntactic Calculus. And the elimination rules \texttt{SlashElim}
and \texttt{BackslashElim} are even more close to AB grammar
rules. Besides, \texttt{replace} operations only appear in
\texttt{DotElim} and \texttt{NatExt} rules. All these characteristics
made natural deduction easier to use for proving facts about Lambek
Calculus.

We have proved a few theorems as simplified version of above primitive
rules, sometimes it's easier to use them instead of the primitive rules:
\begin{alltt}
NatAxiomGeneralized:
\HOLTokenTurnstile{} \HOLConst{natDed} \HOLFreeVar{E} \HOLFreeVar{Gamma} (\HOLConst{deltaTranslation} \HOLFreeVar{Gamma})
DotElimGeneralized:
\HOLTokenTurnstile{} \HOLConst{replaceCommaDot} \HOLFreeVar{Gamma} \HOLFreeVar{Gamma\sp{\prime}} \HOLSymConst{\HOLTokenImp{}}
   \HOLConst{natDed} \HOLFreeVar{E} \HOLFreeVar{Gamma} \HOLFreeVar{C} \HOLSymConst{\HOLTokenImp{}}
   \HOLConst{natDed} \HOLFreeVar{E} \HOLFreeVar{Gamma\sp{\prime}} \HOLFreeVar{C}
NatTermToForm:
\HOLTokenTurnstile{} \HOLConst{natDed} \HOLFreeVar{E} \HOLFreeVar{Gamma} \HOLFreeVar{C} \HOLSymConst{\HOLTokenImp{}}
   \HOLConst{natDed} \HOLFreeVar{E} (\HOLConst{OneForm} (\HOLConst{deltaTranslation} \HOLFreeVar{Gamma})) \HOLFreeVar{C}
NatExtSimpl:
\HOLTokenTurnstile{} \HOLFreeVar{E} \HOLFreeVar{Gamma} \HOLFreeVar{Gamma\sp{\prime}} \HOLSymConst{\HOLTokenConj{}} \HOLConst{natDed} \HOLFreeVar{E} \HOLFreeVar{Gamma} \HOLFreeVar{C} \HOLSymConst{\HOLTokenImp{}} \HOLConst{natDed} \HOLFreeVar{E} \HOLFreeVar{Gamma\sp{\prime}} \HOLFreeVar{C}
\end{alltt}

The main problem for Natural Deduction is the lacking of decision
procedures. When proving any theorem in backward way, one has to
``guess'' many unknown variables to correctly apply those elimination
rules. To see this, let's take a look at one elimination rules,
e. g. \texttt{SlashElim}:
\begin{alltt}
\HOLTokenTurnstile{} \HOLConst{natDed} \HOLFreeVar{E} \HOLFreeVar{Gamma} (\HOLFreeVar{A} \HOLSymConst{/} \HOLFreeVar{B}) \HOLSymConst{\HOLTokenConj{}} \HOLConst{natDed} \HOLFreeVar{E} \HOLFreeVar{Delta} \HOLFreeVar{B} \HOLSymConst{\HOLTokenImp{}}
   \HOLConst{natDed} \HOLFreeVar{E} (\HOLConst{Comma} \HOLFreeVar{Gamma} \HOLFreeVar{Delta}) \HOLFreeVar{A}
\end{alltt}
In this rule, variable $B$ only appears in the antecedents but the
conclusion, and when applying this rule from backwards, we have to
guess out the value of $B$, and different values of $B$ may completely
change the rest proofs (only the correct guess can lead to a success
proof). Thus, automatic proof searching based on these Natural
deduction rules are impossible. But if our final goal is to get a
langauge parser, we have to have a more reasonable set of rules in
which there's essential no guess (this is the so-called ``sub-formula
property'' in sequent calculus, we'll get to this property later).

Thus, to make both manual proving and automatic proof searching
possible, now we represent the final solution for the inference
system of Lambek Calculus: the Gentzen's Sequent Calculus.

\subsection{Gentzen's Sequent Calculus for Lambek Calculus}

Sequent Calculus was original introduced by Gerhard Gentzen in two German
papers \cite{Gentzen:1935gj} \cite{Gentzen:1935dm} written in
1935. It's a great achievements in the proof theory of classical and intuitionistic
logic. To understnad its difference with Natural Deduction, it's
better to watch directly the inference rules for Lambek Calculus. In
our implementation, the Sequent Calculus is defined through an
inductive definition of the relation \HOLinline{\HOLConst{gentzenSequent}} which has
the same type as previous \HOLinline{\HOLConst{natDed}} relation. Here are its
inference rules:
\begin{alltt}
SeqAxiom:
\HOLTokenTurnstile{} \HOLConst{gentzenSequent} \HOLFreeVar{E} (\HOLConst{OneForm} \HOLFreeVar{A}) \HOLFreeVar{A}
RightSlash:
\HOLTokenTurnstile{} \HOLConst{gentzenSequent} \HOLFreeVar{E} (\HOLConst{Comma} \HOLFreeVar{Gamma} (\HOLConst{OneForm} \HOLFreeVar{B})) \HOLFreeVar{A} \HOLSymConst{\HOLTokenImp{}}
   \HOLConst{gentzenSequent} \HOLFreeVar{E} \HOLFreeVar{Gamma} (\HOLFreeVar{A} \HOLSymConst{/} \HOLFreeVar{B})
RightBackslash:
\HOLTokenTurnstile{} \HOLConst{gentzenSequent} \HOLFreeVar{E} (\HOLConst{Comma} (\HOLConst{OneForm} \HOLFreeVar{B}) \HOLFreeVar{Gamma}) \HOLFreeVar{A} \HOLSymConst{\HOLTokenImp{}}
   \HOLConst{gentzenSequent} \HOLFreeVar{E} \HOLFreeVar{Gamma} (\HOLFreeVar{B} \HOLSymConst{\HOLTokenBackslash} \HOLFreeVar{A})
RightDot:
\HOLTokenTurnstile{} \HOLConst{gentzenSequent} \HOLFreeVar{E} \HOLFreeVar{Gamma} \HOLFreeVar{A} \HOLSymConst{\HOLTokenConj{}} \HOLConst{gentzenSequent} \HOLFreeVar{E} \HOLFreeVar{Delta} \HOLFreeVar{B} \HOLSymConst{\HOLTokenImp{}}
   \HOLConst{gentzenSequent} \HOLFreeVar{E} (\HOLConst{Comma} \HOLFreeVar{Gamma} \HOLFreeVar{Delta}) (\HOLFreeVar{A} \HOLSymConst{\HOLTokenProd{}} \HOLFreeVar{B})
LeftSlash:
\HOLTokenTurnstile{} \HOLConst{replace} \HOLFreeVar{Gamma} \HOLFreeVar{Gamma\sp{\prime}} (\HOLConst{OneForm} \HOLFreeVar{A})
     (\HOLConst{Comma} (\HOLConst{OneForm} (\HOLFreeVar{A} \HOLSymConst{/} \HOLFreeVar{B})) \HOLFreeVar{Delta}) \HOLSymConst{\HOLTokenConj{}}
   \HOLConst{gentzenSequent} \HOLFreeVar{E} \HOLFreeVar{Delta} \HOLFreeVar{B} \HOLSymConst{\HOLTokenConj{}} \HOLConst{gentzenSequent} \HOLFreeVar{E} \HOLFreeVar{Gamma} \HOLFreeVar{C} \HOLSymConst{\HOLTokenImp{}}
   \HOLConst{gentzenSequent} \HOLFreeVar{E} \HOLFreeVar{Gamma\sp{\prime}} \HOLFreeVar{C}
LeftBackslash:
\HOLTokenTurnstile{} \HOLConst{replace} \HOLFreeVar{Gamma} \HOLFreeVar{Gamma\sp{\prime}} (\HOLConst{OneForm} \HOLFreeVar{A})
     (\HOLConst{Comma} \HOLFreeVar{Delta} (\HOLConst{OneForm} (\HOLFreeVar{B} \HOLSymConst{\HOLTokenBackslash} \HOLFreeVar{A}))) \HOLSymConst{\HOLTokenConj{}}
   \HOLConst{gentzenSequent} \HOLFreeVar{E} \HOLFreeVar{Delta} \HOLFreeVar{B} \HOLSymConst{\HOLTokenConj{}} \HOLConst{gentzenSequent} \HOLFreeVar{E} \HOLFreeVar{Gamma} \HOLFreeVar{C} \HOLSymConst{\HOLTokenImp{}}
   \HOLConst{gentzenSequent} \HOLFreeVar{E} \HOLFreeVar{Gamma\sp{\prime}} \HOLFreeVar{C}
LeftDot:
\HOLTokenTurnstile{} \HOLConst{replace} \HOLFreeVar{Gamma} \HOLFreeVar{Gamma\sp{\prime}} (\HOLConst{Comma} (\HOLConst{OneForm} \HOLFreeVar{A}) (\HOLConst{OneForm} \HOLFreeVar{B}))
     (\HOLConst{OneForm} (\HOLFreeVar{A} \HOLSymConst{\HOLTokenProd{}} \HOLFreeVar{B})) \HOLSymConst{\HOLTokenConj{}} \HOLConst{gentzenSequent} \HOLFreeVar{E} \HOLFreeVar{Gamma} \HOLFreeVar{C} \HOLSymConst{\HOLTokenImp{}}
   \HOLConst{gentzenSequent} \HOLFreeVar{E} \HOLFreeVar{Gamma\sp{\prime}} \HOLFreeVar{C}
CutRule:
\HOLTokenTurnstile{} \HOLConst{replace} \HOLFreeVar{Gamma} \HOLFreeVar{Gamma\sp{\prime}} (\HOLConst{OneForm} \HOLFreeVar{A}) \HOLFreeVar{Delta} \HOLSymConst{\HOLTokenConj{}}
   \HOLConst{gentzenSequent} \HOLFreeVar{E} \HOLFreeVar{Delta} \HOLFreeVar{A} \HOLSymConst{\HOLTokenConj{}} \HOLConst{gentzenSequent} \HOLFreeVar{E} \HOLFreeVar{Gamma} \HOLFreeVar{C} \HOLSymConst{\HOLTokenImp{}}
   \HOLConst{gentzenSequent} \HOLFreeVar{E} \HOLFreeVar{Gamma\sp{\prime}} \HOLFreeVar{C}
SeqExt:
\HOLTokenTurnstile{} \HOLConst{replace} \HOLFreeVar{Gamma} \HOLFreeVar{Gamma\sp{\prime}} \HOLFreeVar{Delta} \HOLFreeVar{Delta\sp{\prime}} \HOLSymConst{\HOLTokenConj{}} \HOLFreeVar{E} \HOLFreeVar{Delta} \HOLFreeVar{Delta\sp{\prime}} \HOLSymConst{\HOLTokenConj{}}
   \HOLConst{gentzenSequent} \HOLFreeVar{E} \HOLFreeVar{Gamma} \HOLFreeVar{C} \HOLSymConst{\HOLTokenImp{}}
   \HOLConst{gentzenSequent} \HOLFreeVar{E} \HOLFreeVar{Gamma\sp{\prime}} \HOLFreeVar{C}
\end{alltt}

Comparing with Natural Deduction rules, the following differences are
notable:
\begin{enumerate}
\item For each of the three connectives ($\cdot$, $/$ and
  $\setminus$), there're Left and Right rules for them;  For Natural
  deduction, instead they're Introduction and Elimination
  rules.
\item Except for the different relation name, the rule \texttt{SeqAxiom} is the same
  as \texttt{NatAxiom}, \texttt{RightSlash} is the same as
  \texttt{SlashIntro}, \texttt{RightBackslash} is the same as
  \texttt{BackslashIntro}, \texttt{RightDot} is the same as
  \texttt{DotIntro}, and finally the rule \texttt{SeqExt} is the same
  as \texttt{NatExt}.
\item All Left rules are new, and they all make use of the
  \HOLinline{\HOLConst{replace}} predicates. And the result of these rules is to
  introduce one of the three connectives in the antecedents
  \emph{without changing the conclusion}.
\item The rule \texttt{CutRule} is new in Sequent Calculus, it's not
  present in Natural Deduction, nor can be proved in Natural
  Deduction.
\end{enumerate}

Beside above observations, there's one more important fact:
\begin{itemize}
\item All rules except for \texttt{CutRule} satisfy the so-called
  ``sub-formula property'', that is, all variables in the antecedents
  are sub-formulae of the conclusion.
\end{itemize}
To see why this is true, let's take a deeper look at \texttt{LeftSlash}:
\begin{alltt}
\HOLTokenTurnstile{} \HOLConst{replace} \HOLFreeVar{Gamma} \HOLFreeVar{Gamma\sp{\prime}} (\HOLConst{OneForm} \HOLFreeVar{A})
     (\HOLConst{Comma} (\HOLConst{OneForm} (\HOLFreeVar{A} \HOLSymConst{/} \HOLFreeVar{B})) \HOLFreeVar{Delta}) \HOLSymConst{\HOLTokenConj{}}
   \HOLConst{gentzenSequent} \HOLFreeVar{E} \HOLFreeVar{Delta} \HOLFreeVar{B} \HOLSymConst{\HOLTokenConj{}} \HOLConst{gentzenSequent} \HOLFreeVar{E} \HOLFreeVar{Gamma} \HOLFreeVar{C} \HOLSymConst{\HOLTokenImp{}}
   \HOLConst{gentzenSequent} \HOLFreeVar{E} \HOLFreeVar{Gamma\sp{\prime}} \HOLFreeVar{C}
\end{alltt}
If we consider the meaning of \HOLinline{\HOLConst{replace}}, we may rewrite this
rule into the following form:
\begin{prooftree}
\AxiomC{$Delta \vdash B$}
\AxiomC{$Gamma[A] \vdash C$}
\LeftLabel{(LeftSlash)}
\BinaryInfC{$Gamma[(A/B, Delta)] \vdash C$}
\end{prooftree}
To some extents, this rule is just saying $A/B \cdot B \rightarrow A$
as in AB grammar. What's more important is the fact that, all
variables appearing in the conclusion, i.e. $Gamma$, $A$, $B$, $Delta$
and $C$, are variables borrowing from antecedents. The same fact is
true for all other \HOLinline{\HOLConst{gentzenSequent}} rules, except for
\texttt{CutRule}:
\begin{prooftree}
\AxiomC{$Delta \vdash A$}
\AxiomC{$Gamma[A] \vdash C$}
\LeftLabel{(CutRule)}
\BinaryInfC{$Gamma[Delta] \vdash C$}
\end{prooftree}
in which the variable $A$ appears only in antecedents but conclusion.

The sub-formula property is extremely useful for automatic proof
searching, because there's nothing to guess when applying these rules
from backwards.

Based on above primitive inference rules, we have proved a large
amount of derived theorems which is true for all Lambek calculus
(i. e. extended from \textbf{NL}):
\begin{alltt}
SeqAxiomGeneralized:
\HOLTokenTurnstile{} \HOLConst{gentzenSequent} \HOLFreeVar{E} \HOLFreeVar{Gamma} (\HOLConst{deltaTranslation} \HOLFreeVar{Gamma})
LeftDotSimpl:
\HOLTokenTurnstile{} \HOLConst{gentzenSequent} \HOLFreeVar{E} (\HOLConst{Comma} (\HOLConst{OneForm} \HOLFreeVar{A}) (\HOLConst{OneForm} \HOLFreeVar{B})) \HOLFreeVar{C} \HOLSymConst{\HOLTokenImp{}}
   \HOLConst{gentzenSequent} \HOLFreeVar{E} (\HOLConst{OneForm} (\HOLFreeVar{A} \HOLSymConst{\HOLTokenProd{}} \HOLFreeVar{B})) \HOLFreeVar{C}
LeftDotGeneralized:
\HOLTokenTurnstile{} \HOLConst{replaceCommaDot} \HOLFreeVar{T\sb{\mathrm{1}}} \HOLFreeVar{T\sb{\mathrm{2}}} \HOLSymConst{\HOLTokenImp{}}
   \HOLConst{gentzenSequent} \HOLFreeVar{E} \HOLFreeVar{T\sb{\mathrm{1}}} \HOLFreeVar{C} \HOLSymConst{\HOLTokenImp{}}
   \HOLConst{gentzenSequent} \HOLFreeVar{E} \HOLFreeVar{T\sb{\mathrm{2}}} \HOLFreeVar{C}
SeqTermToForm:
\HOLTokenTurnstile{} \HOLConst{gentzenSequent} \HOLFreeVar{E} \HOLFreeVar{Gamma} \HOLFreeVar{C} \HOLSymConst{\HOLTokenImp{}}
   \HOLConst{gentzenSequent} \HOLFreeVar{E} (\HOLConst{OneForm} (\HOLConst{deltaTranslation} \HOLFreeVar{Gamma})) \HOLFreeVar{C}
LeftSlashSimpl:
\HOLTokenTurnstile{} \HOLConst{gentzenSequent} \HOLFreeVar{E} \HOLFreeVar{Gamma} \HOLFreeVar{B} \HOLSymConst{\HOLTokenConj{}} \HOLConst{gentzenSequent} \HOLFreeVar{E} (\HOLConst{OneForm} \HOLFreeVar{A}) \HOLFreeVar{C} \HOLSymConst{\HOLTokenImp{}}
   \HOLConst{gentzenSequent} \HOLFreeVar{E} (\HOLConst{Comma} (\HOLConst{OneForm} (\HOLFreeVar{A} \HOLSymConst{/} \HOLFreeVar{B})) \HOLFreeVar{Gamma}) \HOLFreeVar{C}
LeftBackslashSimpl:
\HOLTokenTurnstile{} \HOLConst{gentzenSequent} \HOLFreeVar{E} \HOLFreeVar{Gamma} \HOLFreeVar{B} \HOLSymConst{\HOLTokenConj{}} \HOLConst{gentzenSequent} \HOLFreeVar{E} (\HOLConst{OneForm} \HOLFreeVar{A}) \HOLFreeVar{C} \HOLSymConst{\HOLTokenImp{}}
   \HOLConst{gentzenSequent} \HOLFreeVar{E} (\HOLConst{Comma} \HOLFreeVar{Gamma} (\HOLConst{OneForm} (\HOLFreeVar{B} \HOLSymConst{\HOLTokenBackslash} \HOLFreeVar{A}))) \HOLFreeVar{C}
CutRuleSimpl:
\HOLTokenTurnstile{} \HOLConst{gentzenSequent} \HOLFreeVar{E} \HOLFreeVar{Gamma} \HOLFreeVar{A} \HOLSymConst{\HOLTokenConj{}} \HOLConst{gentzenSequent} \HOLFreeVar{E} (\HOLConst{OneForm} \HOLFreeVar{A}) \HOLFreeVar{C} \HOLSymConst{\HOLTokenImp{}}
   \HOLConst{gentzenSequent} \HOLFreeVar{E} \HOLFreeVar{Gamma} \HOLFreeVar{C}
DotRightSlash':
\HOLTokenTurnstile{} \HOLConst{gentzenSequent} \HOLFreeVar{E} (\HOLConst{OneForm} \HOLFreeVar{A}) (\HOLFreeVar{C} \HOLSymConst{/} \HOLFreeVar{B}) \HOLSymConst{\HOLTokenImp{}}
   \HOLConst{gentzenSequent} \HOLFreeVar{E} (\HOLConst{OneForm} (\HOLFreeVar{A} \HOLSymConst{\HOLTokenProd{}} \HOLFreeVar{B})) \HOLFreeVar{C}
DotRightBackslash':
\HOLTokenTurnstile{} \HOLConst{gentzenSequent} \HOLFreeVar{E} (\HOLConst{OneForm} \HOLFreeVar{B}) (\HOLFreeVar{A} \HOLSymConst{\HOLTokenBackslash} \HOLFreeVar{C}) \HOLSymConst{\HOLTokenImp{}}
   \HOLConst{gentzenSequent} \HOLFreeVar{E} (\HOLConst{OneForm} (\HOLFreeVar{A} \HOLSymConst{\HOLTokenProd{}} \HOLFreeVar{B})) \HOLFreeVar{C}
SeqExtSimpl:
\HOLTokenTurnstile{} \HOLFreeVar{E} \HOLFreeVar{Gamma} \HOLFreeVar{Gamma\sp{\prime}} \HOLSymConst{\HOLTokenConj{}} \HOLConst{gentzenSequent} \HOLFreeVar{E} \HOLFreeVar{Gamma} \HOLFreeVar{C} \HOLSymConst{\HOLTokenImp{}}
   \HOLConst{gentzenSequent} \HOLFreeVar{E} \HOLFreeVar{Gamma\sp{\prime}} \HOLFreeVar{C}
application:
\HOLTokenTurnstile{} \HOLConst{gentzenSequent} \HOLFreeVar{E} (\HOLConst{OneForm} (\HOLFreeVar{A} \HOLSymConst{/} \HOLFreeVar{B} \HOLSymConst{\HOLTokenProd{}} \HOLFreeVar{B})) \HOLFreeVar{A}
application':
\HOLTokenTurnstile{} \HOLConst{gentzenSequent} \HOLFreeVar{E} (\HOLConst{OneForm} (\HOLFreeVar{B} \HOLSymConst{\HOLTokenProd{}} \HOLFreeVar{B} \HOLSymConst{\HOLTokenBackslash} \HOLFreeVar{A})) \HOLFreeVar{A}
RightSlashDot:
\HOLTokenTurnstile{} \HOLConst{gentzenSequent} \HOLFreeVar{E} (\HOLConst{OneForm} (\HOLFreeVar{A} \HOLSymConst{\HOLTokenProd{}} \HOLFreeVar{C})) \HOLFreeVar{B} \HOLSymConst{\HOLTokenImp{}}
   \HOLConst{gentzenSequent} \HOLFreeVar{E} (\HOLConst{OneForm} \HOLFreeVar{A}) (\HOLFreeVar{B} \HOLSymConst{/} \HOLFreeVar{C})
RightBackslashDot:
\HOLTokenTurnstile{} \HOLConst{gentzenSequent} \HOLFreeVar{E} (\HOLConst{OneForm} (\HOLFreeVar{B} \HOLSymConst{\HOLTokenProd{}} \HOLFreeVar{A})) \HOLFreeVar{C} \HOLSymConst{\HOLTokenImp{}}
   \HOLConst{gentzenSequent} \HOLFreeVar{E} (\HOLConst{OneForm} \HOLFreeVar{A}) (\HOLFreeVar{B} \HOLSymConst{\HOLTokenBackslash} \HOLFreeVar{C})
coApplication:
\HOLTokenTurnstile{} \HOLConst{gentzenSequent} \HOLFreeVar{E} (\HOLConst{OneForm} \HOLFreeVar{A}) (\HOLFreeVar{A} \HOLSymConst{\HOLTokenProd{}} \HOLFreeVar{B} \HOLSymConst{/} \HOLFreeVar{B})
coApplication':
\HOLTokenTurnstile{} \HOLConst{gentzenSequent} \HOLFreeVar{E} (\HOLConst{OneForm} \HOLFreeVar{A}) (\HOLFreeVar{B} \HOLSymConst{\HOLTokenBackslash} (\HOLFreeVar{B} \HOLSymConst{\HOLTokenProd{}} \HOLFreeVar{A}))
mono_E:
\HOLTokenTurnstile{} \HOLConst{gentzenSequent} \HOLFreeVar{E} \HOLFreeVar{Gamma} \HOLFreeVar{A} \HOLSymConst{\HOLTokenImp{}}
   \HOLConst{extends} \HOLFreeVar{E} \HOLFreeVar{E\sp{\prime}} \HOLSymConst{\HOLTokenImp{}}
   \HOLConst{gentzenSequent} \HOLFreeVar{E\sp{\prime}} \HOLFreeVar{Gamma} \HOLFreeVar{A}
monotonicity:
\HOLTokenTurnstile{} \HOLConst{gentzenSequent} \HOLFreeVar{E} (\HOLConst{OneForm} \HOLFreeVar{A}) \HOLFreeVar{B} \HOLSymConst{\HOLTokenConj{}}
   \HOLConst{gentzenSequent} \HOLFreeVar{E} (\HOLConst{OneForm} \HOLFreeVar{C}) \HOLFreeVar{D} \HOLSymConst{\HOLTokenImp{}}
   \HOLConst{gentzenSequent} \HOLFreeVar{E} (\HOLConst{OneForm} (\HOLFreeVar{A} \HOLSymConst{\HOLTokenProd{}} \HOLFreeVar{C})) (\HOLFreeVar{B} \HOLSymConst{\HOLTokenProd{}} \HOLFreeVar{D})
isotonicity:
\HOLTokenTurnstile{} \HOLConst{gentzenSequent} \HOLFreeVar{E} (\HOLConst{OneForm} \HOLFreeVar{A}) \HOLFreeVar{B} \HOLSymConst{\HOLTokenImp{}}
   \HOLConst{gentzenSequent} \HOLFreeVar{E} (\HOLConst{OneForm} (\HOLFreeVar{A} \HOLSymConst{/} \HOLFreeVar{C})) (\HOLFreeVar{B} \HOLSymConst{/} \HOLFreeVar{C})
isotonicity':
\HOLTokenTurnstile{} \HOLConst{gentzenSequent} \HOLFreeVar{E} (\HOLConst{OneForm} \HOLFreeVar{A}) \HOLFreeVar{B} \HOLSymConst{\HOLTokenImp{}}
   \HOLConst{gentzenSequent} \HOLFreeVar{E} (\HOLConst{OneForm} (\HOLFreeVar{C} \HOLSymConst{\HOLTokenBackslash} \HOLFreeVar{A})) (\HOLFreeVar{C} \HOLSymConst{\HOLTokenBackslash} \HOLFreeVar{B})
antitonicity:
\HOLTokenTurnstile{} \HOLConst{gentzenSequent} \HOLFreeVar{E} (\HOLConst{OneForm} \HOLFreeVar{A}) \HOLFreeVar{B} \HOLSymConst{\HOLTokenImp{}}
   \HOLConst{gentzenSequent} \HOLFreeVar{E} (\HOLConst{OneForm} (\HOLFreeVar{C} \HOLSymConst{/} \HOLFreeVar{B})) (\HOLFreeVar{C} \HOLSymConst{/} \HOLFreeVar{A})
antitonicity':
\HOLTokenTurnstile{} \HOLConst{gentzenSequent} \HOLFreeVar{E} (\HOLConst{OneForm} \HOLFreeVar{A}) \HOLFreeVar{B} \HOLSymConst{\HOLTokenImp{}}
   \HOLConst{gentzenSequent} \HOLFreeVar{E} (\HOLConst{OneForm} (\HOLFreeVar{B} \HOLSymConst{\HOLTokenBackslash} \HOLFreeVar{C})) (\HOLFreeVar{A} \HOLSymConst{\HOLTokenBackslash} \HOLFreeVar{C})
lifting:
\HOLTokenTurnstile{} \HOLConst{gentzenSequent} \HOLFreeVar{E} (\HOLConst{OneForm} \HOLFreeVar{A}) (\HOLFreeVar{B} \HOLSymConst{/} \HOLFreeVar{A} \HOLSymConst{\HOLTokenBackslash} \HOLFreeVar{B})
lifting':
\HOLTokenTurnstile{} \HOLConst{gentzenSequent} \HOLFreeVar{E} (\HOLConst{OneForm} \HOLFreeVar{A}) ((\HOLFreeVar{B} \HOLSymConst{/} \HOLFreeVar{A}) \HOLSymConst{\HOLTokenBackslash} \HOLFreeVar{B})
\end{alltt}

For Lambek Calculus extended from \textbf{L},which supports associativity, we have proved some extra
theorems:
\begin{alltt}
LextensionSimpl:
\HOLTokenTurnstile{} \HOLConst{extends} \HOLConst{L_Sequent} \HOLFreeVar{E} \HOLSymConst{\HOLTokenConj{}}
   \HOLConst{gentzenSequent} \HOLFreeVar{E} (\HOLConst{Comma} \HOLFreeVar{T\sb{\mathrm{1}}} (\HOLConst{Comma} \HOLFreeVar{T\sb{\mathrm{2}}} \HOLFreeVar{T\sb{\mathrm{3}}})) \HOLFreeVar{C} \HOLSymConst{\HOLTokenImp{}}
   \HOLConst{gentzenSequent} \HOLFreeVar{E} (\HOLConst{Comma} (\HOLConst{Comma} \HOLFreeVar{T\sb{\mathrm{1}}} \HOLFreeVar{T\sb{\mathrm{2}}}) \HOLFreeVar{T\sb{\mathrm{3}}}) \HOLFreeVar{C}
LextensionSimpl':
\HOLTokenTurnstile{} \HOLConst{extends} \HOLConst{L_Sequent} \HOLFreeVar{E} \HOLSymConst{\HOLTokenConj{}}
   \HOLConst{gentzenSequent} \HOLFreeVar{E} (\HOLConst{Comma} (\HOLConst{Comma} \HOLFreeVar{T\sb{\mathrm{1}}} \HOLFreeVar{T\sb{\mathrm{2}}}) \HOLFreeVar{T\sb{\mathrm{3}}}) \HOLFreeVar{C} \HOLSymConst{\HOLTokenImp{}}
   \HOLConst{gentzenSequent} \HOLFreeVar{E} (\HOLConst{Comma} \HOLFreeVar{T\sb{\mathrm{1}}} (\HOLConst{Comma} \HOLFreeVar{T\sb{\mathrm{2}}} \HOLFreeVar{T\sb{\mathrm{3}}})) \HOLFreeVar{C}
LextensionSimplDot:
\HOLTokenTurnstile{} \HOLConst{extends} \HOLConst{L_Sequent} \HOLFreeVar{E} \HOLSymConst{\HOLTokenConj{}}
   \HOLConst{gentzenSequent} \HOLFreeVar{E} (\HOLConst{OneForm} (\HOLFreeVar{A} \HOLSymConst{\HOLTokenProd{}} (\HOLFreeVar{B} \HOLSymConst{\HOLTokenProd{}} \HOLFreeVar{C}))) \HOLFreeVar{D} \HOLSymConst{\HOLTokenImp{}}
   \HOLConst{gentzenSequent} \HOLFreeVar{E} (\HOLConst{OneForm} (\HOLFreeVar{A} \HOLSymConst{\HOLTokenProd{}} \HOLFreeVar{B} \HOLSymConst{\HOLTokenProd{}} \HOLFreeVar{C})) \HOLFreeVar{D}
LextensionSimplDot':
\HOLTokenTurnstile{} \HOLConst{extends} \HOLConst{L_Sequent} \HOLFreeVar{E} \HOLSymConst{\HOLTokenConj{}}
   \HOLConst{gentzenSequent} \HOLFreeVar{E} (\HOLConst{OneForm} (\HOLFreeVar{A} \HOLSymConst{\HOLTokenProd{}} \HOLFreeVar{B} \HOLSymConst{\HOLTokenProd{}} \HOLFreeVar{C})) \HOLFreeVar{D} \HOLSymConst{\HOLTokenImp{}}
   \HOLConst{gentzenSequent} \HOLFreeVar{E} (\HOLConst{OneForm} (\HOLFreeVar{A} \HOLSymConst{\HOLTokenProd{}} (\HOLFreeVar{B} \HOLSymConst{\HOLTokenProd{}} \HOLFreeVar{C}))) \HOLFreeVar{D}
mainGeach:
\HOLTokenTurnstile{} \HOLConst{extends} \HOLConst{L_Sequent} \HOLFreeVar{E} \HOLSymConst{\HOLTokenImp{}}
   \HOLConst{gentzenSequent} \HOLFreeVar{E} (\HOLConst{OneForm} (\HOLFreeVar{A} \HOLSymConst{/} \HOLFreeVar{B})) (\HOLFreeVar{A} \HOLSymConst{/} \HOLFreeVar{C} \HOLSymConst{/} (\HOLFreeVar{B} \HOLSymConst{/} \HOLFreeVar{C}))
mainGeach':
\HOLTokenTurnstile{} \HOLConst{extends} \HOLConst{L_Sequent} \HOLFreeVar{E} \HOLSymConst{\HOLTokenImp{}}
   \HOLConst{gentzenSequent} \HOLFreeVar{E} (\HOLConst{OneForm} (\HOLFreeVar{B} \HOLSymConst{\HOLTokenBackslash} \HOLFreeVar{A})) ((\HOLFreeVar{C} \HOLSymConst{\HOLTokenBackslash} \HOLFreeVar{B}) \HOLSymConst{\HOLTokenBackslash} \HOLFreeVar{C} \HOLSymConst{\HOLTokenBackslash} \HOLFreeVar{A})
secondaryGeach:
\HOLTokenTurnstile{} \HOLConst{extends} \HOLConst{L_Sequent} \HOLFreeVar{E} \HOLSymConst{\HOLTokenImp{}}
   \HOLConst{gentzenSequent} \HOLFreeVar{E} (\HOLConst{OneForm} (\HOLFreeVar{B} \HOLSymConst{/} \HOLFreeVar{C})) ((\HOLFreeVar{A} \HOLSymConst{/} \HOLFreeVar{B}) \HOLSymConst{\HOLTokenBackslash} (\HOLFreeVar{A} \HOLSymConst{/} \HOLFreeVar{C}))
secondaryGeach':
\HOLTokenTurnstile{} \HOLConst{extends} \HOLConst{L_Sequent} \HOLFreeVar{E} \HOLSymConst{\HOLTokenImp{}}
   \HOLConst{gentzenSequent} \HOLFreeVar{E} (\HOLConst{OneForm} (\HOLFreeVar{C} \HOLSymConst{\HOLTokenBackslash} \HOLFreeVar{B})) (\HOLFreeVar{C} \HOLSymConst{\HOLTokenBackslash} \HOLFreeVar{A} \HOLSymConst{/} \HOLFreeVar{B} \HOLSymConst{\HOLTokenBackslash} \HOLFreeVar{A})
composition:
\HOLTokenTurnstile{} \HOLConst{extends} \HOLConst{L_Sequent} \HOLFreeVar{E} \HOLSymConst{\HOLTokenImp{}}
   \HOLConst{gentzenSequent} \HOLFreeVar{E} (\HOLConst{OneForm} (\HOLFreeVar{A} \HOLSymConst{/} \HOLFreeVar{B} \HOLSymConst{\HOLTokenProd{}} (\HOLFreeVar{B} \HOLSymConst{/} \HOLFreeVar{C}))) (\HOLFreeVar{A} \HOLSymConst{/} \HOLFreeVar{C})
composition':
\HOLTokenTurnstile{} \HOLConst{extends} \HOLConst{L_Sequent} \HOLFreeVar{E} \HOLSymConst{\HOLTokenImp{}}
   \HOLConst{gentzenSequent} \HOLFreeVar{E} (\HOLConst{OneForm} (\HOLFreeVar{C} \HOLSymConst{\HOLTokenBackslash} \HOLFreeVar{B} \HOLSymConst{\HOLTokenProd{}} \HOLFreeVar{B} \HOLSymConst{\HOLTokenBackslash} \HOLFreeVar{A})) (\HOLFreeVar{C} \HOLSymConst{\HOLTokenBackslash} \HOLFreeVar{A})
restructuring:
\HOLTokenTurnstile{} \HOLConst{extends} \HOLConst{L_Sequent} \HOLFreeVar{E} \HOLSymConst{\HOLTokenImp{}}
   \HOLConst{gentzenSequent} \HOLFreeVar{E} (\HOLConst{OneForm} (\HOLFreeVar{A} \HOLSymConst{\HOLTokenBackslash} \HOLFreeVar{B} \HOLSymConst{/} \HOLFreeVar{C})) (\HOLFreeVar{A} \HOLSymConst{\HOLTokenBackslash} (\HOLFreeVar{B} \HOLSymConst{/} \HOLFreeVar{C}))
restructuring':
\HOLTokenTurnstile{} \HOLConst{extends} \HOLConst{L_Sequent} \HOLFreeVar{E} \HOLSymConst{\HOLTokenImp{}}
   \HOLConst{gentzenSequent} \HOLFreeVar{E} (\HOLConst{OneForm} (\HOLFreeVar{A} \HOLSymConst{\HOLTokenBackslash} (\HOLFreeVar{B} \HOLSymConst{/} \HOLFreeVar{C}))) (\HOLFreeVar{A} \HOLSymConst{\HOLTokenBackslash} \HOLFreeVar{B} \HOLSymConst{/} \HOLFreeVar{C})
currying:
\HOLTokenTurnstile{} \HOLConst{extends} \HOLConst{L_Sequent} \HOLFreeVar{E} \HOLSymConst{\HOLTokenImp{}}
   \HOLConst{gentzenSequent} \HOLFreeVar{E} (\HOLConst{OneForm} (\HOLFreeVar{A} \HOLSymConst{/} (\HOLFreeVar{B} \HOLSymConst{\HOLTokenProd{}} \HOLFreeVar{C}))) (\HOLFreeVar{A} \HOLSymConst{/} \HOLFreeVar{C} \HOLSymConst{/} \HOLFreeVar{B})
currying':
\HOLTokenTurnstile{} \HOLConst{extends} \HOLConst{L_Sequent} \HOLFreeVar{E} \HOLSymConst{\HOLTokenImp{}}
   \HOLConst{gentzenSequent} \HOLFreeVar{E} (\HOLConst{OneForm} (\HOLFreeVar{A} \HOLSymConst{/} \HOLFreeVar{C} \HOLSymConst{/} \HOLFreeVar{B})) (\HOLFreeVar{A} \HOLSymConst{/} (\HOLFreeVar{B} \HOLSymConst{\HOLTokenProd{}} \HOLFreeVar{C}))
decurrying:
\HOLTokenTurnstile{} \HOLConst{extends} \HOLConst{L_Sequent} \HOLFreeVar{E} \HOLSymConst{\HOLTokenImp{}}
   \HOLConst{gentzenSequent} \HOLFreeVar{E} (\HOLConst{OneForm} ((\HOLFreeVar{A} \HOLSymConst{\HOLTokenProd{}} \HOLFreeVar{B}) \HOLSymConst{\HOLTokenBackslash} \HOLFreeVar{C})) (\HOLFreeVar{B} \HOLSymConst{\HOLTokenBackslash} \HOLFreeVar{A} \HOLSymConst{\HOLTokenBackslash} \HOLFreeVar{C})
decurrying':
\HOLTokenTurnstile{} \HOLConst{extends} \HOLConst{L_Sequent} \HOLFreeVar{E} \HOLSymConst{\HOLTokenImp{}}
   \HOLConst{gentzenSequent} \HOLFreeVar{E} (\HOLConst{OneForm} (\HOLFreeVar{B} \HOLSymConst{\HOLTokenBackslash} \HOLFreeVar{A} \HOLSymConst{\HOLTokenBackslash} \HOLFreeVar{C})) ((\HOLFreeVar{A} \HOLSymConst{\HOLTokenProd{}} \HOLFreeVar{B}) \HOLSymConst{\HOLTokenBackslash} \HOLFreeVar{C})
\end{alltt}

For Lambek Calculus extended from \textbf{NLP}, which supports both
commutativity we have proved the following extra
theorems:
\begin{alltt}
NLPextensionSimpl:
\HOLTokenTurnstile{} \HOLConst{extends} \HOLConst{NLP_Sequent} \HOLFreeVar{E} \HOLSymConst{\HOLTokenConj{}} \HOLConst{gentzenSequent} \HOLFreeVar{E} (\HOLConst{Comma} \HOLFreeVar{T\sb{\mathrm{1}}} \HOLFreeVar{T\sb{\mathrm{2}}}) \HOLFreeVar{C} \HOLSymConst{\HOLTokenImp{}}
   \HOLConst{gentzenSequent} \HOLFreeVar{E} (\HOLConst{Comma} \HOLFreeVar{T\sb{\mathrm{2}}} \HOLFreeVar{T\sb{\mathrm{1}}}) \HOLFreeVar{C}
NLPextensionSimplDot:
\HOLTokenTurnstile{} \HOLConst{extends} \HOLConst{NLP_Sequent} \HOLFreeVar{E} \HOLSymConst{\HOLTokenConj{}}
   \HOLConst{gentzenSequent} \HOLFreeVar{E} (\HOLConst{OneForm} (\HOLFreeVar{A} \HOLSymConst{\HOLTokenProd{}} \HOLFreeVar{B})) \HOLFreeVar{C} \HOLSymConst{\HOLTokenImp{}}
   \HOLConst{gentzenSequent} \HOLFreeVar{E} (\HOLConst{OneForm} (\HOLFreeVar{B} \HOLSymConst{\HOLTokenProd{}} \HOLFreeVar{A})) \HOLFreeVar{C}
permutation:
\HOLTokenTurnstile{} \HOLConst{extends} \HOLConst{NLP_Sequent} \HOLFreeVar{E} \HOLSymConst{\HOLTokenImp{}}
   \HOLConst{gentzenSequent} \HOLFreeVar{E} (\HOLConst{OneForm} \HOLFreeVar{A}) (\HOLFreeVar{B} \HOLSymConst{\HOLTokenBackslash} \HOLFreeVar{C}) \HOLSymConst{\HOLTokenImp{}}
   \HOLConst{gentzenSequent} \HOLFreeVar{E} (\HOLConst{OneForm} \HOLFreeVar{B}) (\HOLFreeVar{A} \HOLSymConst{\HOLTokenBackslash} \HOLFreeVar{C})
exchange:
\HOLTokenTurnstile{} \HOLConst{extends} \HOLConst{NLP_Sequent} \HOLFreeVar{E} \HOLSymConst{\HOLTokenImp{}}
   \HOLConst{gentzenSequent} \HOLFreeVar{E} (\HOLConst{OneForm} (\HOLFreeVar{A} \HOLSymConst{/} \HOLFreeVar{B})) (\HOLFreeVar{B} \HOLSymConst{\HOLTokenBackslash} \HOLFreeVar{A})
exchange':
\HOLTokenTurnstile{} \HOLConst{extends} \HOLConst{NLP_Sequent} \HOLFreeVar{E} \HOLSymConst{\HOLTokenImp{}}
   \HOLConst{gentzenSequent} \HOLFreeVar{E} (\HOLConst{OneForm} (\HOLFreeVar{B} \HOLSymConst{\HOLTokenBackslash} \HOLFreeVar{A})) (\HOLFreeVar{A} \HOLSymConst{/} \HOLFreeVar{B})
preposing:
\HOLTokenTurnstile{} \HOLConst{extends} \HOLConst{NLP_Sequent} \HOLFreeVar{E} \HOLSymConst{\HOLTokenImp{}}
   \HOLConst{gentzenSequent} \HOLFreeVar{E} (\HOLConst{OneForm} \HOLFreeVar{A}) (\HOLFreeVar{B} \HOLSymConst{/} (\HOLFreeVar{B} \HOLSymConst{/} \HOLFreeVar{A}))
postposing:
\HOLTokenTurnstile{} \HOLConst{extends} \HOLConst{NLP_Sequent} \HOLFreeVar{E} \HOLSymConst{\HOLTokenImp{}}
   \HOLConst{gentzenSequent} \HOLFreeVar{E} (\HOLConst{OneForm} \HOLFreeVar{A}) ((\HOLFreeVar{A} \HOLSymConst{\HOLTokenBackslash} \HOLFreeVar{B}) \HOLSymConst{\HOLTokenBackslash} \HOLFreeVar{B})
\end{alltt}

For Lambek Calculus extended from \textbf{LP}, which supports both commutativity and
associativity, we have proved the following theorems:
\begin{alltt}
mixedComposition:
\HOLTokenTurnstile{} \HOLConst{extends} \HOLConst{LP_Sequent} \HOLFreeVar{E} \HOLSymConst{\HOLTokenImp{}}
   \HOLConst{gentzenSequent} \HOLFreeVar{E} (\HOLConst{OneForm} (\HOLFreeVar{A} \HOLSymConst{/} \HOLFreeVar{B} \HOLSymConst{\HOLTokenProd{}} \HOLFreeVar{C} \HOLSymConst{\HOLTokenBackslash} \HOLFreeVar{B})) (\HOLFreeVar{C} \HOLSymConst{\HOLTokenBackslash} \HOLFreeVar{A})
mixedComposition':
\HOLTokenTurnstile{} \HOLConst{extends} \HOLConst{LP_Sequent} \HOLFreeVar{E} \HOLSymConst{\HOLTokenImp{}}
   \HOLConst{gentzenSequent} \HOLFreeVar{E} (\HOLConst{OneForm} (\HOLFreeVar{B} \HOLSymConst{/} \HOLFreeVar{C} \HOLSymConst{\HOLTokenProd{}} \HOLFreeVar{B} \HOLSymConst{\HOLTokenBackslash} \HOLFreeVar{A})) (\HOLFreeVar{A} \HOLSymConst{/} \HOLFreeVar{C})
\end{alltt}

\section{Equivalences between three deduction systems}

So far we have introduced three deduction systems for Lambek Calculus:
(axiomatic) syntactic calculus, natural deduction and gentzen's
sequent calculus. The convincing of syntactic calculus
comes from intuition and the fact that all axiom rules are extremely
simple, while the correctness of natural deduction rules and sequent
calculus rules is not that clear.  Before fully switching to use
natural deduction or sequent calculus as the alternative deduction system, we have to prove the
equivalences between them and the original syntactic calculus.

\subsection{Equivalence between Syntactic Calculus and Sequent Calculus}

By equivalence, basically we want to know if any theorem about
categories in \HOLinline{\HOLConst{arrow}} relation has also a proof in natural
deduction and sequent calculus, and conversely if any theorems in
latter systems has also a proof in syntactac calculus. In another
words, the set of theorems about syntactic categories proven in these three
deduction systems are exactly the same. However, there're two
difficulties we must resolve first:
\begin{enumerate}
\item the corresponce between arrow extensions (from Form to Form) and
  gentzen extensions (from Term to Term) must be defined and proved.
\item There's unique translation from Term to Form, but the other
  direction is not unique.
\end{enumerate}

The first problem are handled in two directions: from arrow to gentzen
extensions, then from genzen to arrow extensions.   In the first
direction we have the following definition:
\begin{alltt}
\HOLTokenTurnstile{} \HOLConst{arrowToGentzenExt} \HOLFreeVar{X} \HOLFreeVar{E} \HOLSymConst{\HOLTokenEquiv{}}
   \HOLSymConst{\HOLTokenForall{}}\HOLBoundVar{A} \HOLBoundVar{B}. \HOLFreeVar{X} \HOLBoundVar{A} \HOLBoundVar{B} \HOLSymConst{\HOLTokenImp{}} \HOLConst{gentzenSequent} \HOLFreeVar{E} (\HOLConst{OneForm} \HOLBoundVar{A}) \HOLBoundVar{B}
\end{alltt}
Noticed that, this is not a function mapping any arrow extension to a
gentzen extension, instead we need to use \HOLinline{\HOLConst{gentzenSequent}}
relation to handle the \HOLinline{\HOLConst{OneForm}} quoting of the first parameter
of arrow extension. Based on this definition, we have proved the
correspondences between arrow and gentzen extensions for \textbf{NL},
\textbf{L}, \textbf{NLP}, \textbf{LP}:
\begin{alltt}
NLToNL_Sequent:
\HOLTokenTurnstile{} \HOLConst{arrowToGentzenExt} \HOLConst{NL} \HOLConst{NL_Sequent}
NLPToNLP_Sequent:
\HOLTokenTurnstile{} \HOLConst{arrowToGentzenExt} \HOLConst{NLP} \HOLConst{NLP_Sequent}
LToL_Sequent:
\HOLTokenTurnstile{} \HOLConst{arrowToGentzenExt} \HOLConst{L} \HOLConst{L_Sequent}
LPToLP_Sequent:
\HOLTokenTurnstile{} \HOLConst{arrowToGentzenExt} \HOLConst{LP} \HOLConst{LP_Sequent}
\end{alltt}

The other direction is more subtle. From gentzen extension to arrow
extension, since the translation from Term to Form is unique, the
relationship between two kind of extensions can be defined just by
characteristics of themselves:
\begin{alltt}
\HOLTokenTurnstile{} \HOLConst{gentzenToArrowExt} \HOLFreeVar{E} \HOLFreeVar{X} \HOLSymConst{\HOLTokenEquiv{}}
   \HOLSymConst{\HOLTokenForall{}}\HOLBoundVar{T\sb{\mathrm{1}}} \HOLBoundVar{T\sb{\mathrm{2}}}.
     \HOLFreeVar{E} \HOLBoundVar{T\sb{\mathrm{1}}} \HOLBoundVar{T\sb{\mathrm{2}}} \HOLSymConst{\HOLTokenImp{}} \HOLFreeVar{X} (\HOLConst{deltaTranslation} \HOLBoundVar{T\sb{\mathrm{2}}}) (\HOLConst{deltaTranslation} \HOLBoundVar{T\sb{\mathrm{1}}})
\end{alltt}
However, noticed the interesting part: instead of saying \HOLinline{\HOLFreeVar{E} \HOLFreeVar{T\sb{\mathrm{1}}} \HOLFreeVar{T\sb{\mathrm{2}}} \HOLSymConst{\HOLTokenImp{}}
           \HOLFreeVar{X} (\HOLConst{deltaTranslation} \HOLFreeVar{T\sb{\mathrm{1}}}) (\HOLConst{deltaTranslation} \HOLFreeVar{T\sb{\mathrm{2}}})}, we must use
swap the order of \HOLinline{\HOLConst{deltaTranslation} \HOLFreeVar{T\sb{\mathrm{1}}}} and
\HOLinline{\HOLConst{deltaTranslation} \HOLFreeVar{T\sb{\mathrm{2}}}}. The reason seems connected with the
\texttt{CutRule}, and without defing like this we can't prove the
equivalece between the three deduction systems. However, we do can
prove correspondences between arrow and gentzen extensions for \textbf{NL},
\textbf{L}, \textbf{NLP}, \textbf{LP} and other general results using
this definition:
\begin{alltt}
NL_SequentToNL:
\HOLTokenTurnstile{} \HOLConst{gentzenToArrowExt} \HOLConst{NL_Sequent} \HOLConst{NL}
NLP_SequentToNLP:
\HOLTokenTurnstile{} \HOLConst{gentzenToArrowExt} \HOLConst{NLP_Sequent} \HOLConst{NLP}
L_SequentToL:
\HOLTokenTurnstile{} \HOLConst{gentzenToArrowExt} \HOLConst{L_Sequent} \HOLConst{L}
LP_SequentToLP:
\HOLTokenTurnstile{} \HOLConst{gentzenToArrowExt} \HOLConst{LP_Sequent} \HOLConst{LP}
\end{alltt}

Also noticed that, in the definition of \HOLinline{\HOLConst{gentzenToArrowExt}},
when the gentzen extension contains more contents (i.e. more pairs of
Terms satisfy the relation) then what's required by arrow extension,
the whole definition is still satisfied.  To actually define a
function which precisely translate a gentzen extension into unique
arrow extension and make sure the resulting arrow extension satisfy
the definition of \HOLinline{\HOLConst{gentzenToArrowExt}}, we can define this
function like this:
\begin{alltt}
\HOLTokenTurnstile{} \HOLConst{ToArrowExt} \HOLFreeVar{E} \HOLSymConst{=}
   \HOLConst{CURRY}
     \HOLTokenLeftbrace{}(\HOLConst{deltaTranslation} \HOLBoundVar{y}\HOLSymConst{,}\HOLConst{deltaTranslation} \HOLBoundVar{x}) \HOLTokenBar{}
      (\HOLBoundVar{x}\HOLSymConst{,}\HOLBoundVar{y}) \HOLSymConst{\HOLTokenIn{}} \HOLConst{UNCURRY} \HOLFreeVar{E}\HOLTokenRightbrace{}
\end{alltt}
Here we used HOL4's set theory package, and the use of \texttt{CURRY}
and \texttt{UNCURRY} is to converse between standard mathemics
definition of relations (with type ``\HOLinline{\ensuremath{\alpha} \HOLTyOp{Form} \HOLTokenProd{} \ensuremath{\alpha} \HOLTyOp{Form} -> \HOLTyOp{bool}}'') and relations in higher order logics (with type
``\HOLinline{\ensuremath{\alpha} \HOLTyOp{arrow_extension}}''. We can prove that, the output
of this function indeed satisfy the definition of \HOLinline{\HOLConst{gentzenToArrowExt}}:
\begin{alltt}
gentzenToArrowExt_thm:
\HOLTokenTurnstile{} \HOLConst{gentzenToArrowExt} \HOLFreeVar{E} (\HOLConst{ToArrowExt} \HOLFreeVar{E})
\end{alltt}

With all above devices, now the equivalences between (axiomatic) syntactic calculus and
gentzen's Sequent calculus and their common extensions have been
proved by the following theorems:

\begin{alltt}
arrowToGentzen:
\HOLTokenTurnstile{} \HOLConst{arrow} \HOLFreeVar{X} \HOLFreeVar{A} \HOLFreeVar{B} \HOLSymConst{\HOLTokenImp{}}
   \HOLConst{arrowToGentzenExt} \HOLFreeVar{X} \HOLFreeVar{E} \HOLSymConst{\HOLTokenImp{}}
   \HOLConst{gentzenSequent} \HOLFreeVar{E} (\HOLConst{OneForm} \HOLFreeVar{A}) \HOLFreeVar{B}
arrowToGentzenNL:
\HOLTokenTurnstile{} \HOLConst{arrow} \HOLConst{NL} \HOLFreeVar{A} \HOLFreeVar{B} \HOLSymConst{\HOLTokenImp{}} \HOLConst{gentzenSequent} \HOLConst{NL_Sequent} (\HOLConst{OneForm} \HOLFreeVar{A}) \HOLFreeVar{B}
arrowToGentzenNLP:
\HOLTokenTurnstile{} \HOLConst{arrow} \HOLConst{NLP} \HOLFreeVar{A} \HOLFreeVar{B} \HOLSymConst{\HOLTokenImp{}} \HOLConst{gentzenSequent} \HOLConst{NLP_Sequent} (\HOLConst{OneForm} \HOLFreeVar{A}) \HOLFreeVar{B}
arrowToGentzenL:
\HOLTokenTurnstile{} \HOLConst{arrow} \HOLConst{L} \HOLFreeVar{A} \HOLFreeVar{B} \HOLSymConst{\HOLTokenImp{}} \HOLConst{gentzenSequent} \HOLConst{L_Sequent} (\HOLConst{OneForm} \HOLFreeVar{A}) \HOLFreeVar{B}
arrowToGentzenLP:
\HOLTokenTurnstile{} \HOLConst{arrow} \HOLConst{LP} \HOLFreeVar{A} \HOLFreeVar{B} \HOLSymConst{\HOLTokenImp{}} \HOLConst{gentzenSequent} \HOLConst{LP_Sequent} (\HOLConst{OneForm} \HOLFreeVar{A}) \HOLFreeVar{B}

gentzenToArrow:
\HOLTokenTurnstile{} \HOLConst{gentzenToArrowExt} \HOLFreeVar{E} \HOLFreeVar{X} \HOLSymConst{\HOLTokenConj{}} \HOLConst{gentzenSequent} \HOLFreeVar{E} \HOLFreeVar{Gamma} \HOLFreeVar{A} \HOLSymConst{\HOLTokenImp{}}
   \HOLConst{arrow} \HOLFreeVar{X} (\HOLConst{deltaTranslation} \HOLFreeVar{Gamma}) \HOLFreeVar{A}
NLGentzenToArrow:
\HOLTokenTurnstile{} \HOLConst{gentzenSequent} \HOLConst{NL_Sequent} \HOLFreeVar{Gamma} \HOLFreeVar{A} \HOLSymConst{\HOLTokenImp{}}
   \HOLConst{arrow} \HOLConst{NL} (\HOLConst{deltaTranslation} \HOLFreeVar{Gamma}) \HOLFreeVar{A}
NLPGentzenToArrow:
\HOLTokenTurnstile{} \HOLConst{gentzenSequent} \HOLConst{NLP_Sequent} \HOLFreeVar{Gamma} \HOLFreeVar{A} \HOLSymConst{\HOLTokenImp{}}
   \HOLConst{arrow} \HOLConst{NLP} (\HOLConst{deltaTranslation} \HOLFreeVar{Gamma}) \HOLFreeVar{A}
LGentzenToArrow:
\HOLTokenTurnstile{} \HOLConst{gentzenSequent} \HOLConst{L_Sequent} \HOLFreeVar{Gamma} \HOLFreeVar{A} \HOLSymConst{\HOLTokenImp{}}
   \HOLConst{arrow} \HOLConst{L} (\HOLConst{deltaTranslation} \HOLFreeVar{Gamma}) \HOLFreeVar{A}
LPGentzenToArrow:
\HOLTokenTurnstile{} \HOLConst{gentzenSequent} \HOLConst{LP_Sequent} \HOLFreeVar{Gamma} \HOLFreeVar{A} \HOLSymConst{\HOLTokenImp{}}
   \HOLConst{arrow} \HOLConst{LP} (\HOLConst{deltaTranslation} \HOLFreeVar{Gamma}) \HOLFreeVar{A}
\end{alltt}

This means, if we translated all Comma into Dots, gentzen's Sequent
Calculus actually doesn't prove anything new beyond the Syntactic
Calculus, although it has more rules, especially the CutRule.

\subsection{Equivalence between Natural Deduction and Sequent Calculus}

Surprisely, the equivalance between Natural Deduction and Sequent
Calculus requires extra assumptions about properties on gentzen
extensions. Generally speaking, without any restriction on gentzen
extension, Gentzen's sequent calculus has a larger theorem
set, therefore is stronger than Natural Deduction.

From Natural Deduction to Sequence Calculus, i.e. the easy direction, we can prove the following theorem by induction
on the structure of \HOLinline{\HOLConst{gentzenSequent}} relation:
\begin{alltt}
natDedToGentzen:
\HOLTokenTurnstile{} \HOLConst{natDed} \HOLFreeVar{E} \HOLFreeVar{Gamma} \HOLFreeVar{C} \HOLSymConst{\HOLTokenImp{}} \HOLConst{gentzenSequent} \HOLFreeVar{E} \HOLFreeVar{Gamma} \HOLFreeVar{C}
\end{alltt}

The other direction is more difficult to prove, and it contains an extra property called
\HOLinline{\HOLConst{condCutExt}} about the extensions. First we present directly the
following important result:
\begin{alltt}
gentzenToNatDed:
\HOLTokenTurnstile{} \HOLConst{gentzenSequent} \HOLFreeVar{E} \HOLFreeVar{Gamma} \HOLFreeVar{C} \HOLSymConst{\HOLTokenImp{}} \HOLConst{condCutExt} \HOLFreeVar{E} \HOLSymConst{\HOLTokenImp{}} \HOLConst{natDed} \HOLFreeVar{E} \HOLFreeVar{Gamma} \HOLFreeVar{C}
\end{alltt}
its the formal proof of this theorem is the longest proof in our
LambekTheory, and to successfully prove it, many lemmas are needed.

The definition of \HOLinline{\HOLConst{condCutExt}} used in above theorem is defined as follows:
\begin{alltt}
\HOLTokenTurnstile{} \HOLConst{condCutExt} \HOLFreeVar{E} \HOLSymConst{\HOLTokenEquiv{}}
   \HOLSymConst{\HOLTokenForall{}}\HOLBoundVar{Gamma} \HOLBoundVar{T\sb{\mathrm{1}}} \HOLBoundVar{T\sb{\mathrm{2}}} \HOLBoundVar{A} \HOLBoundVar{Delta}.
     \HOLFreeVar{E} \HOLBoundVar{T\sb{\mathrm{1}}} \HOLBoundVar{T\sb{\mathrm{2}}} \HOLSymConst{\HOLTokenImp{}}
     \HOLConst{replace} \HOLBoundVar{T\sb{\mathrm{2}}} \HOLBoundVar{Gamma} (\HOLConst{OneForm} \HOLBoundVar{A}) \HOLBoundVar{Delta} \HOLSymConst{\HOLTokenImp{}}
     \HOLSymConst{\HOLTokenExists{}}\HOLBoundVar{Gamma\sp{\prime}}.
       \HOLFreeVar{E} \HOLBoundVar{Gamma\sp{\prime}} \HOLBoundVar{Gamma} \HOLSymConst{\HOLTokenConj{}} \HOLConst{replace} \HOLBoundVar{T\sb{\mathrm{1}}} \HOLBoundVar{Gamma\sp{\prime}} (\HOLConst{OneForm} \HOLBoundVar{A}) \HOLBoundVar{Delta}
\end{alltt}
to understand the precise meaning of this definition, we may consider the
meaning of the replace operator and rewrite above definition into the
following mathematic definition:
\begin{equation*}
E\enskip T_1[A]\enskip T_2[A] \Longrightarrow \exists T_1[Delta].\enskip E\enskip T_1[Delta]\enskip T_2[Delta]
\end{equation*}
This seems like a kind of completeness of the arrow extension:
whenever we replace any \HOLinline{\HOLConst{OneForm}} term into another Term in the
pairs satisfying the gentzen extension, the resulting pairs must also
satisfy this extension. However, whenever the definition of extension
concerns only about \HOLinline{\HOLConst{Comma}}s, like in \texttt{L_Sequent_def} and
others, this condition is naturally satisfied: (although the formal
proofs are quite long for some of them)
\begin{alltt}
conditionOKNL:
\HOLTokenTurnstile{} \HOLConst{condCutExt} \HOLConst{NL_Sequent}
conditionOKNLP:
\HOLTokenTurnstile{} \HOLConst{condCutExt} \HOLConst{NLP_Sequent}
conditionOKL:
\HOLTokenTurnstile{} \HOLConst{condCutExt} \HOLConst{L_Sequent}
\end{alltt}
Besides, we have proved that, if two gentzen extensions both satisfy above
\HOLinline{\HOLConst{condCutExt}} property, so is their sum relation:
\begin{alltt}
condAddExt:
\HOLTokenTurnstile{} \HOLConst{condCutExt} \HOLFreeVar{E} \HOLSymConst{\HOLTokenConj{}} \HOLConst{condCutExt} \HOLFreeVar{E\sp{\prime}} \HOLSymConst{\HOLTokenImp{}}
   \HOLConst{condCutExt} (\HOLConst{add_extension} \HOLFreeVar{E} \HOLFreeVar{E\sp{\prime}})
\end{alltt}

The formal proofs of \texttt{gentzenToNatDed} also depends on the
following lemmas which themselves have long proofs:
\begin{alltt}
CutNatDed:
\HOLTokenTurnstile{} \HOLConst{natDed} \HOLFreeVar{E} \HOLFreeVar{Gamma} \HOLFreeVar{C} \HOLSymConst{\HOLTokenImp{}}
   \HOLConst{condCutExt} \HOLFreeVar{E} \HOLSymConst{\HOLTokenImp{}}
   \HOLConst{natDed} \HOLFreeVar{E} \HOLFreeVar{Delta} \HOLFreeVar{A} \HOLSymConst{\HOLTokenImp{}}
   \HOLSymConst{\HOLTokenForall{}}\HOLBoundVar{Gamma\sp{\prime}}.
     \HOLConst{replace} \HOLFreeVar{Gamma} \HOLBoundVar{Gamma\sp{\prime}} (\HOLConst{OneForm} \HOLFreeVar{A}) \HOLFreeVar{Delta} \HOLSymConst{\HOLTokenImp{}} \HOLConst{natDed} \HOLFreeVar{E} \HOLBoundVar{Gamma\sp{\prime}} \HOLFreeVar{C}
natDedComposition:
\HOLTokenTurnstile{} \HOLConst{condCutExt} \HOLFreeVar{E} \HOLSymConst{\HOLTokenConj{}} \HOLConst{natDed} \HOLFreeVar{E} \HOLFreeVar{Gamma} \HOLFreeVar{F\sb{\mathrm{1}}} \HOLSymConst{\HOLTokenConj{}}
   \HOLConst{natDed} \HOLFreeVar{E} (\HOLConst{OneForm} \HOLFreeVar{F\sb{\mathrm{1}}}) \HOLFreeVar{F\sb{\mathrm{2}}} \HOLSymConst{\HOLTokenImp{}}
   \HOLConst{natDed} \HOLFreeVar{E} \HOLFreeVar{Gamma} \HOLFreeVar{F\sb{\mathrm{2}}}
\end{alltt}

Finally, we can merge the two direction together and get the following
equivalence theorems about natural deduction and gentzen's sequent
calculus for Lambek Calculus (both general and special cases),
although they're not used anywhere so far:
\begin{alltt}
gentzenEqNatDed:
\HOLTokenTurnstile{} \HOLConst{condCutExt} \HOLFreeVar{E} \HOLSymConst{\HOLTokenImp{}}
   (\HOLConst{gentzenSequent} \HOLFreeVar{E} \HOLFreeVar{Gamma} \HOLFreeVar{C} \HOLSymConst{\HOLTokenEquiv{}} \HOLConst{natDed} \HOLFreeVar{E} \HOLFreeVar{Gamma} \HOLFreeVar{C})
NLgentzenEqNatDed:
\HOLTokenTurnstile{} \HOLConst{gentzenSequent} \HOLConst{NL_Sequent} \HOLFreeVar{Gamma} \HOLFreeVar{C} \HOLSymConst{\HOLTokenEquiv{}}
   \HOLConst{natDed} \HOLConst{NL_Sequent} \HOLFreeVar{Gamma} \HOLFreeVar{C}
LgentzenEqNatDed:
\HOLTokenTurnstile{} \HOLConst{gentzenSequent} \HOLConst{L_Sequent} \HOLFreeVar{Gamma} \HOLFreeVar{C} \HOLSymConst{\HOLTokenEquiv{}} \HOLConst{natDed} \HOLConst{L_Sequent} \HOLFreeVar{Gamma} \HOLFreeVar{C}
NLPgentzenEqNatDed:
\HOLTokenTurnstile{} \HOLConst{gentzenSequent} \HOLConst{NLP_Sequent} \HOLFreeVar{Gamma} \HOLFreeVar{C} \HOLSymConst{\HOLTokenEquiv{}}
   \HOLConst{natDed} \HOLConst{NLP_Sequent} \HOLFreeVar{Gamma} \HOLFreeVar{C}
\end{alltt}

\subsection{Equivalence between Syntactic Calculus and Natural
  Deduction}

Combining results from previous two sections, the equivalence between Syntactic Calculus and Natural
  Deduction can be easily proved with gentzen's Sequent Calculus as
  intermediate steps:
\begin{alltt}
natDedToArrow:
\HOLTokenTurnstile{} \HOLConst{gentzenToArrowExt} \HOLFreeVar{E} \HOLFreeVar{X} \HOLSymConst{\HOLTokenImp{}}
   \HOLConst{natDed} \HOLFreeVar{E} \HOLFreeVar{Gamma} \HOLFreeVar{A} \HOLSymConst{\HOLTokenImp{}}
   \HOLConst{arrow} \HOLFreeVar{X} (\HOLConst{deltaTranslation} \HOLFreeVar{Gamma}) \HOLFreeVar{A}
natDedToArrow_E:
\HOLTokenTurnstile{} \HOLConst{natDed} \HOLFreeVar{E} \HOLFreeVar{Gamma} \HOLFreeVar{A} \HOLSymConst{\HOLTokenImp{}}
   \HOLConst{arrow} (\HOLConst{ToArrowExt} \HOLFreeVar{E}) (\HOLConst{deltaTranslation} \HOLFreeVar{Gamma}) \HOLFreeVar{A}
arrowToNatDed:
\HOLTokenTurnstile{} \HOLConst{condCutExt} \HOLFreeVar{E} \HOLSymConst{\HOLTokenConj{}} \HOLConst{arrowToGentzenExt} \HOLFreeVar{X} \HOLFreeVar{E} \HOLSymConst{\HOLTokenConj{}} \HOLConst{arrow} \HOLFreeVar{X} \HOLFreeVar{A} \HOLFreeVar{B} \HOLSymConst{\HOLTokenImp{}}
   \HOLConst{natDed} \HOLFreeVar{E} (\HOLConst{OneForm} \HOLFreeVar{A}) \HOLFreeVar{B}
\end{alltt}
There seems no way to get equivalence theorems between Syntactic
Calculus and Natural Deduction/Sequent Calculus, because of the
translations between Terms and Forms.

\section{A proof-theoretic formalization of Sequent Calculus}

Above formalizations for Lambek Calculus may have provided a good
basis for proving theorems about categories using any of the three
deduction systems, but one can only do this manually. For the
following two purposes, the current work is not enough:
\begin{enumerate}
\item Proving theorems about Sequent Calculus proofs, e. g. the
  cut-elimination theorem (for any Sequent Calculus proof, there exists a
  corresponding proof without using the CutRule).
\item Finding and generating Sequent Calculus proofs programmatically.
\end{enumerate}

For any of above purpose, we need a data structure to represent a
whole proof. HOL theorem prover has already provided data structures
to represent terms and theorems, but there's no built-in facility to
represent a whole proof. In another word, a ``proof'' is not a first-class
object in HOL.

However, Coq has built-in support for first-class proof object. In
Coq, it's possible to define the ``degree'' of a gentzenSequent proof
like this:
\begin{lstlisting}
Fixpoint degreeProof (Atoms : Set) (Gamma : Term Atoms) 
 (B : Form Atoms) (E : gentzen_extension) (p : gentzenSequent E Gamma B)
 {struct p} : nat :=
  match p with
  | Ax _ => 0
  | RightSlash _ _ _ H => degreeProof H
  | RightBackSlash _ _ _ H => degreeProof H
  | RightDot _ _ _ _ H1 H2 => max (degreeProof H1) (degreeProof H2)
  | LeftSlash _ _ _ _ _ _ R H1 H2 => max (degreeProof H1) (degreeProof H2)
  | LeftBackSlash _ _ _ _ _ _ R H1 H2 =>
      max (degreeProof H1) (degreeProof H2)
  | LeftDot _ _ _ _ _ R H => degreeProof H
  | CutRule _ _ _ A _ R H1 H2 =>
      max (max (degreeFormula A) (degreeProof H1)) (degreeProof H2)
  | SequentExtension _ _ _ _ _ E R H => degreeProof H
  end.
\end{lstlisting}

In HOL, it's impossible to defined the same thing directly, because
\HOLinline{\HOLConst{gentzenSequent} \HOLFreeVar{E} \HOLFreeVar{Gamma} \HOLFreeVar{B}} has the type \HOLinline{\HOLTyOp{bool}}. To fill
the gaps, we have to model a proof by a proof tree "object" by using a datatype 
definition. Our datatype definition is based on similar modelling work in Isabelle/HOL
for Display Logic \cite{Dawson:2006vc}, then all other related
inductive relation definitions and theorems are new. (But once the gaps
are filled, those theorems in Coq has also the same internal structure
in HOL.

\subsection{Proof objects}

A proof object in Sequent Calculus is represented as a datatype called
\HOLinline{\HOLFreeVar{Dertree}}. A \HOLinline{\HOLFreeVar{Dertree}} has at least a sequent as its head,
and by applying a rule it's derived from one or more other
sequents. Thus \HOLinline{\HOLFreeVar{Dertree}} is nothing but a tree of sequent with each node
identified by a rule name. A \HOLinline{\HOLFreeVar{Dertree}} as a proof can also be
``unfinished'', and in this case it contains only a (unproved)
sequent, nothing else:
\begin{lstlisting}
Datatype `Sequent = Sequent ('a gentzen_extension) ('a Term) ('a Form)`;
Datatype `Rule = SeqAxiom
	       | RightSlash | RightBackslash | RightDot
	       | LeftSlash  | LeftBackslash  | LeftDot
	       | CutRule    | SeqExt`;
Datatype `Dertree = Der ('a Sequent) Rule (Dertree list)
		  | Unf ('a Sequent)`;
\end{lstlisting}
Here the datatype \texttt{Sequent} is just a simple
container of three inner objects: a gentzen extension, a Term and a
Form. The datatype \texttt{Rule} is just an atomic  symbol, which
takes values from 9 possible rule names as primitive Sequent Calculus
rules.  So in our formal system, a sequent \HOLinline{\HOLConst{gentzenSequent} \HOLFreeVar{E} \HOLFreeVar{Gamma} \HOLFreeVar{B}} has the type \HOLinline{\HOLTyOp{bool}}, there's no way to access its
internat structure, and when it can appears along as a theorem.
On the other side a sequent \HOLinline{\HOLConst{Sequent} \HOLFreeVar{E} \HOLFreeVar{Gamma} \HOLFreeVar{B}} has the
type \HOLinline{\ensuremath{\alpha} \HOLTyOp{Sequent}}, it's just a container with accessors and its
correctness has to be proved separately.

With above data structures, the following accessors are defined to
access their internal structures:
\begin{alltt}
\HOLTokenTurnstile{} (\HOLSymConst{\HOLTokenForall{}}\HOLBoundVar{seq} \HOLBoundVar{v\sb{\mathrm{0}}} \HOLBoundVar{v\sb{\mathrm{1}}}. \HOLConst{head} (\HOLConst{Der} \HOLBoundVar{seq} \HOLBoundVar{v\sb{\mathrm{0}}} \HOLBoundVar{v\sb{\mathrm{1}}}) \HOLSymConst{=} \HOLBoundVar{seq}) \HOLSymConst{\HOLTokenConj{}}
   \HOLSymConst{\HOLTokenForall{}}\HOLBoundVar{seq}. \HOLConst{head} (\HOLConst{Unf} \HOLBoundVar{seq}) \HOLSymConst{=} \HOLBoundVar{seq}
\HOLTokenTurnstile{} (\HOLConst{concl} (\HOLConst{Unf} (\HOLConst{Sequent} \HOLFreeVar{E} \HOLFreeVar{Delta} \HOLFreeVar{A})) \HOLSymConst{=} \HOLFreeVar{A}) \HOLSymConst{\HOLTokenConj{}}
   (\HOLConst{concl} (\HOLConst{Der} (\HOLConst{Sequent} \HOLFreeVar{E} \HOLFreeVar{Delta} \HOLFreeVar{A}) \HOLFreeVar{v\sb{\mathrm{0}}} \HOLFreeVar{v\sb{\mathrm{1}}}) \HOLSymConst{=} \HOLFreeVar{A})
\HOLTokenTurnstile{} (\HOLConst{prems} (\HOLConst{Unf} (\HOLConst{Sequent} \HOLFreeVar{E} \HOLFreeVar{Delta} \HOLFreeVar{A})) \HOLSymConst{=} \HOLFreeVar{Delta}) \HOLSymConst{\HOLTokenConj{}}
   (\HOLConst{prems} (\HOLConst{Der} (\HOLConst{Sequent} \HOLFreeVar{E} \HOLFreeVar{Delta} \HOLFreeVar{A}) \HOLFreeVar{v\sb{\mathrm{0}}} \HOLFreeVar{v\sb{\mathrm{1}}}) \HOLSymConst{=} \HOLFreeVar{Delta})
\HOLTokenTurnstile{} (\HOLConst{exten} (\HOLConst{Unf} (\HOLConst{Sequent} \HOLFreeVar{E} \HOLFreeVar{Delta} \HOLFreeVar{A})) \HOLSymConst{=} \HOLFreeVar{E}) \HOLSymConst{\HOLTokenConj{}}
   (\HOLConst{exten} (\HOLConst{Der} (\HOLConst{Sequent} \HOLFreeVar{E} \HOLFreeVar{Delta} \HOLFreeVar{A}) \HOLFreeVar{v\sb{\mathrm{0}}} \HOLFreeVar{v\sb{\mathrm{1}}}) \HOLSymConst{=} \HOLFreeVar{E})
\end{alltt}
And now we can define \HOLinline{\HOLConst{degreeProof}} as follows:
\begin{alltt}
\HOLTokenTurnstile{} (\HOLSymConst{\HOLTokenForall{}}\HOLBoundVar{S}. \HOLConst{degreeProof} (\HOLConst{Der} \HOLBoundVar{S} \HOLConst{SeqAxiom} []) \HOLSymConst{=} \HOLNumLit{0}) \HOLSymConst{\HOLTokenConj{}}
   (\HOLSymConst{\HOLTokenForall{}}\HOLBoundVar{S} \HOLBoundVar{H}. \HOLConst{degreeProof} (\HOLConst{Der} \HOLBoundVar{S} \HOLConst{RightSlash} [\HOLBoundVar{H}]) \HOLSymConst{=} \HOLConst{degreeProof} \HOLBoundVar{H}) \HOLSymConst{\HOLTokenConj{}}
   (\HOLSymConst{\HOLTokenForall{}}\HOLBoundVar{S} \HOLBoundVar{H}.
      \HOLConst{degreeProof} (\HOLConst{Der} \HOLBoundVar{S} \HOLConst{RightBackslash} [\HOLBoundVar{H}]) \HOLSymConst{=} \HOLConst{degreeProof} \HOLBoundVar{H}) \HOLSymConst{\HOLTokenConj{}}
   (\HOLSymConst{\HOLTokenForall{}}\HOLBoundVar{S} \HOLBoundVar{H\sb{\mathrm{2}}} \HOLBoundVar{H\sb{\mathrm{1}}}.
      \HOLConst{degreeProof} (\HOLConst{Der} \HOLBoundVar{S} \HOLConst{RightDot} [\HOLBoundVar{H\sb{\mathrm{1}}}; \HOLBoundVar{H\sb{\mathrm{2}}}]) \HOLSymConst{=}
      \HOLConst{MAX} (\HOLConst{degreeProof} \HOLBoundVar{H\sb{\mathrm{1}}}) (\HOLConst{degreeProof} \HOLBoundVar{H\sb{\mathrm{2}}})) \HOLSymConst{\HOLTokenConj{}}
   (\HOLSymConst{\HOLTokenForall{}}\HOLBoundVar{S} \HOLBoundVar{H\sb{\mathrm{2}}} \HOLBoundVar{H\sb{\mathrm{1}}}.
      \HOLConst{degreeProof} (\HOLConst{Der} \HOLBoundVar{S} \HOLConst{LeftSlash} [\HOLBoundVar{H\sb{\mathrm{1}}}; \HOLBoundVar{H\sb{\mathrm{2}}}]) \HOLSymConst{=}
      \HOLConst{MAX} (\HOLConst{degreeProof} \HOLBoundVar{H\sb{\mathrm{1}}}) (\HOLConst{degreeProof} \HOLBoundVar{H\sb{\mathrm{2}}})) \HOLSymConst{\HOLTokenConj{}}
   (\HOLSymConst{\HOLTokenForall{}}\HOLBoundVar{S} \HOLBoundVar{H\sb{\mathrm{2}}} \HOLBoundVar{H\sb{\mathrm{1}}}.
      \HOLConst{degreeProof} (\HOLConst{Der} \HOLBoundVar{S} \HOLConst{LeftBackslash} [\HOLBoundVar{H\sb{\mathrm{1}}}; \HOLBoundVar{H\sb{\mathrm{2}}}]) \HOLSymConst{=}
      \HOLConst{MAX} (\HOLConst{degreeProof} \HOLBoundVar{H\sb{\mathrm{1}}}) (\HOLConst{degreeProof} \HOLBoundVar{H\sb{\mathrm{2}}})) \HOLSymConst{\HOLTokenConj{}}
   (\HOLSymConst{\HOLTokenForall{}}\HOLBoundVar{S} \HOLBoundVar{H}. \HOLConst{degreeProof} (\HOLConst{Der} \HOLBoundVar{S} \HOLConst{LeftDot} [\HOLBoundVar{H}]) \HOLSymConst{=} \HOLConst{degreeProof} \HOLBoundVar{H}) \HOLSymConst{\HOLTokenConj{}}
   (\HOLSymConst{\HOLTokenForall{}}\HOLBoundVar{S} \HOLBoundVar{H}. \HOLConst{degreeProof} (\HOLConst{Der} \HOLBoundVar{S} \HOLConst{SeqExt} [\HOLBoundVar{H}]) \HOLSymConst{=} \HOLConst{degreeProof} \HOLBoundVar{H}) \HOLSymConst{\HOLTokenConj{}}
   \HOLSymConst{\HOLTokenForall{}}\HOLBoundVar{S} \HOLBoundVar{H\sb{\mathrm{2}}} \HOLBoundVar{H\sb{\mathrm{1}}}.
     \HOLConst{degreeProof} (\HOLConst{Der} \HOLBoundVar{S} \HOLConst{CutRule} [\HOLBoundVar{H\sb{\mathrm{1}}}; \HOLBoundVar{H\sb{\mathrm{2}}}]) \HOLSymConst{=}
     \HOLConst{MAX} (\HOLConst{degreeFormula} (\HOLConst{concl} \HOLBoundVar{H\sb{\mathrm{1}}}))
       (\HOLConst{MAX} (\HOLConst{degreeProof} \HOLBoundVar{H\sb{\mathrm{1}}}) (\HOLConst{degreeProof} \HOLBoundVar{H\sb{\mathrm{2}}}))
\end{alltt}

Another closely related concept is the \HOLinline{\HOLConst{degreeFormula}} which
assign each Form as integer as its degree:
\begin{alltt}
\HOLTokenTurnstile{} (\HOLSymConst{\HOLTokenForall{}}\HOLBoundVar{C}. \HOLConst{degreeFormula} (\HOLConst{At} \HOLBoundVar{C}) \HOLSymConst{=} \HOLNumLit{1}) \HOLSymConst{\HOLTokenConj{}}
   (\HOLSymConst{\HOLTokenForall{}}\HOLBoundVar{F\sb{\mathrm{1}}} \HOLBoundVar{F\sb{\mathrm{2}}}.
      \HOLConst{degreeFormula} (\HOLBoundVar{F\sb{\mathrm{1}}} \HOLSymConst{/} \HOLBoundVar{F\sb{\mathrm{2}}}) \HOLSymConst{=}
      \HOLConst{SUC} (\HOLConst{MAX} (\HOLConst{degreeFormula} \HOLBoundVar{F\sb{\mathrm{1}}}) (\HOLConst{degreeFormula} \HOLBoundVar{F\sb{\mathrm{2}}}))) \HOLSymConst{\HOLTokenConj{}}
   (\HOLSymConst{\HOLTokenForall{}}\HOLBoundVar{F\sb{\mathrm{1}}} \HOLBoundVar{F\sb{\mathrm{2}}}.
      \HOLConst{degreeFormula} (\HOLBoundVar{F\sb{\mathrm{1}}} \HOLSymConst{\HOLTokenBackslash} \HOLBoundVar{F\sb{\mathrm{2}}}) \HOLSymConst{=}
      \HOLConst{SUC} (\HOLConst{MAX} (\HOLConst{degreeFormula} \HOLBoundVar{F\sb{\mathrm{1}}}) (\HOLConst{degreeFormula} \HOLBoundVar{F\sb{\mathrm{2}}}))) \HOLSymConst{\HOLTokenConj{}}
   \HOLSymConst{\HOLTokenForall{}}\HOLBoundVar{F\sb{\mathrm{1}}} \HOLBoundVar{F\sb{\mathrm{2}}}.
     \HOLConst{degreeFormula} (\HOLBoundVar{F\sb{\mathrm{1}}} \HOLSymConst{\HOLTokenProd{}} \HOLBoundVar{F\sb{\mathrm{2}}}) \HOLSymConst{=}
     \HOLConst{SUC} (\HOLConst{MAX} (\HOLConst{degreeFormula} \HOLBoundVar{F\sb{\mathrm{1}}}) (\HOLConst{degreeFormula} \HOLBoundVar{F\sb{\mathrm{2}}}))
\end{alltt}

The concept of sub-formula between two Forms that we mentioned several times in
previous sections, is now formally defined as an inductive relation:
\begin{alltt}
\HOLTokenTurnstile{} (\HOLSymConst{\HOLTokenForall{}}\HOLBoundVar{A}. \HOLConst{subFormula} \HOLBoundVar{A} \HOLBoundVar{A}) \HOLSymConst{\HOLTokenConj{}}
   (\HOLSymConst{\HOLTokenForall{}}\HOLBoundVar{A} \HOLBoundVar{B} \HOLBoundVar{C}. \HOLConst{subFormula} \HOLBoundVar{A} \HOLBoundVar{B} \HOLSymConst{\HOLTokenImp{}} \HOLConst{subFormula} \HOLBoundVar{A} (\HOLBoundVar{B} \HOLSymConst{/} \HOLBoundVar{C})) \HOLSymConst{\HOLTokenConj{}}
   (\HOLSymConst{\HOLTokenForall{}}\HOLBoundVar{A} \HOLBoundVar{B} \HOLBoundVar{C}. \HOLConst{subFormula} \HOLBoundVar{A} \HOLBoundVar{B} \HOLSymConst{\HOLTokenImp{}} \HOLConst{subFormula} \HOLBoundVar{A} (\HOLBoundVar{C} \HOLSymConst{/} \HOLBoundVar{B})) \HOLSymConst{\HOLTokenConj{}}
   (\HOLSymConst{\HOLTokenForall{}}\HOLBoundVar{A} \HOLBoundVar{B} \HOLBoundVar{C}. \HOLConst{subFormula} \HOLBoundVar{A} \HOLBoundVar{B} \HOLSymConst{\HOLTokenImp{}} \HOLConst{subFormula} \HOLBoundVar{A} (\HOLBoundVar{B} \HOLSymConst{\HOLTokenBackslash} \HOLBoundVar{C})) \HOLSymConst{\HOLTokenConj{}}
   (\HOLSymConst{\HOLTokenForall{}}\HOLBoundVar{A} \HOLBoundVar{B} \HOLBoundVar{C}. \HOLConst{subFormula} \HOLBoundVar{A} \HOLBoundVar{B} \HOLSymConst{\HOLTokenImp{}} \HOLConst{subFormula} \HOLBoundVar{A} (\HOLBoundVar{C} \HOLSymConst{\HOLTokenBackslash} \HOLBoundVar{B})) \HOLSymConst{\HOLTokenConj{}}
   (\HOLSymConst{\HOLTokenForall{}}\HOLBoundVar{A} \HOLBoundVar{B} \HOLBoundVar{C}. \HOLConst{subFormula} \HOLBoundVar{A} \HOLBoundVar{B} \HOLSymConst{\HOLTokenImp{}} \HOLConst{subFormula} \HOLBoundVar{A} (\HOLBoundVar{B} \HOLSymConst{\HOLTokenProd{}} \HOLBoundVar{C})) \HOLSymConst{\HOLTokenConj{}}
   \HOLSymConst{\HOLTokenForall{}}\HOLBoundVar{A} \HOLBoundVar{B} \HOLBoundVar{C}. \HOLConst{subFormula} \HOLBoundVar{A} \HOLBoundVar{B} \HOLSymConst{\HOLTokenImp{}} \HOLConst{subFormula} \HOLBoundVar{A} (\HOLBoundVar{C} \HOLSymConst{\HOLTokenProd{}} \HOLBoundVar{B})
\end{alltt}
And we have also proved many theorems about \HOLinline{\HOLConst{subFormula}}:
\begin{alltt}
subAt:
\HOLTokenTurnstile{} \HOLConst{subFormula} \HOLFreeVar{A} (\HOLConst{At} \HOLFreeVar{a}) \HOLSymConst{\HOLTokenImp{}} (\HOLFreeVar{A} \HOLSymConst{=} \HOLConst{At} \HOLFreeVar{a})
subSlash:
\HOLTokenTurnstile{} \HOLConst{subFormula} \HOLFreeVar{A} (\HOLFreeVar{B} \HOLSymConst{/} \HOLFreeVar{C}) \HOLSymConst{\HOLTokenImp{}}
   (\HOLFreeVar{A} \HOLSymConst{=} \HOLFreeVar{B} \HOLSymConst{/} \HOLFreeVar{C}) \HOLSymConst{\HOLTokenDisj{}} \HOLConst{subFormula} \HOLFreeVar{A} \HOLFreeVar{B} \HOLSymConst{\HOLTokenDisj{}} \HOLConst{subFormula} \HOLFreeVar{A} \HOLFreeVar{C}
subBackslash:
\HOLTokenTurnstile{} \HOLConst{subFormula} \HOLFreeVar{A} (\HOLFreeVar{B} \HOLSymConst{\HOLTokenBackslash} \HOLFreeVar{C}) \HOLSymConst{\HOLTokenImp{}}
   (\HOLFreeVar{A} \HOLSymConst{=} \HOLFreeVar{B} \HOLSymConst{\HOLTokenBackslash} \HOLFreeVar{C}) \HOLSymConst{\HOLTokenDisj{}} \HOLConst{subFormula} \HOLFreeVar{A} \HOLFreeVar{B} \HOLSymConst{\HOLTokenDisj{}} \HOLConst{subFormula} \HOLFreeVar{A} \HOLFreeVar{C}
subDot:
\HOLTokenTurnstile{} \HOLConst{subFormula} \HOLFreeVar{A} (\HOLFreeVar{B} \HOLSymConst{\HOLTokenProd{}} \HOLFreeVar{C}) \HOLSymConst{\HOLTokenImp{}}
   (\HOLFreeVar{A} \HOLSymConst{=} \HOLFreeVar{B} \HOLSymConst{\HOLTokenProd{}} \HOLFreeVar{C}) \HOLSymConst{\HOLTokenDisj{}} \HOLConst{subFormula} \HOLFreeVar{A} \HOLFreeVar{B} \HOLSymConst{\HOLTokenDisj{}} \HOLConst{subFormula} \HOLFreeVar{A} \HOLFreeVar{C}
subFormulaTrans:
\HOLTokenTurnstile{} \HOLConst{subFormula} \HOLFreeVar{A} \HOLFreeVar{B} \HOLSymConst{\HOLTokenImp{}} \HOLConst{subFormula} \HOLFreeVar{B} \HOLFreeVar{C} \HOLSymConst{\HOLTokenImp{}} \HOLConst{subFormula} \HOLFreeVar{A} \HOLFreeVar{C}
\end{alltt}

The sub-formula between a Form and a Term is called
\HOLinline{\HOLConst{subFormTerm}}, which is defined inductively upon
\HOLinline{\HOLConst{subFormula}}:
\begin{alltt}
\HOLTokenTurnstile{} (\HOLSymConst{\HOLTokenForall{}}\HOLBoundVar{A} \HOLBoundVar{B}. \HOLConst{subFormula} \HOLBoundVar{A} \HOLBoundVar{B} \HOLSymConst{\HOLTokenImp{}} \HOLConst{subFormTerm} \HOLBoundVar{A} (\HOLConst{OneForm} \HOLBoundVar{B})) \HOLSymConst{\HOLTokenConj{}}
   (\HOLSymConst{\HOLTokenForall{}}\HOLBoundVar{A} \HOLBoundVar{T\sb{\mathrm{1}}} \HOLBoundVar{T\sb{\mathrm{2}}}. \HOLConst{subFormTerm} \HOLBoundVar{A} \HOLBoundVar{T\sb{\mathrm{1}}} \HOLSymConst{\HOLTokenImp{}} \HOLConst{subFormTerm} \HOLBoundVar{A} (\HOLConst{Comma} \HOLBoundVar{T\sb{\mathrm{1}}} \HOLBoundVar{T\sb{\mathrm{2}}})) \HOLSymConst{\HOLTokenConj{}}
   \HOLSymConst{\HOLTokenForall{}}\HOLBoundVar{A} \HOLBoundVar{T\sb{\mathrm{1}}} \HOLBoundVar{T\sb{\mathrm{2}}}. \HOLConst{subFormTerm} \HOLBoundVar{A} \HOLBoundVar{T\sb{\mathrm{1}}} \HOLSymConst{\HOLTokenImp{}} \HOLConst{subFormTerm} \HOLBoundVar{A} (\HOLConst{Comma} \HOLBoundVar{T\sb{\mathrm{2}}} \HOLBoundVar{T\sb{\mathrm{1}}})
\end{alltt}
Then we have proved several important theorems about
\HOLinline{\HOLConst{subFormTerm}}, most concering about the \HOLinline{\HOLConst{replace}} operator:
\begin{alltt}
oneFormSubEQ:
\HOLTokenTurnstile{} \HOLConst{subFormTerm} \HOLFreeVar{A} (\HOLConst{OneForm} \HOLFreeVar{B}) \HOLSymConst{\HOLTokenEquiv{}} \HOLConst{subFormula} \HOLFreeVar{A} \HOLFreeVar{B}
comSub:
\HOLTokenTurnstile{} \HOLConst{subFormTerm} \HOLFreeVar{f} (\HOLConst{Comma} \HOLFreeVar{T\sb{\mathrm{1}}} \HOLFreeVar{T\sb{\mathrm{2}}}) \HOLSymConst{\HOLTokenImp{}}
   \HOLConst{subFormTerm} \HOLFreeVar{f} \HOLFreeVar{T\sb{\mathrm{1}}} \HOLSymConst{\HOLTokenDisj{}} \HOLConst{subFormTerm} \HOLFreeVar{f} \HOLFreeVar{T\sb{\mathrm{2}}}
subReplace1:
\HOLTokenTurnstile{} \HOLConst{replace} \HOLFreeVar{T\sb{\mathrm{1}}} \HOLFreeVar{T\sb{\mathrm{2}}} \HOLFreeVar{T\sb{\mathrm{3}}} \HOLFreeVar{T\sb{\mathrm{4}}} \HOLSymConst{\HOLTokenImp{}} \HOLConst{subFormTerm} \HOLFreeVar{f} \HOLFreeVar{T\sb{\mathrm{3}}} \HOLSymConst{\HOLTokenImp{}} \HOLConst{subFormTerm} \HOLFreeVar{f} \HOLFreeVar{T\sb{\mathrm{1}}}
subReplace2:
\HOLTokenTurnstile{} \HOLConst{replace} \HOLFreeVar{T\sb{\mathrm{1}}} \HOLFreeVar{T\sb{\mathrm{2}}} \HOLFreeVar{T\sb{\mathrm{3}}} \HOLFreeVar{T\sb{\mathrm{4}}} \HOLSymConst{\HOLTokenImp{}} \HOLConst{subFormTerm} \HOLFreeVar{f} \HOLFreeVar{T\sb{\mathrm{4}}} \HOLSymConst{\HOLTokenImp{}} \HOLConst{subFormTerm} \HOLFreeVar{f} \HOLFreeVar{T\sb{\mathrm{2}}}
subReplace3:
\HOLTokenTurnstile{} \HOLConst{replace} \HOLFreeVar{T\sb{\mathrm{1}}} \HOLFreeVar{T\sb{\mathrm{2}}} \HOLFreeVar{T\sb{\mathrm{3}}} \HOLFreeVar{T\sb{\mathrm{4}}} \HOLSymConst{\HOLTokenImp{}}
   \HOLConst{subFormTerm} \HOLFreeVar{X} \HOLFreeVar{T\sb{\mathrm{1}}} \HOLSymConst{\HOLTokenImp{}}
   \HOLConst{subFormTerm} \HOLFreeVar{X} \HOLFreeVar{T\sb{\mathrm{2}}} \HOLSymConst{\HOLTokenDisj{}} \HOLConst{subFormTerm} \HOLFreeVar{X} \HOLFreeVar{T\sb{\mathrm{3}}}
\end{alltt}

\subsection{Derivations of proof tree}

So how can we (manually) construct a proof for any theorem in Sequent calculus of
Lambek Calculus and prove the resulting \texttt{Dertree} object is
indeed a valid proof for this theorem?  Unfortunately previous Coq proof scripts
didn't give any hint, for this part the author has built everything needed
from the ground.

The idea comes from beta-reduction in $\lambda$-Calculus. First we
define the \emph{one-step} derivation from any unfinished
\texttt{Dertree} by applying one rule which is equivalent with one of
\HOLinline{\HOLConst{gentzenSequent}} rules:
\begin{alltt}
\HOLTokenTurnstile{} (\HOLSymConst{\HOLTokenForall{}}\HOLBoundVar{E} \HOLBoundVar{A}.
      \HOLConst{derivOne} (\HOLConst{Unf} (\HOLConst{Sequent} \HOLBoundVar{E} (\HOLConst{OneForm} \HOLBoundVar{A}) \HOLBoundVar{A}))
        (\HOLConst{Der} (\HOLConst{Sequent} \HOLBoundVar{E} (\HOLConst{OneForm} \HOLBoundVar{A}) \HOLBoundVar{A}) \HOLConst{SeqAxiom} [])) \HOLSymConst{\HOLTokenConj{}}
   (\HOLSymConst{\HOLTokenForall{}}\HOLBoundVar{E} \HOLBoundVar{Gamma} \HOLBoundVar{A} \HOLBoundVar{B}.
      \HOLConst{derivOne} (\HOLConst{Unf} (\HOLConst{Sequent} \HOLBoundVar{E} \HOLBoundVar{Gamma} (\HOLBoundVar{A} \HOLSymConst{/} \HOLBoundVar{B})))
        (\HOLConst{Der} (\HOLConst{Sequent} \HOLBoundVar{E} \HOLBoundVar{Gamma} (\HOLBoundVar{A} \HOLSymConst{/} \HOLBoundVar{B})) \HOLConst{RightSlash}
           [\HOLConst{Unf} (\HOLConst{Sequent} \HOLBoundVar{E} (\HOLConst{Comma} \HOLBoundVar{Gamma} (\HOLConst{OneForm} \HOLBoundVar{B})) \HOLBoundVar{A})])) \HOLSymConst{\HOLTokenConj{}}
   (\HOLSymConst{\HOLTokenForall{}}\HOLBoundVar{E} \HOLBoundVar{Gamma} \HOLBoundVar{A} \HOLBoundVar{B}.
      \HOLConst{derivOne} (\HOLConst{Unf} (\HOLConst{Sequent} \HOLBoundVar{E} \HOLBoundVar{Gamma} (\HOLBoundVar{B} \HOLSymConst{\HOLTokenBackslash} \HOLBoundVar{A})))
        (\HOLConst{Der} (\HOLConst{Sequent} \HOLBoundVar{E} \HOLBoundVar{Gamma} (\HOLBoundVar{B} \HOLSymConst{\HOLTokenBackslash} \HOLBoundVar{A})) \HOLConst{RightBackslash}
           [\HOLConst{Unf} (\HOLConst{Sequent} \HOLBoundVar{E} (\HOLConst{Comma} (\HOLConst{OneForm} \HOLBoundVar{B}) \HOLBoundVar{Gamma}) \HOLBoundVar{A})])) \HOLSymConst{\HOLTokenConj{}}
   (\HOLSymConst{\HOLTokenForall{}}\HOLBoundVar{E} \HOLBoundVar{Gamma} \HOLBoundVar{Delta} \HOLBoundVar{A} \HOLBoundVar{B}.
      \HOLConst{derivOne} (\HOLConst{Unf} (\HOLConst{Sequent} \HOLBoundVar{E} (\HOLConst{Comma} \HOLBoundVar{Gamma} \HOLBoundVar{Delta}) (\HOLBoundVar{A} \HOLSymConst{\HOLTokenProd{}} \HOLBoundVar{B})))
        (\HOLConst{Der} (\HOLConst{Sequent} \HOLBoundVar{E} (\HOLConst{Comma} \HOLBoundVar{Gamma} \HOLBoundVar{Delta}) (\HOLBoundVar{A} \HOLSymConst{\HOLTokenProd{}} \HOLBoundVar{B})) \HOLConst{RightDot}
           [\HOLConst{Unf} (\HOLConst{Sequent} \HOLBoundVar{E} \HOLBoundVar{Gamma} \HOLBoundVar{A});
            \HOLConst{Unf} (\HOLConst{Sequent} \HOLBoundVar{E} \HOLBoundVar{Delta} \HOLBoundVar{B})])) \HOLSymConst{\HOLTokenConj{}}
   (\HOLSymConst{\HOLTokenForall{}}\HOLBoundVar{E} \HOLBoundVar{Gamma} \HOLBoundVar{Gamma\sp{\prime}} \HOLBoundVar{Delta} \HOLBoundVar{A} \HOLBoundVar{B} \HOLBoundVar{C}.
      \HOLConst{replace} \HOLBoundVar{Gamma} \HOLBoundVar{Gamma\sp{\prime}} (\HOLConst{OneForm} \HOLBoundVar{A})
        (\HOLConst{Comma} (\HOLConst{OneForm} (\HOLBoundVar{A} \HOLSymConst{/} \HOLBoundVar{B})) \HOLBoundVar{Delta}) \HOLSymConst{\HOLTokenImp{}}
      \HOLConst{derivOne} (\HOLConst{Unf} (\HOLConst{Sequent} \HOLBoundVar{E} \HOLBoundVar{Gamma\sp{\prime}} \HOLBoundVar{C}))
        (\HOLConst{Der} (\HOLConst{Sequent} \HOLBoundVar{E} \HOLBoundVar{Gamma\sp{\prime}} \HOLBoundVar{C}) \HOLConst{LeftSlash}
           [\HOLConst{Unf} (\HOLConst{Sequent} \HOLBoundVar{E} \HOLBoundVar{Gamma} \HOLBoundVar{C});
            \HOLConst{Unf} (\HOLConst{Sequent} \HOLBoundVar{E} \HOLBoundVar{Delta} \HOLBoundVar{B})])) \HOLSymConst{\HOLTokenConj{}}
   (\HOLSymConst{\HOLTokenForall{}}\HOLBoundVar{E} \HOLBoundVar{Gamma} \HOLBoundVar{Gamma\sp{\prime}} \HOLBoundVar{Delta} \HOLBoundVar{A} \HOLBoundVar{B} \HOLBoundVar{C}.
      \HOLConst{replace} \HOLBoundVar{Gamma} \HOLBoundVar{Gamma\sp{\prime}} (\HOLConst{OneForm} \HOLBoundVar{A})
        (\HOLConst{Comma} \HOLBoundVar{Delta} (\HOLConst{OneForm} (\HOLBoundVar{B} \HOLSymConst{\HOLTokenBackslash} \HOLBoundVar{A}))) \HOLSymConst{\HOLTokenImp{}}
      \HOLConst{derivOne} (\HOLConst{Unf} (\HOLConst{Sequent} \HOLBoundVar{E} \HOLBoundVar{Gamma\sp{\prime}} \HOLBoundVar{C}))
        (\HOLConst{Der} (\HOLConst{Sequent} \HOLBoundVar{E} \HOLBoundVar{Gamma\sp{\prime}} \HOLBoundVar{C}) \HOLConst{LeftBackslash}
           [\HOLConst{Unf} (\HOLConst{Sequent} \HOLBoundVar{E} \HOLBoundVar{Gamma} \HOLBoundVar{C});
            \HOLConst{Unf} (\HOLConst{Sequent} \HOLBoundVar{E} \HOLBoundVar{Delta} \HOLBoundVar{B})])) \HOLSymConst{\HOLTokenConj{}}
   (\HOLSymConst{\HOLTokenForall{}}\HOLBoundVar{E} \HOLBoundVar{Gamma} \HOLBoundVar{Gamma\sp{\prime}} \HOLBoundVar{A} \HOLBoundVar{B} \HOLBoundVar{C}.
      \HOLConst{replace} \HOLBoundVar{Gamma} \HOLBoundVar{Gamma\sp{\prime}} (\HOLConst{Comma} (\HOLConst{OneForm} \HOLBoundVar{A}) (\HOLConst{OneForm} \HOLBoundVar{B}))
        (\HOLConst{OneForm} (\HOLBoundVar{A} \HOLSymConst{\HOLTokenProd{}} \HOLBoundVar{B})) \HOLSymConst{\HOLTokenImp{}}
      \HOLConst{derivOne} (\HOLConst{Unf} (\HOLConst{Sequent} \HOLBoundVar{E} \HOLBoundVar{Gamma\sp{\prime}} \HOLBoundVar{C}))
        (\HOLConst{Der} (\HOLConst{Sequent} \HOLBoundVar{E} \HOLBoundVar{Gamma\sp{\prime}} \HOLBoundVar{C}) \HOLConst{LeftDot}
           [\HOLConst{Unf} (\HOLConst{Sequent} \HOLBoundVar{E} \HOLBoundVar{Gamma} \HOLBoundVar{C})])) \HOLSymConst{\HOLTokenConj{}}
   (\HOLSymConst{\HOLTokenForall{}}\HOLBoundVar{E} \HOLBoundVar{Delta} \HOLBoundVar{Gamma} \HOLBoundVar{Gamma\sp{\prime}} \HOLBoundVar{A} \HOLBoundVar{C}.
      \HOLConst{replace} \HOLBoundVar{Gamma} \HOLBoundVar{Gamma\sp{\prime}} (\HOLConst{OneForm} \HOLBoundVar{A}) \HOLBoundVar{Delta} \HOLSymConst{\HOLTokenImp{}}
      \HOLConst{derivOne} (\HOLConst{Unf} (\HOLConst{Sequent} \HOLBoundVar{E} \HOLBoundVar{Gamma\sp{\prime}} \HOLBoundVar{C}))
        (\HOLConst{Der} (\HOLConst{Sequent} \HOLBoundVar{E} \HOLBoundVar{Gamma\sp{\prime}} \HOLBoundVar{C}) \HOLConst{CutRule}
           [\HOLConst{Unf} (\HOLConst{Sequent} \HOLBoundVar{E} \HOLBoundVar{Gamma} \HOLBoundVar{C});
            \HOLConst{Unf} (\HOLConst{Sequent} \HOLBoundVar{E} \HOLBoundVar{Delta} \HOLBoundVar{A})])) \HOLSymConst{\HOLTokenConj{}}
   \HOLSymConst{\HOLTokenForall{}}\HOLBoundVar{E} \HOLBoundVar{Gamma} \HOLBoundVar{Gamma\sp{\prime}} \HOLBoundVar{Delta} \HOLBoundVar{Delta\sp{\prime}} \HOLBoundVar{C}.
     \HOLConst{replace} \HOLBoundVar{Gamma} \HOLBoundVar{Gamma\sp{\prime}} \HOLBoundVar{Delta} \HOLBoundVar{Delta\sp{\prime}} \HOLSymConst{\HOLTokenConj{}} \HOLBoundVar{E} \HOLBoundVar{Delta} \HOLBoundVar{Delta\sp{\prime}} \HOLSymConst{\HOLTokenImp{}}
     \HOLConst{derivOne} (\HOLConst{Unf} (\HOLConst{Sequent} \HOLBoundVar{E} \HOLBoundVar{Gamma\sp{\prime}} \HOLBoundVar{C}))
       (\HOLConst{Der} (\HOLConst{Sequent} \HOLBoundVar{E} \HOLBoundVar{Gamma\sp{\prime}} \HOLBoundVar{C}) \HOLConst{SeqExt}
          [\HOLConst{Unf} (\HOLConst{Sequent} \HOLBoundVar{E} \HOLBoundVar{Gamma} \HOLBoundVar{C})])
\end{alltt}

The second step is to define ``structure rules'' as a new inductive
relation \HOLinline{\HOLConst{deriv}} based on \HOLinline{\HOLConst{derivOne}}. With this relation,
one can repeatly apply one-step derivation for all the unfinished
sub-proofs in the derivation tree, until all leaves are finished:
\begin{alltt}
\HOLTokenTurnstile{} (\HOLSymConst{\HOLTokenForall{}}\HOLBoundVar{D\sb{\mathrm{1}}} \HOLBoundVar{D\sb{\mathrm{2}}}. \HOLConst{derivOne} \HOLBoundVar{D\sb{\mathrm{1}}} \HOLBoundVar{D\sb{\mathrm{2}}} \HOLSymConst{\HOLTokenImp{}} \HOLConst{deriv} \HOLBoundVar{D\sb{\mathrm{1}}} \HOLBoundVar{D\sb{\mathrm{2}}}) \HOLSymConst{\HOLTokenConj{}}
   (\HOLSymConst{\HOLTokenForall{}}\HOLBoundVar{S} \HOLBoundVar{R} \HOLBoundVar{D\sb{\mathrm{1}}} \HOLBoundVar{D\sb{\mathrm{1}}\sp{\prime}}.
      \HOLConst{deriv} \HOLBoundVar{D\sb{\mathrm{1}}} \HOLBoundVar{D\sb{\mathrm{1}}\sp{\prime}} \HOLSymConst{\HOLTokenImp{}} \HOLConst{deriv} (\HOLConst{Der} \HOLBoundVar{S} \HOLBoundVar{R} [\HOLBoundVar{D\sb{\mathrm{1}}}]) (\HOLConst{Der} \HOLBoundVar{S} \HOLBoundVar{R} [\HOLBoundVar{D\sb{\mathrm{1}}\sp{\prime}}])) \HOLSymConst{\HOLTokenConj{}}
   (\HOLSymConst{\HOLTokenForall{}}\HOLBoundVar{S} \HOLBoundVar{R} \HOLBoundVar{D\sb{\mathrm{1}}} \HOLBoundVar{D\sb{\mathrm{1}}\sp{\prime}} \HOLBoundVar{D}.
      \HOLConst{deriv} \HOLBoundVar{D\sb{\mathrm{1}}} \HOLBoundVar{D\sb{\mathrm{1}}\sp{\prime}} \HOLSymConst{\HOLTokenImp{}}
      \HOLConst{deriv} (\HOLConst{Der} \HOLBoundVar{S} \HOLBoundVar{R} [\HOLBoundVar{D\sb{\mathrm{1}}}; \HOLBoundVar{D}]) (\HOLConst{Der} \HOLBoundVar{S} \HOLBoundVar{R} [\HOLBoundVar{D\sb{\mathrm{1}}\sp{\prime}}; \HOLBoundVar{D}])) \HOLSymConst{\HOLTokenConj{}}
   (\HOLSymConst{\HOLTokenForall{}}\HOLBoundVar{S} \HOLBoundVar{R} \HOLBoundVar{D\sb{\mathrm{2}}} \HOLBoundVar{D\sb{\mathrm{2}}\sp{\prime}} \HOLBoundVar{D}.
      \HOLConst{deriv} \HOLBoundVar{D\sb{\mathrm{2}}} \HOLBoundVar{D\sb{\mathrm{2}}\sp{\prime}} \HOLSymConst{\HOLTokenImp{}}
      \HOLConst{deriv} (\HOLConst{Der} \HOLBoundVar{S} \HOLBoundVar{R} [\HOLBoundVar{D}; \HOLBoundVar{D\sb{\mathrm{2}}}]) (\HOLConst{Der} \HOLBoundVar{S} \HOLBoundVar{R} [\HOLBoundVar{D}; \HOLBoundVar{D\sb{\mathrm{2}}\sp{\prime}}])) \HOLSymConst{\HOLTokenConj{}}
   \HOLSymConst{\HOLTokenForall{}}\HOLBoundVar{S} \HOLBoundVar{R} \HOLBoundVar{D\sb{\mathrm{1}}} \HOLBoundVar{D\sb{\mathrm{1}}\sp{\prime}} \HOLBoundVar{D\sb{\mathrm{2}}} \HOLBoundVar{D\sb{\mathrm{2}}\sp{\prime}}.
     \HOLConst{deriv} \HOLBoundVar{D\sb{\mathrm{1}}} \HOLBoundVar{D\sb{\mathrm{1}}\sp{\prime}} \HOLSymConst{\HOLTokenConj{}} \HOLConst{deriv} \HOLBoundVar{D\sb{\mathrm{2}}} \HOLBoundVar{D\sb{\mathrm{2}}\sp{\prime}} \HOLSymConst{\HOLTokenImp{}}
     \HOLConst{deriv} (\HOLConst{Der} \HOLBoundVar{S} \HOLBoundVar{R} [\HOLBoundVar{D\sb{\mathrm{1}}}; \HOLBoundVar{D\sb{\mathrm{2}}}]) (\HOLConst{Der} \HOLBoundVar{S} \HOLBoundVar{R} [\HOLBoundVar{D\sb{\mathrm{1}}\sp{\prime}}; \HOLBoundVar{D\sb{\mathrm{2}}\sp{\prime}}])
\end{alltt}

The last step is to define the chain of derivations as a reduction, so
that any two derivation trees are related with each other. This is
done by defining a new relation \HOLinline{\HOLConst{Deriv}} as the reflexitive
transitive closure (RTC) of \HOLinline{\HOLConst{deriv}}:
\begin{alltt}
\HOLTokenTurnstile{} \HOLConst{Deriv} \HOLSymConst{=} \HOLConst{deriv}\HOLSymConst{\HOLTokenSupStar{}}
\end{alltt}

Merging all above definitions together, we have proved all needed
theorems for contructing a whole new proof tree, either manually or automatically:
\begin{alltt}
(One step rules)
DerivSeqAxiom:
\HOLTokenTurnstile{} \HOLConst{Deriv} (\HOLConst{Unf} (\HOLConst{Sequent} \HOLFreeVar{E} (\HOLConst{OneForm} \HOLFreeVar{A}) \HOLFreeVar{A}))
     (\HOLConst{Der} (\HOLConst{Sequent} \HOLFreeVar{E} (\HOLConst{OneForm} \HOLFreeVar{A}) \HOLFreeVar{A}) \HOLConst{SeqAxiom} [])
DerivRightSlash:
\HOLTokenTurnstile{} \HOLConst{Deriv} (\HOLConst{Unf} (\HOLConst{Sequent} \HOLFreeVar{E} \HOLFreeVar{Gamma} (\HOLFreeVar{A} \HOLSymConst{/} \HOLFreeVar{B})))
     (\HOLConst{Der} (\HOLConst{Sequent} \HOLFreeVar{E} \HOLFreeVar{Gamma} (\HOLFreeVar{A} \HOLSymConst{/} \HOLFreeVar{B})) \HOLConst{RightSlash}
        [\HOLConst{Unf} (\HOLConst{Sequent} \HOLFreeVar{E} (\HOLConst{Comma} \HOLFreeVar{Gamma} (\HOLConst{OneForm} \HOLFreeVar{B})) \HOLFreeVar{A})])
DerivRightBackslash:
\HOLTokenTurnstile{} \HOLConst{Deriv} (\HOLConst{Unf} (\HOLConst{Sequent} \HOLFreeVar{E} \HOLFreeVar{Gamma} (\HOLFreeVar{B} \HOLSymConst{\HOLTokenBackslash} \HOLFreeVar{A})))
     (\HOLConst{Der} (\HOLConst{Sequent} \HOLFreeVar{E} \HOLFreeVar{Gamma} (\HOLFreeVar{B} \HOLSymConst{\HOLTokenBackslash} \HOLFreeVar{A})) \HOLConst{RightBackslash}
        [\HOLConst{Unf} (\HOLConst{Sequent} \HOLFreeVar{E} (\HOLConst{Comma} (\HOLConst{OneForm} \HOLFreeVar{B}) \HOLFreeVar{Gamma}) \HOLFreeVar{A})])
DerivRightDot:
\HOLTokenTurnstile{} \HOLConst{Deriv} (\HOLConst{Unf} (\HOLConst{Sequent} \HOLFreeVar{E} (\HOLConst{Comma} \HOLFreeVar{Gamma} \HOLFreeVar{Delta}) (\HOLFreeVar{A} \HOLSymConst{\HOLTokenProd{}} \HOLFreeVar{B})))
     (\HOLConst{Der} (\HOLConst{Sequent} \HOLFreeVar{E} (\HOLConst{Comma} \HOLFreeVar{Gamma} \HOLFreeVar{Delta}) (\HOLFreeVar{A} \HOLSymConst{\HOLTokenProd{}} \HOLFreeVar{B})) \HOLConst{RightDot}
        [\HOLConst{Unf} (\HOLConst{Sequent} \HOLFreeVar{E} \HOLFreeVar{Gamma} \HOLFreeVar{A}); \HOLConst{Unf} (\HOLConst{Sequent} \HOLFreeVar{E} \HOLFreeVar{Delta} \HOLFreeVar{B})])
DerivLeftSlash:
\HOLTokenTurnstile{} \HOLConst{replace} \HOLFreeVar{Gamma} \HOLFreeVar{Gamma\sp{\prime}} (\HOLConst{OneForm} \HOLFreeVar{A})
     (\HOLConst{Comma} (\HOLConst{OneForm} (\HOLFreeVar{A} \HOLSymConst{/} \HOLFreeVar{B})) \HOLFreeVar{Delta}) \HOLSymConst{\HOLTokenImp{}}
   \HOLConst{Deriv} (\HOLConst{Unf} (\HOLConst{Sequent} \HOLFreeVar{E} \HOLFreeVar{Gamma\sp{\prime}} \HOLFreeVar{C}))
     (\HOLConst{Der} (\HOLConst{Sequent} \HOLFreeVar{E} \HOLFreeVar{Gamma\sp{\prime}} \HOLFreeVar{C}) \HOLConst{LeftSlash}
        [\HOLConst{Unf} (\HOLConst{Sequent} \HOLFreeVar{E} \HOLFreeVar{Gamma} \HOLFreeVar{C}); \HOLConst{Unf} (\HOLConst{Sequent} \HOLFreeVar{E} \HOLFreeVar{Delta} \HOLFreeVar{B})])
DerivLeftBackslash:
\HOLTokenTurnstile{} \HOLConst{replace} \HOLFreeVar{Gamma} \HOLFreeVar{Gamma\sp{\prime}} (\HOLConst{OneForm} \HOLFreeVar{A})
     (\HOLConst{Comma} \HOLFreeVar{Delta} (\HOLConst{OneForm} (\HOLFreeVar{B} \HOLSymConst{\HOLTokenBackslash} \HOLFreeVar{A}))) \HOLSymConst{\HOLTokenImp{}}
   \HOLConst{Deriv} (\HOLConst{Unf} (\HOLConst{Sequent} \HOLFreeVar{E} \HOLFreeVar{Gamma\sp{\prime}} \HOLFreeVar{C}))
     (\HOLConst{Der} (\HOLConst{Sequent} \HOLFreeVar{E} \HOLFreeVar{Gamma\sp{\prime}} \HOLFreeVar{C}) \HOLConst{LeftBackslash}
        [\HOLConst{Unf} (\HOLConst{Sequent} \HOLFreeVar{E} \HOLFreeVar{Gamma} \HOLFreeVar{C}); \HOLConst{Unf} (\HOLConst{Sequent} \HOLFreeVar{E} \HOLFreeVar{Delta} \HOLFreeVar{B})])
DerivLeftDot:
\HOLTokenTurnstile{} \HOLConst{replace} \HOLFreeVar{Gamma} \HOLFreeVar{Gamma\sp{\prime}} (\HOLConst{Comma} (\HOLConst{OneForm} \HOLFreeVar{A}) (\HOLConst{OneForm} \HOLFreeVar{B}))
     (\HOLConst{OneForm} (\HOLFreeVar{A} \HOLSymConst{\HOLTokenProd{}} \HOLFreeVar{B})) \HOLSymConst{\HOLTokenImp{}}
   \HOLConst{Deriv} (\HOLConst{Unf} (\HOLConst{Sequent} \HOLFreeVar{E} \HOLFreeVar{Gamma\sp{\prime}} \HOLFreeVar{C}))
     (\HOLConst{Der} (\HOLConst{Sequent} \HOLFreeVar{E} \HOLFreeVar{Gamma\sp{\prime}} \HOLFreeVar{C}) \HOLConst{LeftDot}
        [\HOLConst{Unf} (\HOLConst{Sequent} \HOLFreeVar{E} \HOLFreeVar{Gamma} \HOLFreeVar{C})])
DerivCutRule:
\HOLTokenTurnstile{} \HOLConst{replace} \HOLFreeVar{Gamma} \HOLFreeVar{Gamma\sp{\prime}} (\HOLConst{OneForm} \HOLFreeVar{A}) \HOLFreeVar{Delta} \HOLSymConst{\HOLTokenImp{}}
   \HOLConst{Deriv} (\HOLConst{Unf} (\HOLConst{Sequent} \HOLFreeVar{E} \HOLFreeVar{Gamma\sp{\prime}} \HOLFreeVar{C}))
     (\HOLConst{Der} (\HOLConst{Sequent} \HOLFreeVar{E} \HOLFreeVar{Gamma\sp{\prime}} \HOLFreeVar{C}) \HOLConst{CutRule}
        [\HOLConst{Unf} (\HOLConst{Sequent} \HOLFreeVar{E} \HOLFreeVar{Gamma} \HOLFreeVar{C}); \HOLConst{Unf} (\HOLConst{Sequent} \HOLFreeVar{E} \HOLFreeVar{Delta} \HOLFreeVar{A})])
DerivSeqExt:
\HOLTokenTurnstile{} \HOLConst{replace} \HOLFreeVar{Gamma} \HOLFreeVar{Gamma\sp{\prime}} \HOLFreeVar{Delta} \HOLFreeVar{Delta\sp{\prime}} \HOLSymConst{\HOLTokenConj{}} \HOLFreeVar{E} \HOLFreeVar{Delta} \HOLFreeVar{Delta\sp{\prime}} \HOLSymConst{\HOLTokenImp{}}
   \HOLConst{Deriv} (\HOLConst{Unf} (\HOLConst{Sequent} \HOLFreeVar{E} \HOLFreeVar{Gamma\sp{\prime}} \HOLFreeVar{C}))
     (\HOLConst{Der} (\HOLConst{Sequent} \HOLFreeVar{E} \HOLFreeVar{Gamma\sp{\prime}} \HOLFreeVar{C}) \HOLConst{SeqExt}
        [\HOLConst{Unf} (\HOLConst{Sequent} \HOLFreeVar{E} \HOLFreeVar{Gamma} \HOLFreeVar{C})])

(Structure rules)
DerivOne:
\HOLTokenTurnstile{} \HOLConst{Deriv} \HOLFreeVar{D\sb{\mathrm{1}}} \HOLFreeVar{D\sb{\mathrm{1}}\sp{\prime}} \HOLSymConst{\HOLTokenImp{}} \HOLConst{Deriv} (\HOLConst{Der} \HOLFreeVar{S} \HOLFreeVar{R} [\HOLFreeVar{D\sb{\mathrm{1}}}]) (\HOLConst{Der} \HOLFreeVar{S} \HOLFreeVar{R} [\HOLFreeVar{D\sb{\mathrm{1}}\sp{\prime}}])
DerivLeft:
\HOLTokenTurnstile{} \HOLConst{Deriv} \HOLFreeVar{D\sb{\mathrm{1}}} \HOLFreeVar{D\sb{\mathrm{1}}\sp{\prime}} \HOLSymConst{\HOLTokenImp{}} \HOLConst{Deriv} (\HOLConst{Der} \HOLFreeVar{S} \HOLFreeVar{R} [\HOLFreeVar{D\sb{\mathrm{1}}}; \HOLFreeVar{D}]) (\HOLConst{Der} \HOLFreeVar{S} \HOLFreeVar{R} [\HOLFreeVar{D\sb{\mathrm{1}}\sp{\prime}}; \HOLFreeVar{D}])
DerivRight:
\HOLTokenTurnstile{} \HOLConst{Deriv} \HOLFreeVar{D\sb{\mathrm{2}}} \HOLFreeVar{D\sb{\mathrm{2}}\sp{\prime}} \HOLSymConst{\HOLTokenImp{}} \HOLConst{Deriv} (\HOLConst{Der} \HOLFreeVar{S} \HOLFreeVar{R} [\HOLFreeVar{D}; \HOLFreeVar{D\sb{\mathrm{2}}}]) (\HOLConst{Der} \HOLFreeVar{S} \HOLFreeVar{R} [\HOLFreeVar{D}; \HOLFreeVar{D\sb{\mathrm{2}}\sp{\prime}}])
DerivBoth:
\HOLTokenTurnstile{} \HOLConst{Deriv} \HOLFreeVar{D\sb{\mathrm{1}}} \HOLFreeVar{D\sb{\mathrm{1}}\sp{\prime}} \HOLSymConst{\HOLTokenImp{}}
   \HOLConst{Deriv} \HOLFreeVar{D\sb{\mathrm{2}}} \HOLFreeVar{D\sb{\mathrm{2}}\sp{\prime}} \HOLSymConst{\HOLTokenImp{}}
   \HOLConst{Deriv} (\HOLConst{Der} \HOLFreeVar{S} \HOLFreeVar{R} [\HOLFreeVar{D\sb{\mathrm{1}}}; \HOLFreeVar{D\sb{\mathrm{2}}}]) (\HOLConst{Der} \HOLFreeVar{S} \HOLFreeVar{R} [\HOLFreeVar{D\sb{\mathrm{1}}\sp{\prime}}; \HOLFreeVar{D\sb{\mathrm{2}}\sp{\prime}}])

(RTC rules)
Deriv_refl:
\HOLTokenTurnstile{} \HOLConst{Deriv} \HOLFreeVar{x} \HOLFreeVar{x}
Deriv_trans:
\HOLTokenTurnstile{} \HOLConst{Deriv} \HOLFreeVar{x} \HOLFreeVar{y} \HOLSymConst{\HOLTokenConj{}} \HOLConst{Deriv} \HOLFreeVar{y} \HOLFreeVar{z} \HOLSymConst{\HOLTokenImp{}} \HOLConst{Deriv} \HOLFreeVar{x} \HOLFreeVar{z}
\end{alltt}

A \texttt{Dertree} is a proof if all its leaves are finished
sub-proofs. It has to be defined as another inductive unary relation:
\begin{alltt}
\HOLTokenTurnstile{} (\HOLSymConst{\HOLTokenForall{}}\HOLBoundVar{S} \HOLBoundVar{R}. \HOLConst{Proof} (\HOLConst{Der} \HOLBoundVar{S} \HOLBoundVar{R} [])) \HOLSymConst{\HOLTokenConj{}}
   (\HOLSymConst{\HOLTokenForall{}}\HOLBoundVar{S} \HOLBoundVar{R} \HOLBoundVar{D}. \HOLConst{Proof} \HOLBoundVar{D} \HOLSymConst{\HOLTokenImp{}} \HOLConst{Proof} (\HOLConst{Der} \HOLBoundVar{S} \HOLBoundVar{R} [\HOLBoundVar{D}])) \HOLSymConst{\HOLTokenConj{}}
   \HOLSymConst{\HOLTokenForall{}}\HOLBoundVar{S} \HOLBoundVar{R} \HOLBoundVar{D\sb{\mathrm{1}}} \HOLBoundVar{D\sb{\mathrm{2}}}. \HOLConst{Proof} \HOLBoundVar{D\sb{\mathrm{1}}} \HOLSymConst{\HOLTokenConj{}} \HOLConst{Proof} \HOLBoundVar{D\sb{\mathrm{2}}} \HOLSymConst{\HOLTokenImp{}} \HOLConst{Proof} (\HOLConst{Der} \HOLBoundVar{S} \HOLBoundVar{R} [\HOLBoundVar{D\sb{\mathrm{1}}}; \HOLBoundVar{D\sb{\mathrm{2}}}])
\end{alltt}

Finally we have proved the following theorem which partially
guaranteed the correctness of \HOLinline{\HOLConst{Deriv}} rules. It says for each
true sequent, there's a finished proof tree which is derivable from
a unfinished Dertree having that sequent as head:\footnote{The other
  direction remains unproved: for any sequent, if it leads to a finished Dertree,
  then it must be a true sequent. We leave this difficult theorem to
  future work.}
\begin{alltt}
gentzenToDeriv:
\HOLTokenTurnstile{} \HOLConst{gentzenSequent} \HOLFreeVar{E} \HOLFreeVar{Gamma} \HOLFreeVar{A} \HOLSymConst{\HOLTokenImp{}}
   \HOLSymConst{\HOLTokenExists{}}\HOLBoundVar{D}. \HOLConst{Deriv} (\HOLConst{Unf} (\HOLConst{Sequent} \HOLFreeVar{E} \HOLFreeVar{Gamma} \HOLFreeVar{A})) \HOLBoundVar{D} \HOLSymConst{\HOLTokenConj{}} \HOLConst{Proof} \HOLBoundVar{D}
\end{alltt}

\subsection{Cut-free proofs}

In Gentzen's original Sequent Calculus for intuitionistic propositional logic, the so-called
\emph{cut-elimination theorem} (Hauptsatz) stands at the central
position. Cut-elimination theorem for Lambek Calculus was proved by
Lambek for \textbf{L} \cite{Lambek:1958fh} and \textbf{NL}
\cite{Lambek:1961bx}.

Roughly speaking, this important theorem said that the Cut
rule is admissible. In other words, the Cut rule doesn't increase the
set of theorems provable from other rules. It's this theorem which guaranteed the existence of decision
procedures, because all other rules has the sub-formula property,
which is essential for automatic proof searching.

In our project, due to the complexity and large amount of preparation
before reaching the Cut-elimination theorem, it's not formally proved
yet. Nor the original Coq work has done this proof.\footnote{to our
  knowledge, the Cut-elimination theorem for Lambek calculus is
  never formally verified. This topic along may become another paper
  after this project, since it's big enough.}

A \emph{cut-free} proof is a proof (Dertree) without using the
\texttt{CutRule} of gentzen's Sequent Calculus. One way to define
cut-free proofs is simply through the \HOLinline{\HOLConst{degreeProof}} property:
\begin{alltt}
\HOLTokenTurnstile{} \HOLConst{CutFreeProof} \HOLFreeVar{p} \HOLSymConst{\HOLTokenEquiv{}} (\HOLConst{degreeProof} \HOLFreeVar{p} \HOLSymConst{=} \HOLNumLit{0})
\end{alltt}
This is because, in the definition of \HOLinline{\HOLConst{degreeProof}}, only the
\texttt{CutRule} has a non-zero degree, while all other rules has zero
degrees. So if the entire Dertree has zero degree, it must not contain
any CutRule. Saying the same thing in another way, that's the
following theorem:
\begin{alltt}
notCutFree:
\HOLTokenTurnstile{} \HOLConst{replace} \HOLFreeVar{T\sb{\mathrm{1}}} \HOLFreeVar{T\sb{\mathrm{2}}} (\HOLConst{OneForm} \HOLFreeVar{A}) \HOLFreeVar{D} \HOLSymConst{\HOLTokenConj{}} (\HOLFreeVar{p\sb{\mathrm{1}}} \HOLSymConst{=} \HOLConst{Sequent} \HOLFreeVar{E} \HOLFreeVar{D} \HOLFreeVar{A}) \HOLSymConst{\HOLTokenConj{}}
   (\HOLFreeVar{p\sb{\mathrm{2}}} \HOLSymConst{=} \HOLConst{Sequent} \HOLFreeVar{E} \HOLFreeVar{T\sb{\mathrm{1}}} \HOLFreeVar{C}) \HOLSymConst{\HOLTokenImp{}}
   \HOLSymConst{\HOLTokenNeg{}}\HOLConst{CutFreeProof} (\HOLConst{Der} \HOLFreeVar{\HOLTokenUnderscore{}} \HOLConst{CutRule} [\HOLConst{Der} \HOLFreeVar{p\sb{\mathrm{1}}} \HOLFreeVar{\HOLTokenUnderscore{}} \HOLFreeVar{\HOLTokenUnderscore{}}; \HOLConst{Der} \HOLFreeVar{p\sb{\mathrm{2}}} \HOLFreeVar{\HOLTokenUnderscore{}} \HOLFreeVar{\HOLTokenUnderscore{}}])
\end{alltt}

The next important concept is \emph{sub-proof}. A proof $q$ is a
sub-proof of another proof $p$ if and only if $p$ can be found in one
leave of the Dertree of $p$. The one-step version of this relation is
defined as another inductive relation \HOLinline{\HOLConst{subProofOne}}. Here we
omitted its long definition but only show one example, the two LeftSlash
cases:
\begin{alltt}
ls1:
\HOLTokenTurnstile{} (\HOLFreeVar{p\sb{\mathrm{0}}} \HOLSymConst{=} \HOLConst{Sequent} \HOLFreeVar{E} \HOLFreeVar{Gamma\sp{\prime}} \HOLFreeVar{C}) \HOLSymConst{\HOLTokenConj{}}
   (\HOLFreeVar{p\sb{\mathrm{1}}} \HOLSymConst{=} \HOLConst{Der} (\HOLConst{Sequent} \HOLFreeVar{E} \HOLFreeVar{Delta} \HOLFreeVar{B}) \HOLFreeVar{R} \HOLFreeVar{D}) \HOLSymConst{\HOLTokenConj{}}
   (\HOLFreeVar{p\sb{\mathrm{2}}} \HOLSymConst{=} \HOLConst{Der} (\HOLConst{Sequent} \HOLFreeVar{E} \HOLFreeVar{Gamma} \HOLFreeVar{C}) \HOLFreeVar{R} \HOLFreeVar{D}) \HOLSymConst{\HOLTokenConj{}}
   \HOLConst{replace} \HOLFreeVar{Gamma} \HOLFreeVar{Gamma\sp{\prime}} (\HOLConst{OneForm} \HOLFreeVar{A})
     (\HOLConst{Comma} (\HOLConst{OneForm} (\HOLFreeVar{A} \HOLSymConst{/} \HOLFreeVar{B})) \HOLFreeVar{Delta}) \HOLSymConst{\HOLTokenImp{}}
   \HOLConst{subProofOne} \HOLFreeVar{p\sb{\mathrm{1}}} (\HOLConst{Der} \HOLFreeVar{p\sb{\mathrm{0}}} \HOLConst{LeftSlash} [\HOLFreeVar{p\sb{\mathrm{1}}}; \HOLFreeVar{p\sb{\mathrm{2}}}])
ls2:
\HOLTokenTurnstile{} (\HOLFreeVar{p\sb{\mathrm{0}}} \HOLSymConst{=} \HOLConst{Sequent} \HOLFreeVar{E} \HOLFreeVar{Gamma\sp{\prime}} \HOLFreeVar{C}) \HOLSymConst{\HOLTokenConj{}}
   (\HOLFreeVar{p\sb{\mathrm{1}}} \HOLSymConst{=} \HOLConst{Der} (\HOLConst{Sequent} \HOLFreeVar{E} \HOLFreeVar{Delta} \HOLFreeVar{B}) \HOLFreeVar{R} \HOLFreeVar{D}) \HOLSymConst{\HOLTokenConj{}}
   (\HOLFreeVar{p\sb{\mathrm{2}}} \HOLSymConst{=} \HOLConst{Der} (\HOLConst{Sequent} \HOLFreeVar{E} \HOLFreeVar{Gamma} \HOLFreeVar{C}) \HOLFreeVar{R} \HOLFreeVar{D}) \HOLSymConst{\HOLTokenConj{}}
   \HOLConst{replace} \HOLFreeVar{Gamma} \HOLFreeVar{Gamma\sp{\prime}} (\HOLConst{OneForm} \HOLFreeVar{A})
     (\HOLConst{Comma} (\HOLConst{OneForm} (\HOLFreeVar{A} \HOLSymConst{/} \HOLFreeVar{B})) \HOLFreeVar{Delta}) \HOLSymConst{\HOLTokenImp{}}
   \HOLConst{subProofOne} \HOLFreeVar{p\sb{\mathrm{2}}} (\HOLConst{Der} \HOLFreeVar{p\sb{\mathrm{0}}} \HOLConst{LeftSlash} [\HOLFreeVar{p\sb{\mathrm{1}}}; \HOLFreeVar{p\sb{\mathrm{2}}}])
\end{alltt}
Based on \HOLinline{\HOLConst{subProofOne}}, now the full version \HOLinline{\HOLConst{subProof}} is
just a RTC of \HOLinline{\HOLConst{subProofOne}}: (there's no structure rules here)
\begin{alltt}
\HOLTokenTurnstile{} \HOLConst{subProof} \HOLSymConst{=} \HOLConst{subProofOne}\HOLSymConst{\HOLTokenSupStar{}}
\end{alltt}

Finally we have proved an important sub-formula property if we
already have a Cut-free proof. The one-step version (\texttt{subFormulaPropertyOne}) is the longest
proofs we met in the whole project, the full version is then provable
by doing induction on the one-step version.
\begin{alltt}
subFormulaPropertyOne:
\HOLTokenTurnstile{} \HOLConst{subProofOne} \HOLFreeVar{q} \HOLFreeVar{p} \HOLSymConst{\HOLTokenImp{}}
   \HOLConst{extensionSub} (\HOLConst{exten} \HOLFreeVar{p}) \HOLSymConst{\HOLTokenImp{}}
   \HOLConst{CutFreeProof} \HOLFreeVar{p} \HOLSymConst{\HOLTokenImp{}}
   \HOLSymConst{\HOLTokenForall{}}\HOLBoundVar{x}.
     \HOLConst{subFormTerm} \HOLBoundVar{x} (\HOLConst{prems} \HOLFreeVar{q}) \HOLSymConst{\HOLTokenDisj{}} \HOLConst{subFormula} \HOLBoundVar{x} (\HOLConst{concl} \HOLFreeVar{q}) \HOLSymConst{\HOLTokenImp{}}
     \HOLConst{subFormTerm} \HOLBoundVar{x} (\HOLConst{prems} \HOLFreeVar{p}) \HOLSymConst{\HOLTokenDisj{}} \HOLConst{subFormula} \HOLBoundVar{x} (\HOLConst{concl} \HOLFreeVar{p})
subFormulaPropertyOne':
\HOLTokenTurnstile{} (\HOLFreeVar{p} \HOLSymConst{=} \HOLConst{Der} (\HOLConst{Sequent} \HOLFreeVar{E} \HOLFreeVar{Gamma\sb{\mathrm{1}}} \HOLFreeVar{B}) \HOLFreeVar{\HOLTokenUnderscore{}} \HOLFreeVar{\HOLTokenUnderscore{}}) \HOLSymConst{\HOLTokenImp{}}
   (\HOLFreeVar{q} \HOLSymConst{=} \HOLConst{Der} (\HOLConst{Sequent} \HOLFreeVar{E} \HOLFreeVar{Gamma\sb{\mathrm{2}}} \HOLFreeVar{C}) \HOLFreeVar{\HOLTokenUnderscore{}} \HOLFreeVar{\HOLTokenUnderscore{}}) \HOLSymConst{\HOLTokenImp{}}
   \HOLConst{extensionSub} \HOLFreeVar{E} \HOLSymConst{\HOLTokenImp{}}
   \HOLConst{subProofOne} \HOLFreeVar{q} \HOLFreeVar{p} \HOLSymConst{\HOLTokenImp{}}
   \HOLConst{CutFreeProof} \HOLFreeVar{p} \HOLSymConst{\HOLTokenImp{}}
   \HOLConst{subFormTerm} \HOLFreeVar{x} \HOLFreeVar{Gamma\sb{\mathrm{2}}} \HOLSymConst{\HOLTokenDisj{}} \HOLConst{subFormula} \HOLFreeVar{x} \HOLFreeVar{C} \HOLSymConst{\HOLTokenImp{}}
   \HOLConst{subFormTerm} \HOLFreeVar{x} \HOLFreeVar{Gamma\sb{\mathrm{1}}} \HOLSymConst{\HOLTokenDisj{}} \HOLConst{subFormula} \HOLFreeVar{x} \HOLFreeVar{B}
subFormulaProperty:
\HOLTokenTurnstile{} \HOLConst{subProof} \HOLFreeVar{q} \HOLFreeVar{p} \HOLSymConst{\HOLTokenImp{}}
   \HOLConst{extensionSub} (\HOLConst{exten} \HOLFreeVar{p}) \HOLSymConst{\HOLTokenImp{}}
   \HOLConst{CutFreeProof} \HOLFreeVar{p} \HOLSymConst{\HOLTokenImp{}}
   \HOLSymConst{\HOLTokenForall{}}\HOLBoundVar{x}.
     \HOLConst{subFormTerm} \HOLBoundVar{x} (\HOLConst{prems} \HOLFreeVar{q}) \HOLSymConst{\HOLTokenDisj{}} \HOLConst{subFormula} \HOLBoundVar{x} (\HOLConst{concl} \HOLFreeVar{q}) \HOLSymConst{\HOLTokenImp{}}
     \HOLConst{subFormTerm} \HOLBoundVar{x} (\HOLConst{prems} \HOLFreeVar{p}) \HOLSymConst{\HOLTokenDisj{}} \HOLConst{subFormula} \HOLBoundVar{x} (\HOLConst{concl} \HOLFreeVar{p})
subFormulaProperty':
\HOLTokenTurnstile{} (\HOLFreeVar{p} \HOLSymConst{=} \HOLConst{Der} (\HOLConst{Sequent} \HOLFreeVar{E} \HOLFreeVar{Gamma\sb{\mathrm{1}}} \HOLFreeVar{B}) \HOLFreeVar{\HOLTokenUnderscore{}} \HOLFreeVar{\HOLTokenUnderscore{}}) \HOLSymConst{\HOLTokenImp{}}
   (\HOLFreeVar{q} \HOLSymConst{=} \HOLConst{Der} (\HOLConst{Sequent} \HOLFreeVar{E} \HOLFreeVar{Gamma\sb{\mathrm{2}}} \HOLFreeVar{C}) \HOLFreeVar{\HOLTokenUnderscore{}} \HOLFreeVar{\HOLTokenUnderscore{}}) \HOLSymConst{\HOLTokenImp{}}
   \HOLConst{extensionSub} \HOLFreeVar{E} \HOLSymConst{\HOLTokenImp{}}
   \HOLConst{subProof} \HOLFreeVar{q} \HOLFreeVar{p} \HOLSymConst{\HOLTokenImp{}}
   \HOLConst{CutFreeProof} \HOLFreeVar{p} \HOLSymConst{\HOLTokenImp{}}
   \HOLConst{subFormTerm} \HOLFreeVar{x} \HOLFreeVar{Gamma\sb{\mathrm{2}}} \HOLSymConst{\HOLTokenDisj{}} \HOLConst{subFormula} \HOLFreeVar{x} \HOLFreeVar{C} \HOLSymConst{\HOLTokenImp{}}
   \HOLConst{subFormTerm} \HOLFreeVar{x} \HOLFreeVar{Gamma\sb{\mathrm{1}}} \HOLSymConst{\HOLTokenDisj{}} \HOLConst{subFormula} \HOLFreeVar{x} \HOLFreeVar{B}
\end{alltt}
Basically these theorems said, if we have a cut-free proof, then any
sub-formula in its sub-proofs (either in antecedents or conclusion of
the sequent) is also a sub-formula of the sequent for whole proof.  In
another word, NO new formula appears during the proof searching
process, or NO need to guess anything!   So, if the cut-free proof
indeed exists for \emph{every sequent proof using CutRule}, then we do have the decision
procedure!

The last thing to explain, is the meaning of \HOLinline{\HOLConst{extensionSub}}
appearing in above theorems. The purpose is to make sure the gentzen
extension in the related Lambek Calculus has also the sub-formula property:
\begin{alltt}
\HOLTokenTurnstile{} \HOLConst{extensionSub} \HOLFreeVar{E} \HOLSymConst{\HOLTokenEquiv{}}
   \HOLSymConst{\HOLTokenForall{}}\HOLBoundVar{Form} \HOLBoundVar{T\sb{\mathrm{1}}} \HOLBoundVar{T\sb{\mathrm{2}}}.
     \HOLFreeVar{E} \HOLBoundVar{T\sb{\mathrm{1}}} \HOLBoundVar{T\sb{\mathrm{2}}} \HOLSymConst{\HOLTokenImp{}} \HOLConst{subFormTerm} \HOLBoundVar{Form} \HOLBoundVar{T\sb{\mathrm{1}}} \HOLSymConst{\HOLTokenImp{}} \HOLConst{subFormTerm} \HOLBoundVar{Form} \HOLBoundVar{T\sb{\mathrm{2}}}
\end{alltt}

\section{Examples}

Our first example demonstrates how to use the Lambek Calculus
formulization as a toolkit for manual proving derivations about
categories of sentences. Suppose we have the following minimal Italian lexicon:
\begin{quote}
\begin{tabular}{|c|c|}
\hline
\textbf{word} & \textbf{category} \\
\hline
cosa & $S/(S/np)$ \\
guarda & $S/inf$ \\
passare & $inf/np$ \\
\hline
\end{tabular}
\end{quote}
And the goal is to check if ``cosa guarda passare'' is an Italian
sentance (and then parse it).  There're only two ways to bracketing the
three words ``cosa'', ``guarda'' and ``passare'':
\begin{enumerate}
\item ((``cosa'', ``guarda''), ``passare'');
\item (``cosa'', (``guarda'', ``passare'')).
\end{enumerate}
The first way leads to nothing, while the second way (as also a parsing tree) can be proved to
have the category $S$ in either Natural Deduction or Sequent Calculus:
\begin{alltt}
\HOLTokenTurnstile{} \HOLConst{natDed} \HOLConst{L_Sequent}
     (\HOLConst{Comma} (\HOLConst{OneForm} (\HOLConst{At} \HOLStringLit{S} \HOLSymConst{/} (\HOLConst{At} \HOLStringLit{S} \HOLSymConst{/} \HOLConst{At} \HOLStringLit{np})))
        (\HOLConst{Comma} (\HOLConst{OneForm} (\HOLConst{At} \HOLStringLit{S} \HOLSymConst{/} \HOLConst{At} \HOLStringLit{inf}))
           (\HOLConst{OneForm} (\HOLConst{At} \HOLStringLit{inf} \HOLSymConst{/} \HOLConst{At} \HOLStringLit{np})))) (\HOLConst{At} \HOLStringLit{S})
\HOLTokenTurnstile{} \HOLConst{gentzenSequent} \HOLConst{L_Sequent}
     (\HOLConst{Comma} (\HOLConst{OneForm} (\HOLConst{At} \HOLStringLit{S} \HOLSymConst{/} (\HOLConst{At} \HOLStringLit{S} \HOLSymConst{/} \HOLConst{At} \HOLStringLit{np})))
        (\HOLConst{Comma} (\HOLConst{OneForm} (\HOLConst{At} \HOLStringLit{S} \HOLSymConst{/} \HOLConst{At} \HOLStringLit{inf}))
           (\HOLConst{OneForm} (\HOLConst{At} \HOLStringLit{inf} \HOLSymConst{/} \HOLConst{At} \HOLStringLit{np})))) (\HOLConst{At} \HOLStringLit{S})
\end{alltt}

Here is the proof tree in Natural Deduction
with \textbf{L} extension:
\begin{prooftree}
\AxiomC{$S/(S/np) \vdash S/(S/np)$}

\AxiomC{$S/inf \vdash S/inf$}

\AxiomC{$inf/np \vdash inf/np$}
\AxiomC{$np \vdash np$}
\RightLabel{$/_e$}
\BinaryInfC{$inf/np, np \vdash inf$}

\RightLabel{$/_e$}
\BinaryInfC{$S/inf, (inf/np, np) \vdash S$}

\RightLabel{L_Sequent}
\UnaryInfC{$(S/inf, inf/np), np \vdash S$}

\RightLabel{$/_i$}
\UnaryInfC{$S/inf, inf/np \vdash S/np$}

\RightLabel{$/_e$}
\BinaryInfC{$S/(S/np), (S/inf, inf/np) \vdash S$}

\RightLabel{Lex}
\UnaryInfC{$(\text{``cosa''}, (\text{``guarda''}, \text{``passare''})) \vdash S$}
\end{prooftree}
And the proof tree in Sequent Calculus with \textbf{L} extension:
\begin{prooftree}
\AxiomC{$S \vdash S$}

\AxiomC{$S \vdash S$}
\AxiomC{$inf \vdash inf$}
\RightLabel{$/_L$}
\BinaryInfC{$S/inf, inf \vdash S$}

\AxiomC{$np \vdash np$}
\RightLabel{$/_L$}
\BinaryInfC{$S/inf, (inf/np, np) \vdash S$}
\RightLabel{L_Sequent}
\UnaryInfC{$(S/inf, inf/np), np \vdash S$}
\RightLabel{$/_R$}
\UnaryInfC{$S/inf, inf/np \vdash S/np$}

\RightLabel{$/_L$}
\BinaryInfC{$S/(S/np), (S/inf, inf/np) \vdash S$}

\RightLabel{Lex}
\UnaryInfC{$(\text{``cosa''}, (\text{``guarda''}, \text{``passare''}))
  \vdash S$}
\end{prooftree}

If we compare the two proof trees, we can see that, the Sequent Calculus
proof is ``smaller'' in the sense that, all applications of
\texttt{SeqAxiom} are based on sub-formulae of lower level formulae in
the proof tree (in this case they're all basic categories, but it's
not always like this). To see the
advantages of Sequent Calculus more clearly, for the first time we
give the formal proof scripts in HOL4 for above two theorems:
\begin{lstlisting}
val cosa_guarda_passare_natDed = store_thm (
   "cosa_guarda_passare_natDed",
  ``natDed L_Sequent (Comma (OneForm ^cosa)
                            (Comma (OneForm ^guarda) (OneForm ^passare)))
		     (At "S")``,
    MATCH_MP_TAC SlashElim
 >> EXISTS_TAC ``(At "S") / (At "np")`` (* guess 1 *)
 >> CONJ_TAC (* 2 sub-goals here *)
 >| [ (* goal 1 *)
      REWRITE_TAC [NatAxiom],
      (* goal 2 *)
      MATCH_MP_TAC SlashIntro \\
      MATCH_MP_TAC NatExtSimpl \\
      EXISTS_TAC ``(Comma (OneForm (At "S" / At "inf"))
			  (Comma (OneForm (At "inf" / At "np"))
                                 (OneForm (At "np"))))`` \\
      CONJ_TAC >- REWRITE_TAC [L_Sequent_rules] \\
      MATCH_MP_TAC SlashElim \\
      EXISTS_TAC ``At "inf"`` \\ (* guess 2 *)
      CONJ_TAC >| (* 2 sub-goals here *)
      [ (* goal 2.1 *)
        REWRITE_TAC [NatAxiom],
        (* goal 2.2 *)
        MATCH_MP_TAC SlashElim \\
        EXISTS_TAC ``At "np"`` \\ (* guess 3 *)
        REWRITE_TAC [NatAxiom] ] ]);

val cosa_guarda_passare_gentzenSequent = store_thm (
   "cosa_guarda_passare_gentzenSequent",
  ``gentzenSequent L_Sequent (Comma (OneForm ^cosa)
                                    (Comma (OneForm ^guarda) (OneForm ^passare)))
			     (At "S")``,
    MATCH_MP_TAC LeftSlashSimpl
 >> CONJ_TAC (* 2 sub-goals here *)
 >| [ (* goal 1 *)
      MATCH_MP_TAC RightSlash \\
      MATCH_MP_TAC SeqExtSimpl \\
      EXISTS_TAC ``(Comma (OneForm (At "S" / At "inf"))
			  (Comma (OneForm (At "inf" / At "np"))
                                 (OneForm (At "np"))))`` \\
      CONJ_TAC >- REWRITE_TAC [L_Sequent_rules] \\
      MATCH_MP_TAC LeftSlashSimpl \\
      CONJ_TAC >| (* 2 sub-goals here *)
      [ (* goal 1.1 *)
        MATCH_MP_TAC LeftSlashSimpl \\
        REWRITE_TAC [SeqAxiom],
        (* goal 1.2 *)
        REWRITE_TAC [SeqAxiom] ],
      (* goal 2 *)
      REWRITE_TAC [SeqAxiom] ]);
\end{lstlisting}
In the first proof based on Natural Deduction, after rule applications of \texttt{SlashElim},
\texttt{SlashElim} and the second \texttt{SlashElim}, the parameter of
\texttt{EXISTS_TAC} tactical must be correctly guessed. Since there're
infinite possibilities, automatica proof searching will fail with
Natural Deduction rules.  But in the second proof based on Sequent
Calculus, there's no such guess at all. Automatic proof searching
algorithm could just try each possible rules (the number is finite)
and then does the same strategy for each searching branches, since the
formulae at next levels always become smaller, the proof searching
process will definitely terminate. (And there're even better algorithms
with polinomial time complexity)

The next example is to demonstrates how to manually construct a proof
tree object for above sentence and prove that the Dertree is indeed a valid proof.

At the beginning we have the following unfinished Dertree:
\begin{lstlisting}
val r0 =
  ``(Unf (Sequent L_Sequent (Comma (OneForm (At "S" / (At "S" / At "np")))
				   (Comma (OneForm (At "S" / At "inf"))
                                          (OneForm (At "inf" / At "np"))))
			    (At "S")))``;
\end{lstlisting}

If we try to manually expand this Dertree into the final proof tree
according to above manual proof, at the next step we could have a new
Dertree like this:
\begin{lstlisting}
val r1 =
  ``(Der (Sequent L_Sequent (Comma (OneForm (At "S" / (At "S" / At "np")))
				   (Comma (OneForm (At "S" / At "inf"))
                                          (OneForm (At "inf" / At "np"))))
			    (At "S"))
	 LeftSlash
      [ (Unf (Sequent L_Sequent (OneForm (At "S")) (At "S"))) ;
	(Unf (Sequent L_Sequent (Comma (OneForm (At "S" / At "inf"))
                                       (OneForm (At "inf" / At "np")))
				(At "S" / At "np"))) ])``;
\end{lstlisting}
And we can prove this new Dertree is derived from the last one:
\begin{alltt}
\HOLTokenTurnstile{} \HOLConst{Deriv}
     (\HOLConst{Unf}
        (\HOLConst{Sequent} \HOLConst{L_Sequent}
           (\HOLConst{Comma} (\HOLConst{OneForm} (\HOLConst{At} \HOLStringLit{S} \HOLSymConst{/} (\HOLConst{At} \HOLStringLit{S} \HOLSymConst{/} \HOLConst{At} \HOLStringLit{np})))
              (\HOLConst{Comma} (\HOLConst{OneForm} (\HOLConst{At} \HOLStringLit{S} \HOLSymConst{/} \HOLConst{At} \HOLStringLit{inf}))
                 (\HOLConst{OneForm} (\HOLConst{At} \HOLStringLit{inf} \HOLSymConst{/} \HOLConst{At} \HOLStringLit{np})))) (\HOLConst{At} \HOLStringLit{S})))
     (\HOLConst{Der}
        (\HOLConst{Sequent} \HOLConst{L_Sequent}
           (\HOLConst{Comma} (\HOLConst{OneForm} (\HOLConst{At} \HOLStringLit{S} \HOLSymConst{/} (\HOLConst{At} \HOLStringLit{S} \HOLSymConst{/} \HOLConst{At} \HOLStringLit{np})))
              (\HOLConst{Comma} (\HOLConst{OneForm} (\HOLConst{At} \HOLStringLit{S} \HOLSymConst{/} \HOLConst{At} \HOLStringLit{inf}))
                 (\HOLConst{OneForm} (\HOLConst{At} \HOLStringLit{inf} \HOLSymConst{/} \HOLConst{At} \HOLStringLit{np})))) (\HOLConst{At} \HOLStringLit{S}))
        \HOLConst{LeftSlash}
        [\HOLConst{Unf} (\HOLConst{Sequent} \HOLConst{L_Sequent} (\HOLConst{OneForm} (\HOLConst{At} \HOLStringLit{S})) (\HOLConst{At} \HOLStringLit{S}));
         \HOLConst{Unf}
           (\HOLConst{Sequent} \HOLConst{L_Sequent}
              (\HOLConst{Comma} (\HOLConst{OneForm} (\HOLConst{At} \HOLStringLit{S} \HOLSymConst{/} \HOLConst{At} \HOLStringLit{inf}))
                 (\HOLConst{OneForm} (\HOLConst{At} \HOLStringLit{inf} \HOLSymConst{/} \HOLConst{At} \HOLStringLit{np})))
              (\HOLConst{At} \HOLStringLit{S} \HOLSymConst{/} \HOLConst{At} \HOLStringLit{np}))])
\end{alltt}

If we repeat this process and manually expand the Dertree while prove
the new Dertree is derived from the last Dertree, finally the
transitivity of \HOLinline{\HOLConst{Deriv}} relation will let us prove that, the
final finished Dertree is indeed a valid proof for the original
unfinished Dertree, and thus it's indeed a valid proof for the
original Sequent theorem. We omited the intermediate steps and show
only the proof script of the final step:
\begin{lstlisting}
val r0_to_final = store_thm (
   "r0_to_final", ``Deriv ^r0 ^r_final``,
    ASSUME_TAC r0_to_r1''
 >> ASSUME_TAC (derivToDeriv r1_to_r2)
 >> ASSUME_TAC (derivToDeriv r2_to_r3)
 >> ASSUME_TAC (derivToDeriv r3_to_r4)
 >> ASSUME_TAC (derivToDeriv r4_to_r5)
 >> ASSUME_TAC (derivToDeriv r5_to_r6)
 >> ASSUME_TAC (derivToDeriv r6_to_final)
 >> REPEAT (IMP_RES_TAC Deriv_trans));
\end{lstlisting}
And the actual proved theorem:
\begin{alltt}
\HOLTokenTurnstile{} \HOLConst{Deriv}
     (\HOLConst{Unf}
        (\HOLConst{Sequent} \HOLConst{L_Sequent}
           (\HOLConst{Comma} (\HOLConst{OneForm} (\HOLConst{At} \HOLStringLit{S} \HOLSymConst{/} (\HOLConst{At} \HOLStringLit{S} \HOLSymConst{/} \HOLConst{At} \HOLStringLit{np})))
              (\HOLConst{Comma} (\HOLConst{OneForm} (\HOLConst{At} \HOLStringLit{S} \HOLSymConst{/} \HOLConst{At} \HOLStringLit{inf}))
                 (\HOLConst{OneForm} (\HOLConst{At} \HOLStringLit{inf} \HOLSymConst{/} \HOLConst{At} \HOLStringLit{np})))) (\HOLConst{At} \HOLStringLit{S})))
     (\HOLConst{Der}
        (\HOLConst{Sequent} \HOLConst{L_Sequent}
           (\HOLConst{Comma} (\HOLConst{OneForm} (\HOLConst{At} \HOLStringLit{S} \HOLSymConst{/} (\HOLConst{At} \HOLStringLit{S} \HOLSymConst{/} \HOLConst{At} \HOLStringLit{np})))
              (\HOLConst{Comma} (\HOLConst{OneForm} (\HOLConst{At} \HOLStringLit{S} \HOLSymConst{/} \HOLConst{At} \HOLStringLit{inf}))
                 (\HOLConst{OneForm} (\HOLConst{At} \HOLStringLit{inf} \HOLSymConst{/} \HOLConst{At} \HOLStringLit{np})))) (\HOLConst{At} \HOLStringLit{S}))
        \HOLConst{LeftSlash}
        [\HOLConst{Der} (\HOLConst{Sequent} \HOLConst{L_Sequent} (\HOLConst{OneForm} (\HOLConst{At} \HOLStringLit{S})) (\HOLConst{At} \HOLStringLit{S}))
           \HOLConst{SeqAxiom} [];
         \HOLConst{Der}
           (\HOLConst{Sequent} \HOLConst{L_Sequent}
              (\HOLConst{Comma} (\HOLConst{OneForm} (\HOLConst{At} \HOLStringLit{S} \HOLSymConst{/} \HOLConst{At} \HOLStringLit{inf}))
                 (\HOLConst{OneForm} (\HOLConst{At} \HOLStringLit{inf} \HOLSymConst{/} \HOLConst{At} \HOLStringLit{np})))
              (\HOLConst{At} \HOLStringLit{S} \HOLSymConst{/} \HOLConst{At} \HOLStringLit{np})) \HOLConst{RightSlash}
           [\HOLConst{Der}
              (\HOLConst{Sequent} \HOLConst{L_Sequent}
                 (\HOLConst{Comma}
                    (\HOLConst{Comma} (\HOLConst{OneForm} (\HOLConst{At} \HOLStringLit{S} \HOLSymConst{/} \HOLConst{At} \HOLStringLit{inf}))
                       (\HOLConst{OneForm} (\HOLConst{At} \HOLStringLit{inf} \HOLSymConst{/} \HOLConst{At} \HOLStringLit{np})))
                    (\HOLConst{OneForm} (\HOLConst{At} \HOLStringLit{np}))) (\HOLConst{At} \HOLStringLit{S})) \HOLConst{SeqExt}
              [\HOLConst{Der}
                 (\HOLConst{Sequent} \HOLConst{L_Sequent}
                    (\HOLConst{Comma} (\HOLConst{OneForm} (\HOLConst{At} \HOLStringLit{S} \HOLSymConst{/} \HOLConst{At} \HOLStringLit{inf}))
                       (\HOLConst{Comma} (\HOLConst{OneForm} (\HOLConst{At} \HOLStringLit{inf} \HOLSymConst{/} \HOLConst{At} \HOLStringLit{np}))
                          (\HOLConst{OneForm} (\HOLConst{At} \HOLStringLit{np})))) (\HOLConst{At} \HOLStringLit{S}))
                 \HOLConst{LeftSlash}
                 [\HOLConst{Der}
                    (\HOLConst{Sequent} \HOLConst{L_Sequent}
                       (\HOLConst{Comma} (\HOLConst{OneForm} (\HOLConst{At} \HOLStringLit{S} \HOLSymConst{/} \HOLConst{At} \HOLStringLit{inf}))
                          (\HOLConst{OneForm} (\HOLConst{At} \HOLStringLit{inf}))) (\HOLConst{At} \HOLStringLit{S}))
                    \HOLConst{LeftSlash}
                    [\HOLConst{Der}
                       (\HOLConst{Sequent} \HOLConst{L_Sequent} (\HOLConst{OneForm} (\HOLConst{At} \HOLStringLit{S}))
                          (\HOLConst{At} \HOLStringLit{S})) \HOLConst{SeqAxiom} [];
                     \HOLConst{Der}
                       (\HOLConst{Sequent} \HOLConst{L_Sequent} (\HOLConst{OneForm} (\HOLConst{At} \HOLStringLit{inf}))
                          (\HOLConst{At} \HOLStringLit{inf})) \HOLConst{SeqAxiom} []];
                  \HOLConst{Der}
                    (\HOLConst{Sequent} \HOLConst{L_Sequent} (\HOLConst{OneForm} (\HOLConst{At} \HOLStringLit{np}))
                       (\HOLConst{At} \HOLStringLit{np})) \HOLConst{SeqAxiom} []]]]])
\end{alltt}

The final Dertree is quite long and hard to read, but it does contain
all necessary information about the details of the proof.  If there's
an automatic proof searching algorithm, in theory we can implement it
either as a special tactical for proving theorems about relation
\HOLinline{\HOLConst{gentzenSequent}}, or as a program taking an initial Dertree and
produce a finished Dertree with a related theorem as above one.  If we
need a language parser instead, then the useful output will be the
bracketed binary tree, together with categories at each node of the tree.

\section{Differences between HOL and Coq}
\label{sec:differences-hol-coq}

There're essential differences between HOL and Coq for many
logical definitions that we have ported from Coq, although they looks
similiar.

\subsection{Inductive datatypes and relations}

In Coq, both inductive data types and relations are defined as
Inductive sets. For instance, the definition of type \texttt{Form} and
the \texttt{arrow} relation for Syntactic Calculus:
\begin{lstlisting}
 Inductive Form (Atoms : Set) : Set :=
   | At : Atoms -> Form Atoms
   | Slash : Form Atoms -> Form Atoms -> Form Atoms
   | Dot : Form Atoms -> Form Atoms -> Form Atoms
   | Backslash : Form Atoms -> Form Atoms -> Form Atoms.

 Inductive arrow (Atoms : Set) : Form Atoms -> Form Atoms -> Set :=
   | one : forall A : Form Atoms, arrow A A
   | comp : forall A B C : Form Atoms, arrow A B -> arrow B C -> arrow A C
   | beta :
       forall A B C : Form Atoms, arrow (Dot A B) C -> arrow A (Slash C B)
   | beta' :
       forall A B C : Form Atoms, arrow A (Slash C B) -> arrow (Dot A B) C
   | gamma :
       forall A B C : Form Atoms,
       arrow (Dot A B) C -> arrow B (Backslash A C)
   | gamma' :
       forall A B C : Form Atoms,
       arrow B (Backslash A C) -> arrow (Dot A B) C
   | arrow_plus : forall A B : Form Atoms, X A B -> arrow A B.
\end{lstlisting}
while in HOL, although they're still inductive defintions, but they're
handled differently: the former is defined by \texttt{Define}, and latter is
defined by \texttt{Hol_reln}:
\begin{lstlisting}
val _ = Datatype `Form = At 'a | Slash Form Form | Backslash Form Form | Dot Form Form`;

val (arrow_rules, arrow_ind, arrow_cases) = Hol_reln `
    (!X A. arrow X A A) /\						(* one *)
    (!X A B C. arrow X (Dot A B) C ==> arrow X A (Slash C B)) /\	(* beta *)
    (!X A B C. arrow X A (Slash C B) ==> arrow X (Dot A B) C) /\	(* beta' *)
    (!X A B C. arrow X (Dot A B) C ==> arrow X B (Backslash A C)) /\	(* gamma *)
    (!X A B C. arrow X B (Backslash A C) ==> arrow X (Dot A B) C) /\	(* gamma' *)
    (!X A B C. arrow X A B /\ arrow X B C ==> arrow X A C) /\		(* comp *)
    (!(X :'a arrow_extension) A B. X A B ==> arrow X A B) `;		(* arrow_plus *)
\end{lstlisting}
There's no magic behind HOL's datatype definition, because what the
\texttt{Datatype} does is to prove a series of theorems which
completely characteries the type \texttt{Form}:
\begin{alltt}
Form_TY_DEF:
\HOLTokenTurnstile{} \HOLSymConst{\HOLTokenExists{}}\HOLBoundVar{rep}.
     \HOLConst{TYPE_DEFINITION}
       (\HOLTokenLambda{}\HOLBoundVar{a\sb{\mathrm{0}}\sp{\prime}}.
          \HOLSymConst{\HOLTokenForall{}}\HOLBoundVar{'Form\sp{\prime}} .
            (\HOLSymConst{\HOLTokenForall{}}\HOLBoundVar{a\sb{\mathrm{0}}\sp{\prime}}.
               (\HOLSymConst{\HOLTokenExists{}}\HOLBoundVar{a}.
                  \HOLBoundVar{a\sb{\mathrm{0}}\sp{\prime}} \HOLSymConst{=}
                  (\HOLTokenLambda{}\HOLBoundVar{a}.
                     \HOLSymConst{ind_type\$CONSTR} \HOLNumLit{0} \HOLBoundVar{a} (\HOLTokenLambda{}\HOLBoundVar{n}. \HOLSymConst{ind_type\$BOTTOM}))
                    \HOLBoundVar{a}) \HOLSymConst{\HOLTokenDisj{}}
               (\HOLSymConst{\HOLTokenExists{}}\HOLBoundVar{a\sb{\mathrm{0}}} \HOLBoundVar{a\sb{\mathrm{1}}}.
                  (\HOLBoundVar{a\sb{\mathrm{0}}\sp{\prime}} \HOLSymConst{=}
                   (\HOLTokenLambda{}\HOLBoundVar{a\sb{\mathrm{0}}} \HOLBoundVar{a\sb{\mathrm{1}}}.
                      \HOLSymConst{ind_type\$CONSTR} (\HOLConst{SUC} \HOLNumLit{0}) \HOLConst{ARB}
                        (\HOLSymConst{ind_type\$FCONS} \HOLBoundVar{a\sb{\mathrm{0}}}
                           (\HOLSymConst{ind_type\$FCONS} \HOLBoundVar{a\sb{\mathrm{1}}}
                              (\HOLTokenLambda{}\HOLBoundVar{n}. \HOLSymConst{ind_type\$BOTTOM})))) \HOLBoundVar{a\sb{\mathrm{0}}} \HOLBoundVar{a\sb{\mathrm{1}}}) \HOLSymConst{\HOLTokenConj{}}
                  \HOLBoundVar{'Form\sp{\prime}} \HOLBoundVar{a\sb{\mathrm{0}}} \HOLSymConst{\HOLTokenConj{}} \HOLBoundVar{'Form\sp{\prime}} \HOLBoundVar{a\sb{\mathrm{1}}}) \HOLSymConst{\HOLTokenDisj{}}
               (\HOLSymConst{\HOLTokenExists{}}\HOLBoundVar{a\sb{\mathrm{0}}} \HOLBoundVar{a\sb{\mathrm{1}}}.
                  (\HOLBoundVar{a\sb{\mathrm{0}}\sp{\prime}} \HOLSymConst{=}
                   (\HOLTokenLambda{}\HOLBoundVar{a\sb{\mathrm{0}}} \HOLBoundVar{a\sb{\mathrm{1}}}.
                      \HOLSymConst{ind_type\$CONSTR} (\HOLConst{SUC} (\HOLConst{SUC} \HOLNumLit{0})) \HOLConst{ARB}
                        (\HOLSymConst{ind_type\$FCONS} \HOLBoundVar{a\sb{\mathrm{0}}}
                           (\HOLSymConst{ind_type\$FCONS} \HOLBoundVar{a\sb{\mathrm{1}}}
                              (\HOLTokenLambda{}\HOLBoundVar{n}. \HOLSymConst{ind_type\$BOTTOM})))) \HOLBoundVar{a\sb{\mathrm{0}}} \HOLBoundVar{a\sb{\mathrm{1}}}) \HOLSymConst{\HOLTokenConj{}}
                  \HOLBoundVar{'Form\sp{\prime}} \HOLBoundVar{a\sb{\mathrm{0}}} \HOLSymConst{\HOLTokenConj{}} \HOLBoundVar{'Form\sp{\prime}} \HOLBoundVar{a\sb{\mathrm{1}}}) \HOLSymConst{\HOLTokenDisj{}}
               (\HOLSymConst{\HOLTokenExists{}}\HOLBoundVar{a\sb{\mathrm{0}}} \HOLBoundVar{a\sb{\mathrm{1}}}.
                  (\HOLBoundVar{a\sb{\mathrm{0}}\sp{\prime}} \HOLSymConst{=}
                   (\HOLTokenLambda{}\HOLBoundVar{a\sb{\mathrm{0}}} \HOLBoundVar{a\sb{\mathrm{1}}}.
                      \HOLSymConst{ind_type\$CONSTR} (\HOLConst{SUC} (\HOLConst{SUC} (\HOLConst{SUC} \HOLNumLit{0}))) \HOLConst{ARB}
                        (\HOLSymConst{ind_type\$FCONS} \HOLBoundVar{a\sb{\mathrm{0}}}
                           (\HOLSymConst{ind_type\$FCONS} \HOLBoundVar{a\sb{\mathrm{1}}}
                              (\HOLTokenLambda{}\HOLBoundVar{n}. \HOLSymConst{ind_type\$BOTTOM})))) \HOLBoundVar{a\sb{\mathrm{0}}} \HOLBoundVar{a\sb{\mathrm{1}}}) \HOLSymConst{\HOLTokenConj{}}
                  \HOLBoundVar{'Form\sp{\prime}} \HOLBoundVar{a\sb{\mathrm{0}}} \HOLSymConst{\HOLTokenConj{}} \HOLBoundVar{'Form\sp{\prime}} \HOLBoundVar{a\sb{\mathrm{1}}}) \HOLSymConst{\HOLTokenImp{}}
               \HOLBoundVar{'Form\sp{\prime}} \HOLBoundVar{a\sb{\mathrm{0}}\sp{\prime}}) \HOLSymConst{\HOLTokenImp{}}
            \HOLBoundVar{'Form\sp{\prime}} \HOLBoundVar{a\sb{\mathrm{0}}\sp{\prime}}) \HOLBoundVar{rep}
Form_case_def:
\HOLTokenTurnstile{} (\HOLSymConst{\HOLTokenForall{}}\HOLBoundVar{a} \HOLBoundVar{f} \HOLBoundVar{f\sb{\mathrm{1}}} \HOLBoundVar{f\sb{\mathrm{2}}} \HOLBoundVar{f\sb{\mathrm{3}}}. \HOLConst{Form_CASE} (\HOLConst{At} \HOLBoundVar{a}) \HOLBoundVar{f} \HOLBoundVar{f\sb{\mathrm{1}}} \HOLBoundVar{f\sb{\mathrm{2}}} \HOLBoundVar{f\sb{\mathrm{3}}} \HOLSymConst{=} \HOLBoundVar{f} \HOLBoundVar{a}) \HOLSymConst{\HOLTokenConj{}}
   (\HOLSymConst{\HOLTokenForall{}}\HOLBoundVar{a\sb{\mathrm{0}}} \HOLBoundVar{a\sb{\mathrm{1}}} \HOLBoundVar{f} \HOLBoundVar{f\sb{\mathrm{1}}} \HOLBoundVar{f\sb{\mathrm{2}}} \HOLBoundVar{f\sb{\mathrm{3}}}.
      \HOLConst{Form_CASE} (\HOLBoundVar{a\sb{\mathrm{0}}} \HOLSymConst{/} \HOLBoundVar{a\sb{\mathrm{1}}}) \HOLBoundVar{f} \HOLBoundVar{f\sb{\mathrm{1}}} \HOLBoundVar{f\sb{\mathrm{2}}} \HOLBoundVar{f\sb{\mathrm{3}}} \HOLSymConst{=} \HOLBoundVar{f\sb{\mathrm{1}}} \HOLBoundVar{a\sb{\mathrm{0}}} \HOLBoundVar{a\sb{\mathrm{1}}}) \HOLSymConst{\HOLTokenConj{}}
   (\HOLSymConst{\HOLTokenForall{}}\HOLBoundVar{a\sb{\mathrm{0}}} \HOLBoundVar{a\sb{\mathrm{1}}} \HOLBoundVar{f} \HOLBoundVar{f\sb{\mathrm{1}}} \HOLBoundVar{f\sb{\mathrm{2}}} \HOLBoundVar{f\sb{\mathrm{3}}}.
      \HOLConst{Form_CASE} (\HOLBoundVar{a\sb{\mathrm{0}}} \HOLSymConst{\HOLTokenBackslash} \HOLBoundVar{a\sb{\mathrm{1}}}) \HOLBoundVar{f} \HOLBoundVar{f\sb{\mathrm{1}}} \HOLBoundVar{f\sb{\mathrm{2}}} \HOLBoundVar{f\sb{\mathrm{3}}} \HOLSymConst{=} \HOLBoundVar{f\sb{\mathrm{2}}} \HOLBoundVar{a\sb{\mathrm{0}}} \HOLBoundVar{a\sb{\mathrm{1}}}) \HOLSymConst{\HOLTokenConj{}}
   \HOLSymConst{\HOLTokenForall{}}\HOLBoundVar{a\sb{\mathrm{0}}} \HOLBoundVar{a\sb{\mathrm{1}}} \HOLBoundVar{f} \HOLBoundVar{f\sb{\mathrm{1}}} \HOLBoundVar{f\sb{\mathrm{2}}} \HOLBoundVar{f\sb{\mathrm{3}}}. \HOLConst{Form_CASE} (\HOLBoundVar{a\sb{\mathrm{0}}} \HOLSymConst{\HOLTokenProd{}} \HOLBoundVar{a\sb{\mathrm{1}}}) \HOLBoundVar{f} \HOLBoundVar{f\sb{\mathrm{1}}} \HOLBoundVar{f\sb{\mathrm{2}}} \HOLBoundVar{f\sb{\mathrm{3}}} \HOLSymConst{=} \HOLBoundVar{f\sb{\mathrm{3}}} \HOLBoundVar{a\sb{\mathrm{0}}} \HOLBoundVar{a\sb{\mathrm{1}}}
Form_size_def:
\HOLTokenTurnstile{} (\HOLSymConst{\HOLTokenForall{}}\HOLBoundVar{f} \HOLBoundVar{a}. \HOLConst{Form_size} \HOLBoundVar{f} (\HOLConst{At} \HOLBoundVar{a}) \HOLSymConst{=} \HOLNumLit{1} \HOLSymConst{+} \HOLBoundVar{f} \HOLBoundVar{a}) \HOLSymConst{\HOLTokenConj{}}
   (\HOLSymConst{\HOLTokenForall{}}\HOLBoundVar{f} \HOLBoundVar{a\sb{\mathrm{0}}} \HOLBoundVar{a\sb{\mathrm{1}}}.
      \HOLConst{Form_size} \HOLBoundVar{f} (\HOLBoundVar{a\sb{\mathrm{0}}} \HOLSymConst{/} \HOLBoundVar{a\sb{\mathrm{1}}}) \HOLSymConst{=}
      \HOLNumLit{1} \HOLSymConst{+} (\HOLConst{Form_size} \HOLBoundVar{f} \HOLBoundVar{a\sb{\mathrm{0}}} \HOLSymConst{+} \HOLConst{Form_size} \HOLBoundVar{f} \HOLBoundVar{a\sb{\mathrm{1}}})) \HOLSymConst{\HOLTokenConj{}}
   (\HOLSymConst{\HOLTokenForall{}}\HOLBoundVar{f} \HOLBoundVar{a\sb{\mathrm{0}}} \HOLBoundVar{a\sb{\mathrm{1}}}.
      \HOLConst{Form_size} \HOLBoundVar{f} (\HOLBoundVar{a\sb{\mathrm{0}}} \HOLSymConst{\HOLTokenBackslash} \HOLBoundVar{a\sb{\mathrm{1}}}) \HOLSymConst{=}
      \HOLNumLit{1} \HOLSymConst{+} (\HOLConst{Form_size} \HOLBoundVar{f} \HOLBoundVar{a\sb{\mathrm{0}}} \HOLSymConst{+} \HOLConst{Form_size} \HOLBoundVar{f} \HOLBoundVar{a\sb{\mathrm{1}}})) \HOLSymConst{\HOLTokenConj{}}
   \HOLSymConst{\HOLTokenForall{}}\HOLBoundVar{f} \HOLBoundVar{a\sb{\mathrm{0}}} \HOLBoundVar{a\sb{\mathrm{1}}}.
     \HOLConst{Form_size} \HOLBoundVar{f} (\HOLBoundVar{a\sb{\mathrm{0}}} \HOLSymConst{\HOLTokenProd{}} \HOLBoundVar{a\sb{\mathrm{1}}}) \HOLSymConst{=}
     \HOLNumLit{1} \HOLSymConst{+} (\HOLConst{Form_size} \HOLBoundVar{f} \HOLBoundVar{a\sb{\mathrm{0}}} \HOLSymConst{+} \HOLConst{Form_size} \HOLBoundVar{f} \HOLBoundVar{a\sb{\mathrm{1}}})
Form_11:
\HOLTokenTurnstile{} (\HOLSymConst{\HOLTokenForall{}}\HOLBoundVar{a} \HOLBoundVar{a\sp{\prime}}. (\HOLConst{At} \HOLBoundVar{a} \HOLSymConst{=} \HOLConst{At} \HOLBoundVar{a\sp{\prime}}) \HOLSymConst{\HOLTokenEquiv{}} (\HOLBoundVar{a} \HOLSymConst{=} \HOLBoundVar{a\sp{\prime}})) \HOLSymConst{\HOLTokenConj{}}
   (\HOLSymConst{\HOLTokenForall{}}\HOLBoundVar{a\sb{\mathrm{0}}} \HOLBoundVar{a\sb{\mathrm{1}}} \HOLBoundVar{a\sb{\mathrm{0}}\sp{\prime}} \HOLBoundVar{a\sb{\mathrm{1}}\sp{\prime}}.
      (\HOLBoundVar{a\sb{\mathrm{0}}} \HOLSymConst{/} \HOLBoundVar{a\sb{\mathrm{1}}} \HOLSymConst{=} \HOLBoundVar{a\sb{\mathrm{0}}\sp{\prime}} \HOLSymConst{/} \HOLBoundVar{a\sb{\mathrm{1}}\sp{\prime}}) \HOLSymConst{\HOLTokenEquiv{}} (\HOLBoundVar{a\sb{\mathrm{0}}} \HOLSymConst{=} \HOLBoundVar{a\sb{\mathrm{0}}\sp{\prime}}) \HOLSymConst{\HOLTokenConj{}} (\HOLBoundVar{a\sb{\mathrm{1}}} \HOLSymConst{=} \HOLBoundVar{a\sb{\mathrm{1}}\sp{\prime}})) \HOLSymConst{\HOLTokenConj{}}
   (\HOLSymConst{\HOLTokenForall{}}\HOLBoundVar{a\sb{\mathrm{0}}} \HOLBoundVar{a\sb{\mathrm{1}}} \HOLBoundVar{a\sb{\mathrm{0}}\sp{\prime}} \HOLBoundVar{a\sb{\mathrm{1}}\sp{\prime}}.
      (\HOLBoundVar{a\sb{\mathrm{0}}} \HOLSymConst{\HOLTokenBackslash} \HOLBoundVar{a\sb{\mathrm{1}}} \HOLSymConst{=} \HOLBoundVar{a\sb{\mathrm{0}}\sp{\prime}} \HOLSymConst{\HOLTokenBackslash} \HOLBoundVar{a\sb{\mathrm{1}}\sp{\prime}}) \HOLSymConst{\HOLTokenEquiv{}} (\HOLBoundVar{a\sb{\mathrm{0}}} \HOLSymConst{=} \HOLBoundVar{a\sb{\mathrm{0}}\sp{\prime}}) \HOLSymConst{\HOLTokenConj{}} (\HOLBoundVar{a\sb{\mathrm{1}}} \HOLSymConst{=} \HOLBoundVar{a\sb{\mathrm{1}}\sp{\prime}})) \HOLSymConst{\HOLTokenConj{}}
   \HOLSymConst{\HOLTokenForall{}}\HOLBoundVar{a\sb{\mathrm{0}}} \HOLBoundVar{a\sb{\mathrm{1}}} \HOLBoundVar{a\sb{\mathrm{0}}\sp{\prime}} \HOLBoundVar{a\sb{\mathrm{1}}\sp{\prime}}.
     (\HOLBoundVar{a\sb{\mathrm{0}}} \HOLSymConst{\HOLTokenProd{}} \HOLBoundVar{a\sb{\mathrm{1}}} \HOLSymConst{=} \HOLBoundVar{a\sb{\mathrm{0}}\sp{\prime}} \HOLSymConst{\HOLTokenProd{}} \HOLBoundVar{a\sb{\mathrm{1}}\sp{\prime}}) \HOLSymConst{\HOLTokenEquiv{}} (\HOLBoundVar{a\sb{\mathrm{0}}} \HOLSymConst{=} \HOLBoundVar{a\sb{\mathrm{0}}\sp{\prime}}) \HOLSymConst{\HOLTokenConj{}} (\HOLBoundVar{a\sb{\mathrm{1}}} \HOLSymConst{=} \HOLBoundVar{a\sb{\mathrm{1}}\sp{\prime}})
Form_Axiom:
\HOLTokenTurnstile{} \HOLSymConst{\HOLTokenExists{}}\HOLBoundVar{fn}.
     (\HOLSymConst{\HOLTokenForall{}}\HOLBoundVar{a}. \HOLBoundVar{fn} (\HOLConst{At} \HOLBoundVar{a}) \HOLSymConst{=} \HOLFreeVar{f\sb{\mathrm{0}}} \HOLBoundVar{a}) \HOLSymConst{\HOLTokenConj{}}
     (\HOLSymConst{\HOLTokenForall{}}\HOLBoundVar{a\sb{\mathrm{0}}} \HOLBoundVar{a\sb{\mathrm{1}}}. \HOLBoundVar{fn} (\HOLBoundVar{a\sb{\mathrm{0}}} \HOLSymConst{/} \HOLBoundVar{a\sb{\mathrm{1}}}) \HOLSymConst{=} \HOLFreeVar{f\sb{\mathrm{1}}} \HOLBoundVar{a\sb{\mathrm{0}}} \HOLBoundVar{a\sb{\mathrm{1}}} (\HOLBoundVar{fn} \HOLBoundVar{a\sb{\mathrm{0}}}) (\HOLBoundVar{fn} \HOLBoundVar{a\sb{\mathrm{1}}})) \HOLSymConst{\HOLTokenConj{}}
     (\HOLSymConst{\HOLTokenForall{}}\HOLBoundVar{a\sb{\mathrm{0}}} \HOLBoundVar{a\sb{\mathrm{1}}}. \HOLBoundVar{fn} (\HOLBoundVar{a\sb{\mathrm{0}}} \HOLSymConst{\HOLTokenBackslash} \HOLBoundVar{a\sb{\mathrm{1}}}) \HOLSymConst{=} \HOLFreeVar{f\sb{\mathrm{2}}} \HOLBoundVar{a\sb{\mathrm{0}}} \HOLBoundVar{a\sb{\mathrm{1}}} (\HOLBoundVar{fn} \HOLBoundVar{a\sb{\mathrm{0}}}) (\HOLBoundVar{fn} \HOLBoundVar{a\sb{\mathrm{1}}})) \HOLSymConst{\HOLTokenConj{}}
     \HOLSymConst{\HOLTokenForall{}}\HOLBoundVar{a\sb{\mathrm{0}}} \HOLBoundVar{a\sb{\mathrm{1}}}. \HOLBoundVar{fn} (\HOLBoundVar{a\sb{\mathrm{0}}} \HOLSymConst{\HOLTokenProd{}} \HOLBoundVar{a\sb{\mathrm{1}}}) \HOLSymConst{=} \HOLFreeVar{f\sb{\mathrm{3}}} \HOLBoundVar{a\sb{\mathrm{0}}} \HOLBoundVar{a\sb{\mathrm{1}}} (\HOLBoundVar{fn} \HOLBoundVar{a\sb{\mathrm{0}}}) (\HOLBoundVar{fn} \HOLBoundVar{a\sb{\mathrm{1}}})
Form_case_cong:
\HOLTokenTurnstile{} (\HOLFreeVar{M} \HOLSymConst{=} \HOLFreeVar{M\sp{\prime}}) \HOLSymConst{\HOLTokenConj{}} (\HOLSymConst{\HOLTokenForall{}}\HOLBoundVar{a}. (\HOLFreeVar{M\sp{\prime}} \HOLSymConst{=} \HOLConst{At} \HOLBoundVar{a}) \HOLSymConst{\HOLTokenImp{}} (\HOLFreeVar{f} \HOLBoundVar{a} \HOLSymConst{=} \HOLFreeVar{f\sp{\prime}} \HOLBoundVar{a})) \HOLSymConst{\HOLTokenConj{}}
   (\HOLSymConst{\HOLTokenForall{}}\HOLBoundVar{a\sb{\mathrm{0}}} \HOLBoundVar{a\sb{\mathrm{1}}}. (\HOLFreeVar{M\sp{\prime}} \HOLSymConst{=} \HOLBoundVar{a\sb{\mathrm{0}}} \HOLSymConst{/} \HOLBoundVar{a\sb{\mathrm{1}}}) \HOLSymConst{\HOLTokenImp{}} (\HOLFreeVar{f\sb{\mathrm{1}}} \HOLBoundVar{a\sb{\mathrm{0}}} \HOLBoundVar{a\sb{\mathrm{1}}} \HOLSymConst{=} \HOLFreeVar{f\sb{\mathrm{1}}\sp{\prime}} \HOLBoundVar{a\sb{\mathrm{0}}} \HOLBoundVar{a\sb{\mathrm{1}}})) \HOLSymConst{\HOLTokenConj{}}
   (\HOLSymConst{\HOLTokenForall{}}\HOLBoundVar{a\sb{\mathrm{0}}} \HOLBoundVar{a\sb{\mathrm{1}}}. (\HOLFreeVar{M\sp{\prime}} \HOLSymConst{=} \HOLBoundVar{a\sb{\mathrm{0}}} \HOLSymConst{\HOLTokenBackslash} \HOLBoundVar{a\sb{\mathrm{1}}}) \HOLSymConst{\HOLTokenImp{}} (\HOLFreeVar{f\sb{\mathrm{2}}} \HOLBoundVar{a\sb{\mathrm{0}}} \HOLBoundVar{a\sb{\mathrm{1}}} \HOLSymConst{=} \HOLFreeVar{f\sb{\mathrm{2}}\sp{\prime}} \HOLBoundVar{a\sb{\mathrm{0}}} \HOLBoundVar{a\sb{\mathrm{1}}})) \HOLSymConst{\HOLTokenConj{}}
   (\HOLSymConst{\HOLTokenForall{}}\HOLBoundVar{a\sb{\mathrm{0}}} \HOLBoundVar{a\sb{\mathrm{1}}}. (\HOLFreeVar{M\sp{\prime}} \HOLSymConst{=} \HOLBoundVar{a\sb{\mathrm{0}}} \HOLSymConst{\HOLTokenProd{}} \HOLBoundVar{a\sb{\mathrm{1}}}) \HOLSymConst{\HOLTokenImp{}} (\HOLFreeVar{f\sb{\mathrm{3}}} \HOLBoundVar{a\sb{\mathrm{0}}} \HOLBoundVar{a\sb{\mathrm{1}}} \HOLSymConst{=} \HOLFreeVar{f\sb{\mathrm{3}}\sp{\prime}} \HOLBoundVar{a\sb{\mathrm{0}}} \HOLBoundVar{a\sb{\mathrm{1}}})) \HOLSymConst{\HOLTokenImp{}}
   (\HOLConst{Form_CASE} \HOLFreeVar{M} \HOLFreeVar{f} \HOLFreeVar{f\sb{\mathrm{1}}} \HOLFreeVar{f\sb{\mathrm{2}}} \HOLFreeVar{f\sb{\mathrm{3}}} \HOLSymConst{=} \HOLConst{Form_CASE} \HOLFreeVar{M\sp{\prime}} \HOLFreeVar{f\sp{\prime}} \HOLFreeVar{f\sb{\mathrm{1}}\sp{\prime}} \HOLFreeVar{f\sb{\mathrm{2}}\sp{\prime}} \HOLFreeVar{f\sb{\mathrm{3}}\sp{\prime}})
Form_distinct:
\HOLTokenTurnstile{} (\HOLSymConst{\HOLTokenForall{}}\HOLBoundVar{a\sb{\mathrm{1}}} \HOLBoundVar{a\sb{\mathrm{0}}} \HOLBoundVar{a}. \HOLConst{At} \HOLBoundVar{a} \HOLSymConst{\HOLTokenNotEqual{}} \HOLBoundVar{a\sb{\mathrm{0}}} \HOLSymConst{/} \HOLBoundVar{a\sb{\mathrm{1}}}) \HOLSymConst{\HOLTokenConj{}} (\HOLSymConst{\HOLTokenForall{}}\HOLBoundVar{a\sb{\mathrm{1}}} \HOLBoundVar{a\sb{\mathrm{0}}} \HOLBoundVar{a}. \HOLConst{At} \HOLBoundVar{a} \HOLSymConst{\HOLTokenNotEqual{}} \HOLBoundVar{a\sb{\mathrm{0}}} \HOLSymConst{\HOLTokenBackslash} \HOLBoundVar{a\sb{\mathrm{1}}}) \HOLSymConst{\HOLTokenConj{}}
   (\HOLSymConst{\HOLTokenForall{}}\HOLBoundVar{a\sb{\mathrm{1}}} \HOLBoundVar{a\sb{\mathrm{0}}} \HOLBoundVar{a}. \HOLConst{At} \HOLBoundVar{a} \HOLSymConst{\HOLTokenNotEqual{}} \HOLBoundVar{a\sb{\mathrm{0}}} \HOLSymConst{\HOLTokenProd{}} \HOLBoundVar{a\sb{\mathrm{1}}}) \HOLSymConst{\HOLTokenConj{}}
   (\HOLSymConst{\HOLTokenForall{}}\HOLBoundVar{a\sb{\mathrm{1}}\sp{\prime}} \HOLBoundVar{a\sb{\mathrm{1}}} \HOLBoundVar{a\sb{\mathrm{0}}\sp{\prime}} \HOLBoundVar{a\sb{\mathrm{0}}}. \HOLBoundVar{a\sb{\mathrm{0}}} \HOLSymConst{/} \HOLBoundVar{a\sb{\mathrm{1}}} \HOLSymConst{\HOLTokenNotEqual{}} \HOLBoundVar{a\sb{\mathrm{0}}\sp{\prime}} \HOLSymConst{\HOLTokenBackslash} \HOLBoundVar{a\sb{\mathrm{1}}\sp{\prime}}) \HOLSymConst{\HOLTokenConj{}}
   (\HOLSymConst{\HOLTokenForall{}}\HOLBoundVar{a\sb{\mathrm{1}}\sp{\prime}} \HOLBoundVar{a\sb{\mathrm{1}}} \HOLBoundVar{a\sb{\mathrm{0}}\sp{\prime}} \HOLBoundVar{a\sb{\mathrm{0}}}. \HOLBoundVar{a\sb{\mathrm{0}}} \HOLSymConst{/} \HOLBoundVar{a\sb{\mathrm{1}}} \HOLSymConst{\HOLTokenNotEqual{}} \HOLBoundVar{a\sb{\mathrm{0}}\sp{\prime}} \HOLSymConst{\HOLTokenProd{}} \HOLBoundVar{a\sb{\mathrm{1}}\sp{\prime}}) \HOLSymConst{\HOLTokenConj{}}
   \HOLSymConst{\HOLTokenForall{}}\HOLBoundVar{a\sb{\mathrm{1}}\sp{\prime}} \HOLBoundVar{a\sb{\mathrm{1}}} \HOLBoundVar{a\sb{\mathrm{0}}\sp{\prime}} \HOLBoundVar{a\sb{\mathrm{0}}}. \HOLBoundVar{a\sb{\mathrm{0}}} \HOLSymConst{\HOLTokenBackslash} \HOLBoundVar{a\sb{\mathrm{1}}} \HOLSymConst{\HOLTokenNotEqual{}} \HOLBoundVar{a\sb{\mathrm{0}}\sp{\prime}} \HOLSymConst{\HOLTokenProd{}} \HOLBoundVar{a\sb{\mathrm{1}}\sp{\prime}}
Form_induction:
\HOLTokenTurnstile{} (\HOLSymConst{\HOLTokenForall{}}\HOLBoundVar{a}. \HOLFreeVar{P} (\HOLConst{At} \HOLBoundVar{a})) \HOLSymConst{\HOLTokenConj{}} (\HOLSymConst{\HOLTokenForall{}}\HOLBoundVar{F\sp{\prime}} \HOLBoundVar{F\sb{\mathrm{0}}}. \HOLFreeVar{P} \HOLBoundVar{F\sp{\prime}} \HOLSymConst{\HOLTokenConj{}} \HOLFreeVar{P} \HOLBoundVar{F\sb{\mathrm{0}}} \HOLSymConst{\HOLTokenImp{}} \HOLFreeVar{P} (\HOLBoundVar{F\sp{\prime}} \HOLSymConst{/} \HOLBoundVar{F\sb{\mathrm{0}}})) \HOLSymConst{\HOLTokenConj{}}
   (\HOLSymConst{\HOLTokenForall{}}\HOLBoundVar{F\sp{\prime}} \HOLBoundVar{F\sb{\mathrm{0}}}. \HOLFreeVar{P} \HOLBoundVar{F\sp{\prime}} \HOLSymConst{\HOLTokenConj{}} \HOLFreeVar{P} \HOLBoundVar{F\sb{\mathrm{0}}} \HOLSymConst{\HOLTokenImp{}} \HOLFreeVar{P} (\HOLBoundVar{F\sp{\prime}} \HOLSymConst{\HOLTokenBackslash} \HOLBoundVar{F\sb{\mathrm{0}}})) \HOLSymConst{\HOLTokenConj{}}
   (\HOLSymConst{\HOLTokenForall{}}\HOLBoundVar{F\sp{\prime}} \HOLBoundVar{F\sb{\mathrm{0}}}. \HOLFreeVar{P} \HOLBoundVar{F\sp{\prime}} \HOLSymConst{\HOLTokenConj{}} \HOLFreeVar{P} \HOLBoundVar{F\sb{\mathrm{0}}} \HOLSymConst{\HOLTokenImp{}} \HOLFreeVar{P} (\HOLBoundVar{F\sp{\prime}} \HOLSymConst{\HOLTokenProd{}} \HOLBoundVar{F\sb{\mathrm{0}}})) \HOLSymConst{\HOLTokenImp{}}
   \HOLSymConst{\HOLTokenForall{}}\HOLBoundVar{F\sp{\prime}}. \HOLFreeVar{P} \HOLBoundVar{F\sp{\prime}}
Form_nchotomy:
\HOLTokenTurnstile{} (\HOLSymConst{\HOLTokenExists{}}\HOLBoundVar{a}. \HOLFreeVar{F'F\sp{\prime}} \HOLSymConst{=} \HOLConst{At} \HOLBoundVar{a}) \HOLSymConst{\HOLTokenDisj{}} (\HOLSymConst{\HOLTokenExists{}}\HOLBoundVar{F\sp{\prime}} \HOLBoundVar{F\sb{\mathrm{0}}}. \HOLFreeVar{F'F\sp{\prime}} \HOLSymConst{=} \HOLBoundVar{F\sp{\prime}} \HOLSymConst{/} \HOLBoundVar{F\sb{\mathrm{0}}}) \HOLSymConst{\HOLTokenDisj{}}
   (\HOLSymConst{\HOLTokenExists{}}\HOLBoundVar{F\sp{\prime}} \HOLBoundVar{F\sb{\mathrm{0}}}. \HOLFreeVar{F'F\sp{\prime}} \HOLSymConst{=} \HOLBoundVar{F\sp{\prime}} \HOLSymConst{\HOLTokenBackslash} \HOLBoundVar{F\sb{\mathrm{0}}}) \HOLSymConst{\HOLTokenDisj{}} \HOLSymConst{\HOLTokenExists{}}\HOLBoundVar{F\sp{\prime}} \HOLBoundVar{F\sb{\mathrm{0}}}. \HOLFreeVar{F'F\sp{\prime}} \HOLSymConst{=} \HOLBoundVar{F\sp{\prime}} \HOLSymConst{\HOLTokenProd{}} \HOLBoundVar{F\sb{\mathrm{0}}}
\end{alltt}
where the constant \texttt{TYPE\_DEFINITION} is defined in the theory
    \texttt{bool} by:
\begin{alltt}
\HOLTokenTurnstile{} \HOLConst{TYPE_DEFINITION} \HOLSymConst{=}
   (\HOLTokenLambda{}\HOLBoundVar{P} \HOLBoundVar{rep}.
      (\HOLSymConst{\HOLTokenForall{}}\HOLBoundVar{x\sp{\prime}} \HOLBoundVar{x\sp{\prime\prime}}. (\HOLBoundVar{rep} \HOLBoundVar{x\sp{\prime}} \HOLSymConst{=} \HOLBoundVar{rep} \HOLBoundVar{x\sp{\prime\prime}}) \HOLSymConst{\HOLTokenImp{}} (\HOLBoundVar{x\sp{\prime}} \HOLSymConst{=} \HOLBoundVar{x\sp{\prime\prime}})) \HOLSymConst{\HOLTokenConj{}}
      \HOLSymConst{\HOLTokenForall{}}\HOLBoundVar{x}. \HOLBoundVar{P} \HOLBoundVar{x} \HOLSymConst{\HOLTokenEquiv{}} \HOLSymConst{\HOLTokenExists{}}\HOLBoundVar{x\sp{\prime}}. \HOLBoundVar{x} \HOLSymConst{=} \HOLBoundVar{rep} \HOLBoundVar{x\sp{\prime}})
\end{alltt}
With all above theorems, any theorem in which the type \texttt{Form}
is used, can be handled by combining above theorems with other related
theorems, although most of time, some HOL's tacticals can benefit from
these generated theorems implicitly.  In Coq, datatypes are handled in
black-box like ways: user has no direct access to any theorem related
to datatype themselves.

Similarily, an induction relation \texttt{arrow} in HOL is just defined by some
generated theorems:
\begin{alltt}
arrow_def:
\HOLTokenTurnstile{} \HOLConst{arrow} \HOLSymConst{=}
   (\HOLTokenLambda{}\HOLBoundVar{a\sb{\mathrm{0}}} \HOLBoundVar{a\sb{\mathrm{1}}} \HOLBoundVar{a\sb{\mathrm{2}}}.
      \HOLSymConst{\HOLTokenForall{}}\HOLBoundVar{arrow\sp{\prime}}.
        (\HOLSymConst{\HOLTokenForall{}}\HOLBoundVar{a\sb{\mathrm{0}}} \HOLBoundVar{a\sb{\mathrm{1}}} \HOLBoundVar{a\sb{\mathrm{2}}}.
           (\HOLBoundVar{a\sb{\mathrm{2}}} \HOLSymConst{=} \HOLBoundVar{a\sb{\mathrm{1}}}) \HOLSymConst{\HOLTokenDisj{}}
           (\HOLSymConst{\HOLTokenExists{}}\HOLBoundVar{B} \HOLBoundVar{C}. (\HOLBoundVar{a\sb{\mathrm{2}}} \HOLSymConst{=} \HOLBoundVar{C} \HOLSymConst{/} \HOLBoundVar{B}) \HOLSymConst{\HOLTokenConj{}} \HOLBoundVar{arrow\sp{\prime}} \HOLBoundVar{a\sb{\mathrm{0}}} (\HOLBoundVar{a\sb{\mathrm{1}}} \HOLSymConst{\HOLTokenProd{}} \HOLBoundVar{B}) \HOLBoundVar{C}) \HOLSymConst{\HOLTokenDisj{}}
           (\HOLSymConst{\HOLTokenExists{}}\HOLBoundVar{A} \HOLBoundVar{B}. (\HOLBoundVar{a\sb{\mathrm{1}}} \HOLSymConst{=} \HOLBoundVar{A} \HOLSymConst{\HOLTokenProd{}} \HOLBoundVar{B}) \HOLSymConst{\HOLTokenConj{}} \HOLBoundVar{arrow\sp{\prime}} \HOLBoundVar{a\sb{\mathrm{0}}} \HOLBoundVar{A} (\HOLBoundVar{a\sb{\mathrm{2}}} \HOLSymConst{/} \HOLBoundVar{B})) \HOLSymConst{\HOLTokenDisj{}}
           (\HOLSymConst{\HOLTokenExists{}}\HOLBoundVar{A} \HOLBoundVar{C}. (\HOLBoundVar{a\sb{\mathrm{2}}} \HOLSymConst{=} \HOLBoundVar{A} \HOLSymConst{\HOLTokenBackslash} \HOLBoundVar{C}) \HOLSymConst{\HOLTokenConj{}} \HOLBoundVar{arrow\sp{\prime}} \HOLBoundVar{a\sb{\mathrm{0}}} (\HOLBoundVar{A} \HOLSymConst{\HOLTokenProd{}} \HOLBoundVar{a\sb{\mathrm{1}}}) \HOLBoundVar{C}) \HOLSymConst{\HOLTokenDisj{}}
           (\HOLSymConst{\HOLTokenExists{}}\HOLBoundVar{A} \HOLBoundVar{B}. (\HOLBoundVar{a\sb{\mathrm{1}}} \HOLSymConst{=} \HOLBoundVar{A} \HOLSymConst{\HOLTokenProd{}} \HOLBoundVar{B}) \HOLSymConst{\HOLTokenConj{}} \HOLBoundVar{arrow\sp{\prime}} \HOLBoundVar{a\sb{\mathrm{0}}} \HOLBoundVar{B} (\HOLBoundVar{A} \HOLSymConst{\HOLTokenBackslash} \HOLBoundVar{a\sb{\mathrm{2}}})) \HOLSymConst{\HOLTokenDisj{}}
           (\HOLSymConst{\HOLTokenExists{}}\HOLBoundVar{B}. \HOLBoundVar{arrow\sp{\prime}} \HOLBoundVar{a\sb{\mathrm{0}}} \HOLBoundVar{a\sb{\mathrm{1}}} \HOLBoundVar{B} \HOLSymConst{\HOLTokenConj{}} \HOLBoundVar{arrow\sp{\prime}} \HOLBoundVar{a\sb{\mathrm{0}}} \HOLBoundVar{B} \HOLBoundVar{a\sb{\mathrm{2}}}) \HOLSymConst{\HOLTokenDisj{}} \HOLBoundVar{a\sb{\mathrm{0}}} \HOLBoundVar{a\sb{\mathrm{1}}} \HOLBoundVar{a\sb{\mathrm{2}}} \HOLSymConst{\HOLTokenImp{}}
           \HOLBoundVar{arrow\sp{\prime}} \HOLBoundVar{a\sb{\mathrm{0}}} \HOLBoundVar{a\sb{\mathrm{1}}} \HOLBoundVar{a\sb{\mathrm{2}}}) \HOLSymConst{\HOLTokenImp{}}
        \HOLBoundVar{arrow\sp{\prime}} \HOLBoundVar{a\sb{\mathrm{0}}} \HOLBoundVar{a\sb{\mathrm{1}}} \HOLBoundVar{a\sb{\mathrm{2}}})
arrow_rules:
\HOLTokenTurnstile{} (\HOLSymConst{\HOLTokenForall{}}\HOLBoundVar{X} \HOLBoundVar{A}. \HOLConst{arrow} \HOLBoundVar{X} \HOLBoundVar{A} \HOLBoundVar{A}) \HOLSymConst{\HOLTokenConj{}}
   (\HOLSymConst{\HOLTokenForall{}}\HOLBoundVar{X} \HOLBoundVar{A} \HOLBoundVar{B} \HOLBoundVar{C}. \HOLConst{arrow} \HOLBoundVar{X} (\HOLBoundVar{A} \HOLSymConst{\HOLTokenProd{}} \HOLBoundVar{B}) \HOLBoundVar{C} \HOLSymConst{\HOLTokenImp{}} \HOLConst{arrow} \HOLBoundVar{X} \HOLBoundVar{A} (\HOLBoundVar{C} \HOLSymConst{/} \HOLBoundVar{B})) \HOLSymConst{\HOLTokenConj{}}
   (\HOLSymConst{\HOLTokenForall{}}\HOLBoundVar{X} \HOLBoundVar{A} \HOLBoundVar{B} \HOLBoundVar{C}. \HOLConst{arrow} \HOLBoundVar{X} \HOLBoundVar{A} (\HOLBoundVar{C} \HOLSymConst{/} \HOLBoundVar{B}) \HOLSymConst{\HOLTokenImp{}} \HOLConst{arrow} \HOLBoundVar{X} (\HOLBoundVar{A} \HOLSymConst{\HOLTokenProd{}} \HOLBoundVar{B}) \HOLBoundVar{C}) \HOLSymConst{\HOLTokenConj{}}
   (\HOLSymConst{\HOLTokenForall{}}\HOLBoundVar{X} \HOLBoundVar{A} \HOLBoundVar{B} \HOLBoundVar{C}. \HOLConst{arrow} \HOLBoundVar{X} (\HOLBoundVar{A} \HOLSymConst{\HOLTokenProd{}} \HOLBoundVar{B}) \HOLBoundVar{C} \HOLSymConst{\HOLTokenImp{}} \HOLConst{arrow} \HOLBoundVar{X} \HOLBoundVar{B} (\HOLBoundVar{A} \HOLSymConst{\HOLTokenBackslash} \HOLBoundVar{C})) \HOLSymConst{\HOLTokenConj{}}
   (\HOLSymConst{\HOLTokenForall{}}\HOLBoundVar{X} \HOLBoundVar{A} \HOLBoundVar{B} \HOLBoundVar{C}. \HOLConst{arrow} \HOLBoundVar{X} \HOLBoundVar{B} (\HOLBoundVar{A} \HOLSymConst{\HOLTokenBackslash} \HOLBoundVar{C}) \HOLSymConst{\HOLTokenImp{}} \HOLConst{arrow} \HOLBoundVar{X} (\HOLBoundVar{A} \HOLSymConst{\HOLTokenProd{}} \HOLBoundVar{B}) \HOLBoundVar{C}) \HOLSymConst{\HOLTokenConj{}}
   (\HOLSymConst{\HOLTokenForall{}}\HOLBoundVar{X} \HOLBoundVar{A} \HOLBoundVar{B} \HOLBoundVar{C}. \HOLConst{arrow} \HOLBoundVar{X} \HOLBoundVar{A} \HOLBoundVar{B} \HOLSymConst{\HOLTokenConj{}} \HOLConst{arrow} \HOLBoundVar{X} \HOLBoundVar{B} \HOLBoundVar{C} \HOLSymConst{\HOLTokenImp{}} \HOLConst{arrow} \HOLBoundVar{X} \HOLBoundVar{A} \HOLBoundVar{C}) \HOLSymConst{\HOLTokenConj{}}
   \HOLSymConst{\HOLTokenForall{}}\HOLBoundVar{X} \HOLBoundVar{A} \HOLBoundVar{B}. \HOLBoundVar{X} \HOLBoundVar{A} \HOLBoundVar{B} \HOLSymConst{\HOLTokenImp{}} \HOLConst{arrow} \HOLBoundVar{X} \HOLBoundVar{A} \HOLBoundVar{B}
arrow_strongind:
\HOLTokenTurnstile{} (\HOLSymConst{\HOLTokenForall{}}\HOLBoundVar{X} \HOLBoundVar{A}. \HOLFreeVar{arrow\sp{\prime}} \HOLBoundVar{X} \HOLBoundVar{A} \HOLBoundVar{A}) \HOLSymConst{\HOLTokenConj{}}
   (\HOLSymConst{\HOLTokenForall{}}\HOLBoundVar{X} \HOLBoundVar{A} \HOLBoundVar{B} \HOLBoundVar{C}.
      \HOLConst{arrow} \HOLBoundVar{X} (\HOLBoundVar{A} \HOLSymConst{\HOLTokenProd{}} \HOLBoundVar{B}) \HOLBoundVar{C} \HOLSymConst{\HOLTokenConj{}} \HOLFreeVar{arrow\sp{\prime}} \HOLBoundVar{X} (\HOLBoundVar{A} \HOLSymConst{\HOLTokenProd{}} \HOLBoundVar{B}) \HOLBoundVar{C} \HOLSymConst{\HOLTokenImp{}}
      \HOLFreeVar{arrow\sp{\prime}} \HOLBoundVar{X} \HOLBoundVar{A} (\HOLBoundVar{C} \HOLSymConst{/} \HOLBoundVar{B})) \HOLSymConst{\HOLTokenConj{}}
   (\HOLSymConst{\HOLTokenForall{}}\HOLBoundVar{X} \HOLBoundVar{A} \HOLBoundVar{B} \HOLBoundVar{C}.
      \HOLConst{arrow} \HOLBoundVar{X} \HOLBoundVar{A} (\HOLBoundVar{C} \HOLSymConst{/} \HOLBoundVar{B}) \HOLSymConst{\HOLTokenConj{}} \HOLFreeVar{arrow\sp{\prime}} \HOLBoundVar{X} \HOLBoundVar{A} (\HOLBoundVar{C} \HOLSymConst{/} \HOLBoundVar{B}) \HOLSymConst{\HOLTokenImp{}}
      \HOLFreeVar{arrow\sp{\prime}} \HOLBoundVar{X} (\HOLBoundVar{A} \HOLSymConst{\HOLTokenProd{}} \HOLBoundVar{B}) \HOLBoundVar{C}) \HOLSymConst{\HOLTokenConj{}}
   (\HOLSymConst{\HOLTokenForall{}}\HOLBoundVar{X} \HOLBoundVar{A} \HOLBoundVar{B} \HOLBoundVar{C}.
      \HOLConst{arrow} \HOLBoundVar{X} (\HOLBoundVar{A} \HOLSymConst{\HOLTokenProd{}} \HOLBoundVar{B}) \HOLBoundVar{C} \HOLSymConst{\HOLTokenConj{}} \HOLFreeVar{arrow\sp{\prime}} \HOLBoundVar{X} (\HOLBoundVar{A} \HOLSymConst{\HOLTokenProd{}} \HOLBoundVar{B}) \HOLBoundVar{C} \HOLSymConst{\HOLTokenImp{}}
      \HOLFreeVar{arrow\sp{\prime}} \HOLBoundVar{X} \HOLBoundVar{B} (\HOLBoundVar{A} \HOLSymConst{\HOLTokenBackslash} \HOLBoundVar{C})) \HOLSymConst{\HOLTokenConj{}}
   (\HOLSymConst{\HOLTokenForall{}}\HOLBoundVar{X} \HOLBoundVar{A} \HOLBoundVar{B} \HOLBoundVar{C}.
      \HOLConst{arrow} \HOLBoundVar{X} \HOLBoundVar{B} (\HOLBoundVar{A} \HOLSymConst{\HOLTokenBackslash} \HOLBoundVar{C}) \HOLSymConst{\HOLTokenConj{}} \HOLFreeVar{arrow\sp{\prime}} \HOLBoundVar{X} \HOLBoundVar{B} (\HOLBoundVar{A} \HOLSymConst{\HOLTokenBackslash} \HOLBoundVar{C}) \HOLSymConst{\HOLTokenImp{}}
      \HOLFreeVar{arrow\sp{\prime}} \HOLBoundVar{X} (\HOLBoundVar{A} \HOLSymConst{\HOLTokenProd{}} \HOLBoundVar{B}) \HOLBoundVar{C}) \HOLSymConst{\HOLTokenConj{}}
   (\HOLSymConst{\HOLTokenForall{}}\HOLBoundVar{X} \HOLBoundVar{A} \HOLBoundVar{B} \HOLBoundVar{C}.
      \HOLConst{arrow} \HOLBoundVar{X} \HOLBoundVar{A} \HOLBoundVar{B} \HOLSymConst{\HOLTokenConj{}} \HOLFreeVar{arrow\sp{\prime}} \HOLBoundVar{X} \HOLBoundVar{A} \HOLBoundVar{B} \HOLSymConst{\HOLTokenConj{}} \HOLConst{arrow} \HOLBoundVar{X} \HOLBoundVar{B} \HOLBoundVar{C} \HOLSymConst{\HOLTokenConj{}} \HOLFreeVar{arrow\sp{\prime}} \HOLBoundVar{X} \HOLBoundVar{B} \HOLBoundVar{C} \HOLSymConst{\HOLTokenImp{}}
      \HOLFreeVar{arrow\sp{\prime}} \HOLBoundVar{X} \HOLBoundVar{A} \HOLBoundVar{C}) \HOLSymConst{\HOLTokenConj{}} (\HOLSymConst{\HOLTokenForall{}}\HOLBoundVar{X} \HOLBoundVar{A} \HOLBoundVar{B}. \HOLBoundVar{X} \HOLBoundVar{A} \HOLBoundVar{B} \HOLSymConst{\HOLTokenImp{}} \HOLFreeVar{arrow\sp{\prime}} \HOLBoundVar{X} \HOLBoundVar{A} \HOLBoundVar{B}) \HOLSymConst{\HOLTokenImp{}}
   \HOLSymConst{\HOLTokenForall{}}\HOLBoundVar{a\sb{\mathrm{0}}} \HOLBoundVar{a\sb{\mathrm{1}}} \HOLBoundVar{a\sb{\mathrm{2}}}. \HOLConst{arrow} \HOLBoundVar{a\sb{\mathrm{0}}} \HOLBoundVar{a\sb{\mathrm{1}}} \HOLBoundVar{a\sb{\mathrm{2}}} \HOLSymConst{\HOLTokenImp{}} \HOLFreeVar{arrow\sp{\prime}} \HOLBoundVar{a\sb{\mathrm{0}}} \HOLBoundVar{a\sb{\mathrm{1}}} \HOLBoundVar{a\sb{\mathrm{2}}}
arrow_ind:
\HOLTokenTurnstile{} (\HOLSymConst{\HOLTokenForall{}}\HOLBoundVar{X} \HOLBoundVar{A}. \HOLFreeVar{arrow\sp{\prime}} \HOLBoundVar{X} \HOLBoundVar{A} \HOLBoundVar{A}) \HOLSymConst{\HOLTokenConj{}}
   (\HOLSymConst{\HOLTokenForall{}}\HOLBoundVar{X} \HOLBoundVar{A} \HOLBoundVar{B} \HOLBoundVar{C}. \HOLFreeVar{arrow\sp{\prime}} \HOLBoundVar{X} (\HOLBoundVar{A} \HOLSymConst{\HOLTokenProd{}} \HOLBoundVar{B}) \HOLBoundVar{C} \HOLSymConst{\HOLTokenImp{}} \HOLFreeVar{arrow\sp{\prime}} \HOLBoundVar{X} \HOLBoundVar{A} (\HOLBoundVar{C} \HOLSymConst{/} \HOLBoundVar{B})) \HOLSymConst{\HOLTokenConj{}}
   (\HOLSymConst{\HOLTokenForall{}}\HOLBoundVar{X} \HOLBoundVar{A} \HOLBoundVar{B} \HOLBoundVar{C}. \HOLFreeVar{arrow\sp{\prime}} \HOLBoundVar{X} \HOLBoundVar{A} (\HOLBoundVar{C} \HOLSymConst{/} \HOLBoundVar{B}) \HOLSymConst{\HOLTokenImp{}} \HOLFreeVar{arrow\sp{\prime}} \HOLBoundVar{X} (\HOLBoundVar{A} \HOLSymConst{\HOLTokenProd{}} \HOLBoundVar{B}) \HOLBoundVar{C}) \HOLSymConst{\HOLTokenConj{}}
   (\HOLSymConst{\HOLTokenForall{}}\HOLBoundVar{X} \HOLBoundVar{A} \HOLBoundVar{B} \HOLBoundVar{C}. \HOLFreeVar{arrow\sp{\prime}} \HOLBoundVar{X} (\HOLBoundVar{A} \HOLSymConst{\HOLTokenProd{}} \HOLBoundVar{B}) \HOLBoundVar{C} \HOLSymConst{\HOLTokenImp{}} \HOLFreeVar{arrow\sp{\prime}} \HOLBoundVar{X} \HOLBoundVar{B} (\HOLBoundVar{A} \HOLSymConst{\HOLTokenBackslash} \HOLBoundVar{C})) \HOLSymConst{\HOLTokenConj{}}
   (\HOLSymConst{\HOLTokenForall{}}\HOLBoundVar{X} \HOLBoundVar{A} \HOLBoundVar{B} \HOLBoundVar{C}. \HOLFreeVar{arrow\sp{\prime}} \HOLBoundVar{X} \HOLBoundVar{B} (\HOLBoundVar{A} \HOLSymConst{\HOLTokenBackslash} \HOLBoundVar{C}) \HOLSymConst{\HOLTokenImp{}} \HOLFreeVar{arrow\sp{\prime}} \HOLBoundVar{X} (\HOLBoundVar{A} \HOLSymConst{\HOLTokenProd{}} \HOLBoundVar{B}) \HOLBoundVar{C}) \HOLSymConst{\HOLTokenConj{}}
   (\HOLSymConst{\HOLTokenForall{}}\HOLBoundVar{X} \HOLBoundVar{A} \HOLBoundVar{B} \HOLBoundVar{C}. \HOLFreeVar{arrow\sp{\prime}} \HOLBoundVar{X} \HOLBoundVar{A} \HOLBoundVar{B} \HOLSymConst{\HOLTokenConj{}} \HOLFreeVar{arrow\sp{\prime}} \HOLBoundVar{X} \HOLBoundVar{B} \HOLBoundVar{C} \HOLSymConst{\HOLTokenImp{}} \HOLFreeVar{arrow\sp{\prime}} \HOLBoundVar{X} \HOLBoundVar{A} \HOLBoundVar{C}) \HOLSymConst{\HOLTokenConj{}}
   (\HOLSymConst{\HOLTokenForall{}}\HOLBoundVar{X} \HOLBoundVar{A} \HOLBoundVar{B}. \HOLBoundVar{X} \HOLBoundVar{A} \HOLBoundVar{B} \HOLSymConst{\HOLTokenImp{}} \HOLFreeVar{arrow\sp{\prime}} \HOLBoundVar{X} \HOLBoundVar{A} \HOLBoundVar{B}) \HOLSymConst{\HOLTokenImp{}}
   \HOLSymConst{\HOLTokenForall{}}\HOLBoundVar{a\sb{\mathrm{0}}} \HOLBoundVar{a\sb{\mathrm{1}}} \HOLBoundVar{a\sb{\mathrm{2}}}. \HOLConst{arrow} \HOLBoundVar{a\sb{\mathrm{0}}} \HOLBoundVar{a\sb{\mathrm{1}}} \HOLBoundVar{a\sb{\mathrm{2}}} \HOLSymConst{\HOLTokenImp{}} \HOLFreeVar{arrow\sp{\prime}} \HOLBoundVar{a\sb{\mathrm{0}}} \HOLBoundVar{a\sb{\mathrm{1}}} \HOLBoundVar{a\sb{\mathrm{2}}}
arrow_cases:
\HOLTokenTurnstile{} \HOLConst{arrow} \HOLFreeVar{a\sb{\mathrm{0}}} \HOLFreeVar{a\sb{\mathrm{1}}} \HOLFreeVar{a\sb{\mathrm{2}}} \HOLSymConst{\HOLTokenEquiv{}}
   (\HOLFreeVar{a\sb{\mathrm{2}}} \HOLSymConst{=} \HOLFreeVar{a\sb{\mathrm{1}}}) \HOLSymConst{\HOLTokenDisj{}} (\HOLSymConst{\HOLTokenExists{}}\HOLBoundVar{B} \HOLBoundVar{C}. (\HOLFreeVar{a\sb{\mathrm{2}}} \HOLSymConst{=} \HOLBoundVar{C} \HOLSymConst{/} \HOLBoundVar{B}) \HOLSymConst{\HOLTokenConj{}} \HOLConst{arrow} \HOLFreeVar{a\sb{\mathrm{0}}} (\HOLFreeVar{a\sb{\mathrm{1}}} \HOLSymConst{\HOLTokenProd{}} \HOLBoundVar{B}) \HOLBoundVar{C}) \HOLSymConst{\HOLTokenDisj{}}
   (\HOLSymConst{\HOLTokenExists{}}\HOLBoundVar{A} \HOLBoundVar{B}. (\HOLFreeVar{a\sb{\mathrm{1}}} \HOLSymConst{=} \HOLBoundVar{A} \HOLSymConst{\HOLTokenProd{}} \HOLBoundVar{B}) \HOLSymConst{\HOLTokenConj{}} \HOLConst{arrow} \HOLFreeVar{a\sb{\mathrm{0}}} \HOLBoundVar{A} (\HOLFreeVar{a\sb{\mathrm{2}}} \HOLSymConst{/} \HOLBoundVar{B})) \HOLSymConst{\HOLTokenDisj{}}
   (\HOLSymConst{\HOLTokenExists{}}\HOLBoundVar{A} \HOLBoundVar{C}. (\HOLFreeVar{a\sb{\mathrm{2}}} \HOLSymConst{=} \HOLBoundVar{A} \HOLSymConst{\HOLTokenBackslash} \HOLBoundVar{C}) \HOLSymConst{\HOLTokenConj{}} \HOLConst{arrow} \HOLFreeVar{a\sb{\mathrm{0}}} (\HOLBoundVar{A} \HOLSymConst{\HOLTokenProd{}} \HOLFreeVar{a\sb{\mathrm{1}}}) \HOLBoundVar{C}) \HOLSymConst{\HOLTokenDisj{}}
   (\HOLSymConst{\HOLTokenExists{}}\HOLBoundVar{A} \HOLBoundVar{B}. (\HOLFreeVar{a\sb{\mathrm{1}}} \HOLSymConst{=} \HOLBoundVar{A} \HOLSymConst{\HOLTokenProd{}} \HOLBoundVar{B}) \HOLSymConst{\HOLTokenConj{}} \HOLConst{arrow} \HOLFreeVar{a\sb{\mathrm{0}}} \HOLBoundVar{B} (\HOLBoundVar{A} \HOLSymConst{\HOLTokenBackslash} \HOLFreeVar{a\sb{\mathrm{2}}})) \HOLSymConst{\HOLTokenDisj{}}
   (\HOLSymConst{\HOLTokenExists{}}\HOLBoundVar{B}. \HOLConst{arrow} \HOLFreeVar{a\sb{\mathrm{0}}} \HOLFreeVar{a\sb{\mathrm{1}}} \HOLBoundVar{B} \HOLSymConst{\HOLTokenConj{}} \HOLConst{arrow} \HOLFreeVar{a\sb{\mathrm{0}}} \HOLBoundVar{B} \HOLFreeVar{a\sb{\mathrm{2}}}) \HOLSymConst{\HOLTokenDisj{}} \HOLFreeVar{a\sb{\mathrm{0}}} \HOLFreeVar{a\sb{\mathrm{1}}} \HOLFreeVar{a\sb{\mathrm{2}}}
\end{alltt}
With all above theorems, any theorem in which the relation \texttt{arrow}
is used, can be handled by combining above theorems with other related
theorems.

Here is the essential differences between Coq and HOL that we have
observed on inductive relations: In HOL, terms like \HOLinline{\HOLConst{arrow} \HOLFreeVar{X} \HOLFreeVar{A} \HOLFreeVar{B}} has type
 \texttt{bool}; while in Coq, its type is \texttt{Set}.

Here is the consequence: in HOL, in terms like \HOLinline{\HOLConst{arrow} \HOLFreeVar{X} \HOLFreeVar{A} \HOLFreeVar{B} \HOLSymConst{\HOLTokenConj{}} \HOLConst{arrow} \HOLFreeVar{X} \HOLFreeVar{B} \HOLFreeVar{C}} or \HOLinline{\HOLConst{arrow} \HOLFreeVar{X} \HOLFreeVar{A} \HOLFreeVar{B} \HOLSymConst{\HOLTokenImp{}} \HOLConst{arrow} \HOLFreeVar{X} \HOLFreeVar{A} \HOLFreeVar{C}}, they're normal logical connectives between boolean
values (first is ``and'', second is ``implies''). But in Coq, logical
connectives never appears between two \texttt{Set}s. Instead, theorems
like \texttt{A /\ B ==> C} were always represented as \texttt{A -> B
  -> C} in which \texttt{->} serves as logical implication but
actually has more complex meanings.

\subsection{Further on logical connectives}

In HOL, basic Boolean connectives and first-order logic quantifiers like ``forall'', ``exists'', ``and'', ``or'', ``not'' and even
``true'', ``false'', are all defined as $\lambda$ terms:
\begin{alltt}
\HOLTokenTurnstile{} \HOLConst{T} \HOLSymConst{\HOLTokenEquiv{}} ((\HOLTokenLambda{}\HOLBoundVar{x}. \HOLBoundVar{x}) \HOLSymConst{=} (\HOLTokenLambda{}\HOLBoundVar{x}. \HOLBoundVar{x}))
\HOLTokenTurnstile{} (\HOLSymConst{\HOLTokenForall{}}) \HOLSymConst{=} (\HOLTokenLambda{}\HOLBoundVar{P}. \HOLBoundVar{P} \HOLSymConst{=} (\HOLTokenLambda{}\HOLBoundVar{x}. \HOLConst{T}))
\HOLTokenTurnstile{} (\HOLSymConst{\HOLTokenExists{}}) \HOLSymConst{=} (\HOLTokenLambda{}\HOLBoundVar{P}. \HOLBoundVar{P} ((\HOLSymConst{\HOLTokenHilbert{}}) \HOLBoundVar{P}))
\HOLTokenTurnstile{} (\HOLSymConst{\HOLTokenConj{}}) \HOLSymConst{=} (\HOLTokenLambda{}\HOLBoundVar{t\sb{\mathrm{1}}} \HOLBoundVar{t\sb{\mathrm{2}}}. \HOLSymConst{\HOLTokenForall{}}\HOLBoundVar{t}. (\HOLBoundVar{t\sb{\mathrm{1}}} \HOLSymConst{\HOLTokenImp{}} \HOLBoundVar{t\sb{\mathrm{2}}} \HOLSymConst{\HOLTokenImp{}} \HOLBoundVar{t}) \HOLSymConst{\HOLTokenImp{}} \HOLBoundVar{t})
\HOLTokenTurnstile{} (\HOLSymConst{\HOLTokenDisj{}}) \HOLSymConst{=} (\HOLTokenLambda{}\HOLBoundVar{t\sb{\mathrm{1}}} \HOLBoundVar{t\sb{\mathrm{2}}}. \HOLSymConst{\HOLTokenForall{}}\HOLBoundVar{t}. (\HOLBoundVar{t\sb{\mathrm{1}}} \HOLSymConst{\HOLTokenImp{}} \HOLBoundVar{t}) \HOLSymConst{\HOLTokenImp{}} (\HOLBoundVar{t\sb{\mathrm{2}}} \HOLSymConst{\HOLTokenImp{}} \HOLBoundVar{t}) \HOLSymConst{\HOLTokenImp{}} \HOLBoundVar{t})
\HOLTokenTurnstile{} \HOLConst{F} \HOLSymConst{\HOLTokenEquiv{}} \HOLSymConst{\HOLTokenForall{}}\HOLBoundVar{t}. \HOLBoundVar{t}
\HOLTokenTurnstile{} (\HOLSymConst{\HOLTokenNeg{}}) \HOLSymConst{=} (\HOLTokenLambda{}\HOLBoundVar{t}. \HOLBoundVar{t} \HOLSymConst{\HOLTokenImp{}} \HOLConst{F})
\HOLTokenTurnstile{} (\HOLSymConst{\HOLTokenUnique{}}) \HOLSymConst{=} (\HOLTokenLambda{}\HOLBoundVar{P}. (\HOLSymConst{\HOLTokenExists{}}) \HOLBoundVar{P} \HOLSymConst{\HOLTokenConj{}} \HOLSymConst{\HOLTokenForall{}}\HOLBoundVar{x} \HOLBoundVar{y}. \HOLBoundVar{P} \HOLBoundVar{x} \HOLSymConst{\HOLTokenConj{}} \HOLBoundVar{P} \HOLBoundVar{y} \HOLSymConst{\HOLTokenImp{}} (\HOLBoundVar{x} \HOLSymConst{=} \HOLBoundVar{y}))
\end{alltt}
Since they're all $\lambda$ terms, any facts about their relationship
can be proved within the framework of $\lambda$-calculus, using
$\beta$-reductions and other basic deduction rules. That's so clear!

While in Coq, it's surprised to notice that, the quantifier
\texttt{forall} is something quite primitive: it's a keyword in
Coq. While there's no \texttt{exists} keyword at all. And to express
the existence of something in any theorem, user must write it as a
lambda function.  For instance, our \texttt{replace_inv2} theorem in
HOL contains two existence quantifier variables:
\begin{alltt}
\HOLTokenTurnstile{} \HOLConst{replace} (\HOLConst{Comma} \HOLFreeVar{Gamma\sb{\mathrm{1}}} \HOLFreeVar{Gamma\sb{\mathrm{2}}}) \HOLFreeVar{Gamma\sp{\prime}} (\HOLConst{OneForm} \HOLFreeVar{X}) \HOLFreeVar{Delta} \HOLSymConst{\HOLTokenImp{}}
   (\HOLSymConst{\HOLTokenExists{}}\HOLBoundVar{G}.
      (\HOLFreeVar{Gamma\sp{\prime}} \HOLSymConst{=} \HOLConst{Comma} \HOLBoundVar{G} \HOLFreeVar{Gamma\sb{\mathrm{2}}}) \HOLSymConst{\HOLTokenConj{}}
      \HOLConst{replace} \HOLFreeVar{Gamma\sb{\mathrm{1}}} \HOLBoundVar{G} (\HOLConst{OneForm} \HOLFreeVar{X}) \HOLFreeVar{Delta}) \HOLSymConst{\HOLTokenDisj{}}
   \HOLSymConst{\HOLTokenExists{}}\HOLBoundVar{G}.
     (\HOLFreeVar{Gamma\sp{\prime}} \HOLSymConst{=} \HOLConst{Comma} \HOLFreeVar{Gamma\sb{\mathrm{1}}} \HOLBoundVar{G}) \HOLSymConst{\HOLTokenConj{}}
     \HOLConst{replace} \HOLFreeVar{Gamma\sb{\mathrm{2}}} \HOLBoundVar{G} (\HOLConst{OneForm} \HOLFreeVar{X}) \HOLFreeVar{Delta}
\end{alltt}
The meaning of above theorem is quite clear just by reading it.  While
in Coq, the same theorem must be expressed in this strange way:
\begin{lstlisting}
Lemma replace_inv2 :
 forall (Gamma1 Gamma2 Gamma' Delta : Term Atoms) (X : Form Atoms),
 replace (Comma Gamma1 Gamma2) Gamma' (OneForm X) Delta ->
 sigS
   (fun Gamma'1 : Term Atoms =>
    {x_ : replace Gamma1 Gamma'1 (OneForm X) Delta |
    Gamma' = Comma Gamma'1 Gamma2}) +
 sigS
   (fun Gamma'2 : Term Atoms =>
    {x_ : replace Gamma2 Gamma'2 (OneForm X) Delta |
    Gamma' = Comma Gamma1 Gamma'2}).
\end{lstlisting}
please notice that, how logical ``and'' and ``or'' must be expressed
as \texttt{|} and \texttt{+} between \texttt{Set}s in Coq.  The author
hope these examples could convince the readers that, Coq is very
unnatural for representing logical theorems.

\subsection{On Coq's built-in supports of proofs}

In previous section, we have mentioned that, Coq has built-in supports
on ``proofs''.  This fact seems coming from the fact that, in Coq,
each logical theorems has essentially the type \texttt{Set} which is
indeed a mathematical set, and each element in such sets is one
possible ``proof'' for that theorem!  This is really a convenient
feature when people need to prove results about proof themselves, but
the drawback is, most of such theorems has large amount of variables.

For instance, our last theorem in \texttt{CutFreeTheory} is
\texttt{subFormulaProperty}, which has the following representation in
HOL:
\begin{alltt}
\HOLTokenTurnstile{} \HOLConst{subProof} \HOLFreeVar{q} \HOLFreeVar{p} \HOLSymConst{\HOLTokenImp{}}
   \HOLConst{extensionSub} (\HOLConst{exten} \HOLFreeVar{p}) \HOLSymConst{\HOLTokenImp{}}
   \HOLConst{CutFreeProof} \HOLFreeVar{p} \HOLSymConst{\HOLTokenImp{}}
   \HOLSymConst{\HOLTokenForall{}}\HOLBoundVar{x}.
     \HOLConst{subFormTerm} \HOLBoundVar{x} (\HOLConst{prems} \HOLFreeVar{q}) \HOLSymConst{\HOLTokenDisj{}} \HOLConst{subFormula} \HOLBoundVar{x} (\HOLConst{concl} \HOLFreeVar{q}) \HOLSymConst{\HOLTokenImp{}}
     \HOLConst{subFormTerm} \HOLBoundVar{x} (\HOLConst{prems} \HOLFreeVar{p}) \HOLSymConst{\HOLTokenDisj{}} \HOLConst{subFormula} \HOLBoundVar{x} (\HOLConst{concl} \HOLFreeVar{p})
\end{alltt}

This theorem was actually ported from Coq, from the following theorem:
\begin{lstlisting}
Lemma subFormulaProperty :
 forall (Atoms : Set) (Gamma1 Gamma2 : Term Atoms) 
   (B C x : Form Atoms) (E : gentzen_extension)
   (p : gentzenSequent E Gamma1 B) (q : gentzenSequent E Gamma2 C),
 extensionSub Atoms E ->
 subProof q p ->
 CutFreeProof p ->
 subFormTerm x Gamma2 \/ subFormula x C ->
 subFormTerm x Gamma1 \/ subFormula x B.
\end{lstlisting}
Now let's count how many variables are used in above theorem in Coq: \texttt{Atoms},
\texttt{Gamma1}, \texttt{Gamma2}, \texttt{B}, \texttt{C}, \texttt{x},
\texttt{E}, \texttt{p}, \texttt{q}, 9 variables totally.  While in the
HOL version, 3 variables are just enough (two ``proofs'' plus one Term).

Once we have proved above HOL theorem, we can also easily prove the following
theorem which has the same amount of variables as Coq:
\begin{alltt}
\HOLTokenTurnstile{} (\HOLFreeVar{p} \HOLSymConst{=} \HOLConst{Der} (\HOLConst{Sequent} \HOLFreeVar{E} \HOLFreeVar{Gamma\sb{\mathrm{1}}} \HOLFreeVar{B}) \HOLFreeVar{\HOLTokenUnderscore{}} \HOLFreeVar{\HOLTokenUnderscore{}}) \HOLSymConst{\HOLTokenImp{}}
   (\HOLFreeVar{q} \HOLSymConst{=} \HOLConst{Der} (\HOLConst{Sequent} \HOLFreeVar{E} \HOLFreeVar{Gamma\sb{\mathrm{2}}} \HOLFreeVar{C}) \HOLFreeVar{\HOLTokenUnderscore{}} \HOLFreeVar{\HOLTokenUnderscore{}}) \HOLSymConst{\HOLTokenImp{}}
   \HOLConst{extensionSub} \HOLFreeVar{E} \HOLSymConst{\HOLTokenImp{}}
   \HOLConst{subProof} \HOLFreeVar{q} \HOLFreeVar{p} \HOLSymConst{\HOLTokenImp{}}
   \HOLConst{CutFreeProof} \HOLFreeVar{p} \HOLSymConst{\HOLTokenImp{}}
   \HOLConst{subFormTerm} \HOLFreeVar{x} \HOLFreeVar{Gamma\sb{\mathrm{2}}} \HOLSymConst{\HOLTokenDisj{}} \HOLConst{subFormula} \HOLFreeVar{x} \HOLFreeVar{C} \HOLSymConst{\HOLTokenImp{}}
   \HOLConst{subFormTerm} \HOLFreeVar{x} \HOLFreeVar{Gamma\sb{\mathrm{1}}} \HOLSymConst{\HOLTokenDisj{}} \HOLConst{subFormula} \HOLFreeVar{x} \HOLFreeVar{B}
\end{alltt}
but the other direction is not easy (maybe just impossible): if we
have already this last theorem in HOL, no way to get the previous
theorem with only 3 variables.

Nevertheless, for Coq's built-in supports of proof, HOL users can
prove the same theorems, although they have to invent the concept and
structure of ``proofs'' from almost ground, and case by case. But the
author thinks, it's quite fair to say that, all theorems provers have
exactly the same set of theorems that they're capable to prove.  So
the remain question is how to choose between them. Most of time, it's
just a matter of personal preferences, experiences and environment
requirements (e.g. when you were doing formalization studies in France, you have to use Coq and OCaml, which
are both invented by Franch people).

\section{Future directions}

From the view of theorem proving, the following theorems are worth to
prove in the future:
\begin{enumerate}
\item Cut-elimination theorem for Lambek Calculus with arbitrary
  sequent extensions.
\item Lambek calculus \textbf{L} is context-free. \cite{Pentus:1993cy}
\end{enumerate}
The cut-elimination theorem can be proved directly using the existing
framework in our CutFreeTheory, while for the second goal, a full
treatment of Lambek Calculus in model-theoretic approach with many
new foundamental definitions and theorems must be done first.

From the view of language parsing, the following goals remain to be
finished:
\begin{enumerate}
\item Implement an automatic proof searching algorithm for Sequent Calculus as a
  HOL tactical.
\item Implement the same algorithm for generating first-class proof
  trees (Dertree).
\item Implement a language parser as ML functions which generates both
  parsing trees and validation theorems.
\item Design a lexicon data structure for holding large amount of
  words, each with more than one categories.
\item Design and implement machine learning algorithms to build a
  large enough lexicon for Italian language.
\end{enumerate}

\section{Conclusions}

In this project, we have implemented a rather complete proof-theoretical formalization of Lambek
Calculus (non-associative, with arbitrary extensions).

The current status is enough as a tool kit for manually proving
category theorems in three deduction systems of Lambek Calculus:
Axiomatic Syntactic Calculus, Natural Deduction (in Gentzen style) and
Gentzen's Sequent
Calculus. It can also be considered as a base framework for further formalization
of more deep theorems of Lambek Calculus, e.g. the cut-elimination
theorem of Gentzen's Sequent Calculus of Lambek Calculus.

This work is based on the Lambek Calculus formalization in Coq, by
Houda ANOUN and Pierre Casteran in 2002-2003. For the modules that we have
migrated from Coq, we improved definitions and proved many new
theorems. For the proof-theoretic formalization of Sequent Calculus
proofs, our work (first-class proof trees in HOL, including related
derivation definitions and theorems) is completely new.

Thanks to Prof. Fabio Tambrini, who has introduced Lambek Calculus to the
author in his NLP course at University of Bologna.

\bibliography{hol-lambek}{}
\bibliographystyle{splncs}

\end{document}